\documentclass[10pt,journal,compsoc]{IEEEtran}

\ifCLASSOPTIONcompsoc
  \usepackage[nocompress]{cite}
\else
  \usepackage{cite}
\fi

\usepackage[inline]{enumitem}
\usepackage[normalem]{ulem}
\usepackage[skins,breakable]{tcolorbox}
\usepackage{amsmath,amsfonts}
\usepackage{array}
\usepackage{hhline}
\usepackage[group-separator={,},group-minimum-digits=4]{siunitx}

\usepackage{microtype}
\usepackage{framed}
\usepackage{lipsum}
\usepackage{listings}
\usepackage{longtable}
\usepackage{mathtools}
\usepackage{multirow}
\usepackage{pbox}
\usepackage{relsize}
\usepackage{rotating}
\usepackage{tabularx}
\newcolumntype{Y}{>{\centering\arraybackslash}X}
\usepackage{booktabs}
\usepackage{url}
\usepackage{subcaption}
\usepackage{graphicx}
\usepackage{xspace}
\usepackage{makecell}
\usepackage{float}
\usepackage{xcolor}

\definecolor{backcolour}{rgb}{0.95,0.95,0.92}
\definecolor{maroon}{rgb}{0.5,0,0}
\definecolor{blue-violet}{rgb}{0.54, 0.17, 0.89}

\newcommand{\BT}{Behavior Trees\xspace}
\newcommand{\Bts}{Behavior trees\xspace}
\newcommand{\Bt}{Behavior tree\xspace}
\newcommand{\bt}{behavior tree\xspace}
\newcommand{\bts}{behavior trees\xspace}

\newcommand{\sm}{state machine\xspace}

\newcommand{\Sms}{State machines\xspace}
\newcommand{\sms}{state machines\xspace}

\newcommand{\hsm}{hierarchical state-machine\xspace}

\newcommand{\BTCPP}{\texttt{Behavior\-Tree.CPP}\xspace}
\newcommand{\pytrees}{\lstinline{PyTrees}\xspace}
\newcommand{\pytreesros}{\lstinline{PyTrees_ros}\xspace}
\newcommand{\smach}{\lstinline{SMACH}\xspace}
\newcommand{\smachros}{\lstinline{smach_ros}\xspace}
\newcommand{\flexbe}{\lstinline{FlexBe}\xspace}

\usepackage{hyperref}
\hypersetup{
	 hidelinks = true 
}

\definecolor{xtextBlue}{RGB}{42,8,255}
\definecolor{blue(ncs)}{rgb}{0.0, 0.53, 0.74}
\newcommand{\MYhref}[3][blue(ncs)]{\href{#2}{\color{#1}{#3}}}%

\usepackage[capitalize]{cleveref}
\crefname{section}{Sect.}{sections}
\Crefname{section}{Section}{Sections}
\crefname{table}{Table}{tables}
\Crefname{table}{Table}{Tables}

\crefformat{footnote}{#2\footnotemark[#1]#3}

\newcommand{\chg}[1]{\textcolor{black}{{#1}}} 
\newcommand{\bchg}[1]{\textcolor{black}{{#1}}} 

\usepackage{ifthen}
\usepackage{amssymb}
\newboolean{showcomments}
\setboolean{showcomments}{true} 
\ifthenelse{\boolean{showcomments}}
{\newcommand{\nb}[2]{
		\fcolorbox{gray}{yellow}{\bfseries\sffamily\scriptsize#1}
		{\sf\small$\blacktriangleright$\textit{#2}$\blacktriangleleft$}
	}
	
}
{\newcommand{\nb}[2]{}
	
}

\newcounter{ObservationIdx}
\newenvironment{observation}%
{\begin{leftbar}
		\refstepcounter{ObservationIdx}
		\noindent\textsc{Observation\,\theObservationIdx.}\@ }%
	{\end{leftbar}}

\makeatletter
\def\endthebibliography{%
	\def\@noitemerr{\@latex@warning{Empty `thebibliography' environment}}%
	\endlist
}
\makeatother

\begin{document}


\title{Behavior Trees and State Machines \\ in Robotics Applications}

\author{Razan~Ghzouli, 
	Thorsten~Berger,
	Einar~Broch~Johnsen,
	Andrzej~Wasowski
	and Swaib~Dragule 
}

%

\IEEEtitleabstractindextext{%

\begin{abstract}
  Autonomous robots combine skills to form increasingly complex
  behaviors, called missions. While skills are often programmed at a
  relatively low abstraction level, their coordination is
  architecturally separated and often expressed in higher-level
  languages or frameworks. \Sms have been the go-to language \chg{to
    model behavior} for decades, but recently, \bts have gained
  attention among roboticists. Originally designed to model autonomous
  actors in computer games, \bts offer an extensible tree-based
  representation of missions and are claimed to support modular design
  and code reuse. Although several implementations of \bts are in use,
  little is known about their usage and scope in the real world.  How
  do concepts offered by \bts relate to traditional languages, such as
  state machines? How are concepts in \bts and \sms used in actual
  applications?

\medskip

This paper is a study of the key language concepts in \bts \chg{as
  realized in \chg{domain-specific languages (DSLs)}, internal and external DSLs offered as libraries,} and their use in
\chg{open-source} robotic applications \chg{supported by the Robot
  Operating System (ROS)}. We analyze behavior-tree \chg{DSLs} and compare
\chg{them} to the \chg{standard} language \chg{for behavior models} in
robotics: \sms. \chg{We identify DSLs for both behavior-modeling languages, and we analyze five in-depth.}
We mine \chg{open-source} repositories for robotic
applications that use the \chg{analyzed DSLs} and analyze their usage.  We
identify similarities between \bts and \sms in terms of language
design and \chg{the concepts offered} to accommodate the needs of the
robotics domain.  \chg{We observed that the} usage of behavior-tree DSLs in open-source projects is
increasing rapidly. \chg{We observed similar} usage patterns \chg{at model structure}
and at code reuse  \chg{in the behavior-tree and state-machine models within the mined open-source projects}. We
contribute all extracted models as a dataset, hoping to
inspire the community to use and further develop \bts, associated
tools, and analysis techniques.
\end{abstract}

}

\maketitle

\IEEEdisplaynontitleabstractindextext

%
\IEEEpeerreviewmaketitle

\section{Introduction}
\label{sec:intro}
The robots are coming!  They can perform tasks in environments that
defy human presence, such as fire fighting in dangerous areas or
disinfection in contaminated hospitals.  Robots can handle
increasingly difficult tasks, ranging from pick-and-place operations
to complex services performed while navigating in dynamic
environments.  Robots combine skills to form complex behaviors, known
as missions\,\cite{garcia2020robotics,menghi.ea:2019:tse,dragule2021bookchapter}. While
skills are typically programmed at a relatively low level of
abstraction (such as controllers for sensors and actuators), the
coordination of skills to form missions is either programmed at a
low level, intimately tied to the implementation of skills, or with
higher-level representations using behavior models. With the
increasing complexity of robot systems, higher-level representations
of missions have become increasingly important to improve software
quality and
maintainability\,\cite{michaud2016behavior,kortenkamp2016robotic,brugali2010component}.

\Sms are among the most common notations for \chg{describing behavior
  models in robotic missions} \cite{garcia2020robotics,
  colledanchise2018behavior, chen2018development}. Recently, \bts are
attracting the attention of roboticists to express such high-level
coordination. \chg{Both models describe pre-defined missions with
  limited decision making.}  \Bts were invented to model the behavior
of autonomous non-player characters in computer games. Similar to
autonomous robots, these are reactive and make decisions in complex
environments \cite{heckel2010representational,isla2005gdc,
  Mcquillan2015}.

\bchg{Many researchers have observed that \bts can offer ease of
  reuse, modularity, and flexibility when modeling reactive behavior
  \cite{biggar2021expressiveness, iovino2022programming,
    colledanchise2018behavior, bagnell2012integrated,
    colledanchise2016advantages, colledanchise2016behavior,
    colledanchise2018learning, marzinotto2014towards,
    ogren2012increasing, rovida2017extended, garcia2019high,
    klockner2013interfacing}. These works mainly focus on theoretical
  and proof-of-concept aspects of behavior-tree models. There is
  a trend in the robotics community to create domain-specific
  languages (DSLs)\,\cite{dslbook} and model-driven tools to engineer software for
  robotics systems \cite{de2021survey, casalaro2021model,
    nordmann2014survey}, but we need to better understand and improve
  these solutions to make them usable in practice \cite{casalaro2021model}.  Multiple DSLs have been created to
  support the implementation of behavior-tree and state-machine
  models, but we lack studies to understand these implementations from
  a software-engineering perspective and their use in real-world
  projects.}

\bchg{Our study bridges this knowledge gap by studying the actual concepts offered in real implementations of behavior-model DSLs and their usage in practice. We compare DSLs for \bts, the
  emerging language in robotics, to \sms, the traditional choice of
  roboticists. In this paper, we explore the concepts offered by \bts
  and \sms as realized in five DSLs (internal and external DSLs offered as libraries) and compare
  them. The scope of our comparison is behavior-tree DSLs and their offered concepts. We further study how the language concepts are exploited by
  the users of the analyzed libraries, based on an analysis of their
  usage in open-source ROS projects. Our goal is to obtain empirical data on the use of these behavior-modeling languages in practice.}

\subsubsection*{Goal and Research Questions}
\bchg{We present an exploratory study of behavior-tree and
  state-machine languages and their use in open-source robotic
  applications. While mostly qualitative, we provide quantitative data
  about the models we mined, and indicate the frequency of different
  concepts and phenomena we observe in practice. We formulated the
  following research questions:}

\medskip
\noindent\textbf{RQ1.} \emph{What modeling concepts from \bts and \sms are available in
  language implementations (libraries)?}

We identified five \chg{DSLs} (libraries) for \bts and \sms that
support ROS \cite{quigley2009ros}---a middleware and framework for
developing robotics applications, surrounded by the largest currently
existing ecosystem of robotics libraries (ROS packages)---and are
actively maintained and documented. Then, we identified and analyzed
the concepts \chg{offered by these DSLs}.  Our goal was to understand
the available behavior-tree and state-machine concepts in practice,
and the similarities and differences between these.

\medskip

\noindent\textbf{RQ2.}
\emph{How are these languages (libraries) engineered in practice?}

For the identified \chg{behavior-tree} and \chg{state-machine}
\chg{DSLs}, we studied the key design principles and language
implementations.  By studying the language implementations, we capture
the techniques and practices that are used in the robotics
community. By analyzing languages that originate from the practice of
roboticists, we can understand the design concepts needed in robotics
better and discuss potential improvements.

\medskip

\noindent\textbf{RQ3.} \emph{How are behavior-tree and state-machine models used in
  robotics projects?}

We mined GitHub for open-source repositories that use the identified
behavior-tree and state-machine \chg{DSLs}, and we analyzed their
usage in robotics applications. We checked the usage trends of the
languages, which can indicate their popularity. We extracted a sample
of the mined projects and analyzed the usage of behavior-tree and
state-machine concepts and the structure of the models in these
projects. Finally, we investigated reuse mechanisms in the projects by
a combination of visual and code-level inspections. In summary, we
report on empirical results regarding the use of \bts and \sms in
open-source \chg{ROS} robotics projects in terms of the popularity of
the \chg{DSLs}, the structure of the models, the usage frequency of
identified concepts, and the state-of-practice of reuse in
behavior-tree and state-machine \chg{DSLs}.

\subsubsection*{Journal Extension}
\looseness=-1 This paper extends a study of \bts, presented at SLE
2020 \cite{ghzouli2020behavior}. Compared to that paper, we here
broaden the scope of our study to include \sms. We extract
state-machine modeling concepts and compare them to behavior-tree
concepts extracted in the former work. We update the collection of
\chg{open-source} behavior-tree models and additionally mine
\chg{open-source} robotics projects that uses state-machine
\chg{DSLs}. \bchg{We improve the selection process of included
  projects (filtration) and automate the process using scripts. The
  new filtration process is applied to all the mined
  projects. Finally, we qualitatively and quantitatively analyze a
  random sample of the mined state-machine models that matches the
  number of previously analyzed behavior-tree models, to understand
  the usage of the studied behavior-modeling languages in open-source
  projects.} Thus, compared to the former paper, our focus is shifted
from only \bts to also include a comparison with \sms in terms of
concepts and robotics applications.

\subsubsection*{Results}
Our analysis of the modeling concepts offered in behavior-tree and
state-machine \chg{DSLs} (\textbf{RQ1}) shows similarities in terms of
language design and offered concepts. The languages are open and
support domain-specific patterns, which reflect common needs in
robotics.  Openness is a common feature in the studied \chg{DSLs},
which do not enforce a fixed model, but allow, even expect, concrete
projects to extend the DSLs by-need. Another observation is that all
studied \chg{DSLs} offer constructs for frequent control-flow
patterns. Although the range of support differs between behavior-tree
and state-machine \chg{DSLs}, accommodating the needs of the users of
a specific domain is good language design practice.

We have observed in our analysis of the \chg{DSL} design
(\textbf{RQ2}) that having a visualization tool for model construction
and monitoring could provide a better understanding of the models and
improved code reuse. In the \chg{DSLs} that feature a graphical
notation with a GUI-based editor, projects were more often using
built-in language constructs, such as Decorators in \bts and
Concurrency containers in \sms (introduced in
\Cref{sec:concepts}). Projects using \chg{DSLs} without GUIs leaked
these constructs to the code instead of the model. This is closely
related to the fact that \chg{DSLs} with no GUI are considered
internal while DSLs with GUIs are external. External DSLs enforce the
use of the DSL constructs (like decorators), while it is easy in
internal DSLs to deviate and use ordinary programming-language
constructs. Although it is a pragmatic practice, internal DSLs will
hinder maintainability and analyzability of the behavior model in the
long term.  Another observation is that all the \chg{DSLs} follow the
\emph{models-at-runtime}
paradigm\,\cite{bencomo14modelsatruntime,blair2009models}: models
coordinate skills, actions, and tasks, which are implemented by
lower-level means, such as ROS components.
 
Our analysis of the sampled models of robotics projects using behavior-tree and
state-machine \chg{DSLs} (\textbf{RQ3}) has provided insights into the
usage of these languages in practice. First, the usage of \bts is
rapidly increasing within \chg{open-source robotics projects}.
Second, the \chg{DSL} usage among developers was \chg{similar} from a
structural perspective. Developers kept the structure of the models
fairly simple; shallow and moderately sized models were observed in
the majority of the \chg{sampled} projects both for \bts and
\sms. From our experience in analyzing \chg{the} sampled models,
keeping the behavior model simple, regardless of the type, helps in
its understandability.
\bchg{Finally, we observed three code-reuse
  patterns in the sampled models,
  with similarities in how these code-reuse patterns were used. To
  reuse a skill (action), the main reuse pattern was inter-model
  referencing; to reuse a task (multiple actions combined in a
  sub-tree or state machine), the main reuse pattern was
  clone-and-own.}

\subsubsection*{Perspectives}
\looseness=-1 With this paper, we hope to raise the interest of two
research communities---software language engineering and software
modeling---in languages for robotics behavior. We hope that observing
the current state-of-practice can help to improve on it. We also hope
that this analysis can inspire designers of behavior-tree and
state-machine languages to revisit, or at least justify, some design
choices. In addition, it seems beneficial to take improvement points
from each other, since some of these tools are built with model-based
design concepts, and other good language design principles in mind.

Finally, we contribute a dataset of \chg{open-source} behavior models,
to inspire the community to use and further develop these languages,
associated tools, and analysis techniques. An accompanying
online appendix\,\cite{appendix:online} contains these
models, our mining and analysis scripts, and further details.


\section{Background}
\label{sec:background}

\bchg{In the robotics community, different control structures are used
  to coordinate agent behavior, including \bts, \sms, teleo-reactive
  architecture, subsumption architecture, sequential behavior
  composition, flowcharts, and decision trees
  \cite{colledanchise2017behavior, colledanchise2018behavior,
    chen2018development}. Each of these structures has its own
  advantages and disadvantages \cite{colledanchise2017behavior,
    colledanchise2018behavior}.}

\bchg{Many of these control structures are offered as DSLs (domain-specific
  languages)\,\cite{dslbook} to developers. For mobile robots,
  many of these DSLs are even end-user-oriented, offering a visual
  syntax, see our previous study\,\cite{dragule2021survey}.  There, we
  observed that most of these DSLs are still imperative at a
  relatively low level of abstraction, similar to programming
  languages. Interestingly, we observed that many of these DSLs are
  realized by cutting down a real programming language and
  implementing a visual syntax for the remaining language concepts:
  expressions, declarations, and statements (including robot-specific
  library method calls). The visual syntax is often realized using
  Scratch or Blockly, offering a simple, block-based programming
  interface. These end-user-oriented DSLs target technically skilled
  users, while the other DSLs, especially the behavior-tree and
  state-machine DSLs we study in the sequel, clearly target
  developers.}

\subsection{Behavior Trees}
	\bchg{In
  recent years, \bts have become a popular behavior-modeling
  language to specify missions and coordinate their control flow. \Bts
  have been shown to generalize multiple control architectures
  \cite{colledanchise2016behavior, colledanchise2016behavior2}. \Sms
  have been the traditional go-to model in robotics for their
  simplicity and their easy-to-understand control-switching mechanism, but
  they become hard to maintain when missions grow in complexity
  \cite{chen2018development, colledanchise2016advantages}.  \Bts are
  claimed to overcome this obstacle, due to their modularity and
  flexibility \cite{colledanchise2016advantages,
    colledanchise2016behavior, colledanchise2018behavior}. Missions
  specified using \bts can be mapped into \sms and vice versa
  \cite{colledanchise2016behavior, marzinotto2014towards}.  In
  contrast, in this paper we study their use in real robotics
  projects, which has not been done before.}

\subsubsection*{Applications of Behavior Trees}
\Bts are well-suited to express the runtime behavior of agents, which
has fueled applications in computer games and robotics.  High-profile
games, such as Halo\,\cite{isla2005gdc}, use \bts. \bchg{However, the
  implementations of \bts in gaming differ from their implementations in
  robotics. For example, the \emph{Unreal Engine 4 (UE4) Behavior
    Tree}, probably the world's most used behavior-tree DSL, emphasizes
  event-driven programming rather than time-triggered control, which
  is a major concern in robotics.}

\begin{figure}[t]
\vspace{-0.2cm}
  \begin{center}
    \includegraphics[
      width=.9\linewidth
    ]{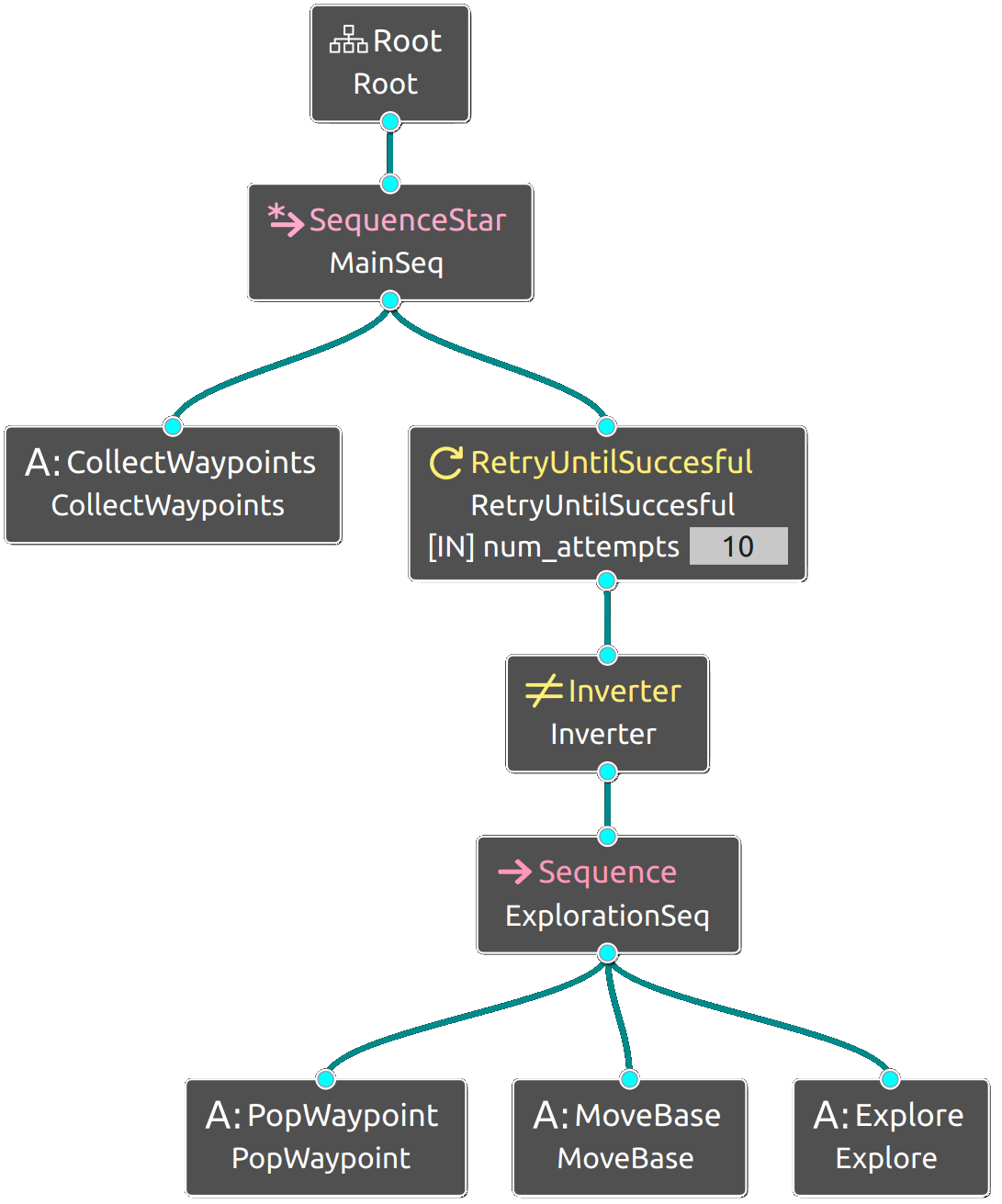}
  \end{center}
  \vspace{-3mm}
  \caption[Behavior Tree Example]{A \bt of a health and safety robot inspector from a GitHub project  \texttt{kmi-robots/hans\--ros-supervisor}, shown in the Groot editing and animation tool from \BTCPP.}%
  \label{fig:figure-1}

  \vspace{-5mm}

\end{figure}

In the robotics community, the interest in \bts is currently growing.
\bchg{Hierarchical state machines were the main task-orchestrating
  mechanism in the ROS navigation system supported by different
  languages \cite{conner2017flexible}. With the upgrade to the newer
  version of ROS, ROS~2, the main customization mechanism navigation
  was changed to \bts \cite{macenski2020marathon}. ROS~2 still
  supports hierarchical state-machine languages \cite{zutell2022ros},
  but they do not use them as the main customization
  mechanism. Further witnessing the increasing popularity of \bts, }
IROS'19, one of the key research conferences in robotics, hosted a dedicated
workshop on \bts in robotics (\textsf{\href{https://behavior-trees-iros-workshop.github.io/}{behavior-trees-iros-workshop.github.io}}).  In addition, multiple projects in
RobMoSys (\textsf{\href{https://robmosys.eu}{robmosys.eu}}), one of the leading
model-driven communities in robotics, have been launched to create a
set of best practices and tools for \bts, such as
CARVE (\textsf{\href{https://carve-robmosys.github.io/}{carve-robmosys.github.io}}) and
MOOD2Be (\textsf{\href{https://robmosys.eu/mood2be}{robmosys.eu/mood2be}}). The EU project
Co4Robots (\textsf{\href{http://www.co4robots.eu}{co4robots.eu}}) developed a
mission-specification DSL for multiple robots based on behavior-tree
concepts\,\cite{garcia2019high,garcia2020icsedemo}.  \bchg{Finally,
  \bts are used for autonomous-driving systems. Autoware
  \cite{kato2018autoware}, a leading open-source platform for
  self-driving vehicles, adopted \bts for the coordination between the
  supported pre-defined driving
  scenarios (\textsf{\href{https://autowarefoundation.github.io/autoware.universe/main/planning/behavior_path_planner/\#behavior-tree}{autowarefoundation.github.io/autoware.universe/main/pla\-nning/beha\-vior\_path\_planner/\#behavior-tree}}).
  CARLA\,\cite{Dosovitskiy17}, a high profile open-source simulator for
  autonomous driving research, uses a behavior-tree DSL, \pytrees,
  which is among the DSLs we study in the paper, to define non-ego
  vehicle behavior \cite{carlarunner}.}

\subsubsection*{Behavior-Tree Example}
\looseness=-1
\Cref{fig:figure-1} presents an example of a behavior-tree model of a
health and safety inspector robot from the Knowledge Media
Institute (\textsf{\href{http://kmi.open.ac.uk/}{kmi.open.ac.uk}}). The robot performs
an exploration sequence for an area. The main operation is placed at
the bottom, in the sub-tree under \lstinline!ExplorationSeq!: it
consists of obtaining the next waypoint, moving the mobile base to the
waypoint, and exploring the area.  If obtaining a new waypoint fails
(empty stack), the first task fails, which is converted into a success
by an \lstinline!Inverter! such that the sequence of motions has
been completed.  Otherwise, the robot keeps repeating the same
operation (next waypoint, move, explore) up to ten times, as long as
the stack is not empty.  The entire computation is placed in a loop
that alternates between obtaining new waypoints and performing the
exploration sequence (\lstinline!MainSeq!) until the success of all
children.

\begin{figure}[t]
	\small
	\begin{tabular}{
			>{\hspace{-2mm}}l
			c
			>{\hspace{-.5mm}}l
			c
			>{\hspace{-1.5mm}}l
			c}
		\raisebox{2ex}{Root}
		& \includegraphics[
		height=6.2mm,
		clip,
		trim=1.8mm 0mm 0mm 0mm]{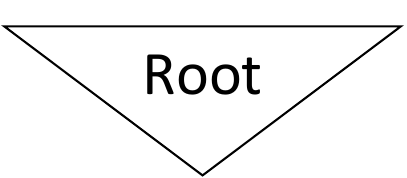}
		& \raisebox{2ex}{Seque\rlap{nce}}
		& \includegraphics[
		height=5.2mm,
		clip,
		trim=0mm 0mm 0mm 0mm]{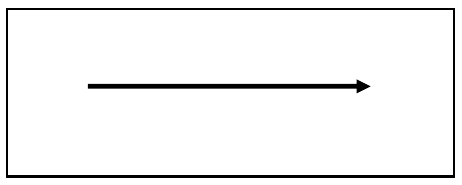}
		& \raisebox{2ex}{Select\rlap{or}}
		& \includegraphics[
		height=5.5mm,
		clip,
		trim=0mm 0mm 0mm 0mm]{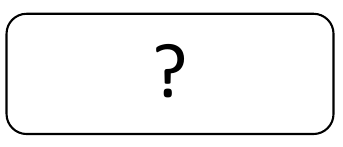}
		\\
		\raisebox{2ex}{Parallel}
		& \includegraphics[
		height=5.5mm,
		clip,
		trim=0mm 0mm 0mm 0mm]{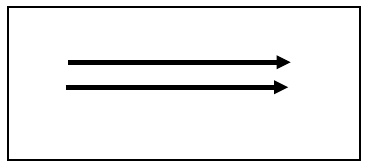}
		& \raisebox{2ex}{Decora\rlap{tor}}
		& \includegraphics[
		height=5.0mm,
		clip,
		trim=1mm 0mm 0mm 0mm]{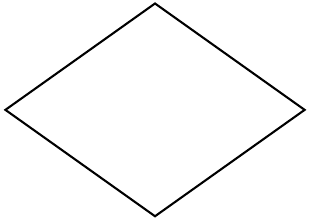}
		& \raisebox{2ex}{Acti\rlap{on}}
		& \includegraphics[
		height=5.0mm,
		clip,
		trim=0mm 0mm 0mm 0mm]{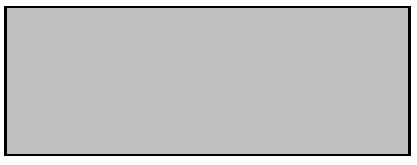}
		\\
		\raisebox{2ex}{Condit\rlap{ion}}
		& \includegraphics[
		height=5.5mm,
		clip,
		trim=0mm 0mm 0mm 0mm]{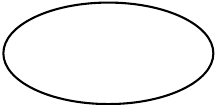}
	\end{tabular}
	\vspace{-2mm}
	\caption{Node types in behavior trees (visual syntax)}%
	\label{fig:btsyntax}
	\vspace{-4mm}
\end{figure}

\subsubsection*{Behavior-Tree Concepts}

In general, \bts can be seen as graphical models that are shaped as
directed trees, with a dedicated root node, non-leaf nodes called
\emph{control-flow nodes}, and leaf nodes called \emph{execution
  nodes}. A \bt is executed by sending signals called \emph{ticks}
from the root node, traversing the tree according to the semantics of
the control-flow nodes. Ticks are issued with a specific frequency
\cite{iovino2022survey, colledanchise2018behavior}. Upon receiving a
tick, a node executes a task, which can be a control-flow task or, if
a leaf node is ticked, some specific robotic task (a.k.a. skills). The latter
classifies into actions (e.g., \lstinline!MoveBase! in
\cref{fig:figure-1}) and conditions, which can test propositions
(e.g., whether the robot is at its base) used to control task
execution.  A ticked node returns its status to its parent: (1)
\emph{success} when a task is completed successfully, (2)
\emph{failure} when a task execution failed, and (3) \emph{running}
when a task is still under execution.

\bchg{This execution semantics is interesting and can be seen as the
  main difference to other behavior-modeling languages. As we will
  discuss for our concrete behavior-tree DSLs in the sequel, the
  current state of execution is not explicitly represented in behavior
  trees. In most other languages, such as state machines or
  flowcharts, the state advances (as prescribed by the control-flow
  elements, such as transitions) by moving the current point of
  execution (the state) forward. Consider, for instance, flowcharts,
  where one action is executed after another, controlled by
  control-flow elements between the actions. Notably, behavior trees
  are also actions controlled by control-flow elements. In other
  words, behavior trees are like many other imperative
  programs---recall also the many end-user-oriented robot-control DSLs
  studied in our previous work\,\cite{dragule2021survey}, most of
  which are subsets of imperative programming languages with a visual
  syntax represented in Scratch or Blockly. Now, in behavior trees,
  the tree is re-executed regularly in short time intervals, which
  allows to re-execute actions. Intuitively, one can use that to put
  reactive control more to the upper-left of the tree, where it gets
  regularly re-executed by every tick (determined by a control loop),
  while more deliberative skills (actions) are more to the bottom and
  right of the tree. Adding such reactive control to other languages,
  including state machines and flowcharts, is not easily possible, not
  even with hierarchical state machines. In other words, this special
  execution semantics fosters modularity and allows to unify other
  control architectures, such as the subsumption
  architecture\,\cite{colledanchise2016behavior}.}

\looseness=-1
\bchg{The benefit of using \bts lies in their ability to express task
  coordination using a small, but extensible set of control-flow
  nodes. Most behavior-tree languages offer the following control-flow
  nodes: sequence, selector, decorator, and parallel. The example in
  \cref{fig:figure-1}} illustrates two sequence nodes
(\lstinline!MainSeq! and \lstinline!ExplorationSeq!) and two decorator
nodes (\lstinline!Inverter! and
\lstinline!RetryUntilSuccesful!). Intuitively, sequence nodes tick all
children and require that they all succeed for the sequence node to
succeed, while selector nodes only require one child's success to
succeed. Decorator nodes allow more complex control flow, including
for- and while-loops. They are also extensible; developers can
implement custom decorator nodes. \bchg{Parallel nodes are a
  misnomer. They are generalizations of sequence and selector nodes,
  allowing custom policies, such as cardinalities specifying the
  minimum or maximum number of nodes that need to succeed. In the
  literature, there is no clear agreement if parallel nodes execute
  child nodes simultaneously, as the name suggests, or concurrently
  \cite{iovino2022survey, colledanchise2019analysis,
    millington2009artificial, Mcquillan2015,
    colledanchise2018behavior, colledanchise2017behavior}.}

\bchg{The visual presentation of the main node types, as illustrated
  in studies of \bts in robotics and
  games\,\cite{colledanchise2018behavior,iovino2022survey,
    millington2009artificial}, is summarized in
  \cref{fig:btsyntax}. Our example in \cref{fig:figure-1} shows the
  visual syntax used in a popular behavior-tree DSL called \BTCPP,
  which is among the DSLs we study in the remainder of the
  paper. Although the shapes of the nodes might differ between
  publications and actual language implementations, the inner symbol
  of each node-type is usually the same; e.g., the selector node
  always uses the symbol ``?''. }

\subsection{State Machines}
State machines are probably the most popular and well-researched
language for behavior specification of systems. They provide the basis
not only for modeling, but also for analysis (e.g., for model
checking) or synthesis of software controlling real systems. There is
a wealth of literature on state machines in non-robotics contexts. As
such, we do not discuss the applications in detail here, as opposed to
behavior trees above, but to further illustrate behavior trees in the
well-known state-machine syntax, we provide an example and then recap
the main state-machine concepts again.

\subsubsection*{State-Machine Example}

\begin{figure}[t]
	\centering
        \includegraphics[width=\linewidth]{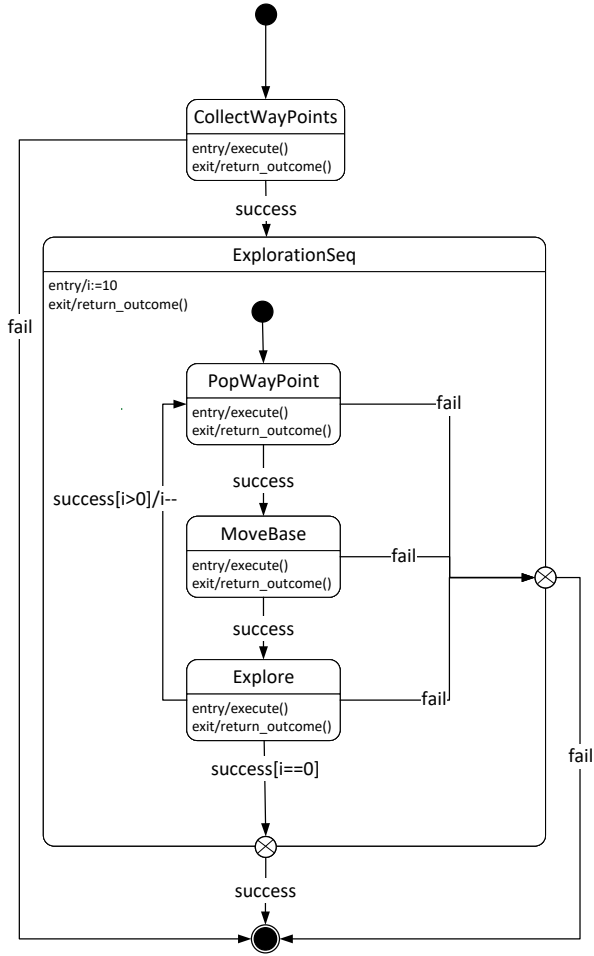}
		\caption[State machine Example]{A \sm  representing the same mission as in \cref{fig:figure-1}, using UML syntax.}%
	\label{fig:smexm}
	\vspace{-4mm}
\end{figure}

\Cref{fig:smexm} shows the same mission as in
  \cref{fig:figure-1} for a health and safety inspector robot,
  represented as a state machine, using the notation of the Unified
  Modeling Language (UML) (see \cref{fig:smsyntax} for the UML
  syntax).
	Since state machine DSLs offer a rich notation for modeling behavior, there is no single mapping from behavior trees to state machines. We modeled the actions and the control flow in \cref{fig:figure-1} as follows.
	
	States represent some situation that the modeled system is in for some period of time. From the actions in our behavior tree, we model the collection of way points and the exploration sequence as states, where the latter consists of the states \lstinline!MoveBase! and \lstinline!Explore!. These states have the respective actions from the behavior tree, executed on state entry or state exit. In contrast, the action \lstinline!PopWaypoint! is rather atomic, so we do not represent it as a state, but execute it when the state \lstinline!MoveBase! is entered.

\looseness=-1	
The \lstinline!ExplorationSeq! of the \bt model is placed
  in a nested, also called composite, state
  \lstinline!Exploration_seq!. At the entry of
  \lstinline!Exploration_seq!, a variable \lstinline!i! is set to the
  desired number of attempts (here, ten). For each waypoint, the \lstinline!MoveBase! component is activated and the model waits in the state until the sensors and the localization component detect that a waypoint has been reached (or a timeout occurs). The model exits from
  \lstinline!Exploration_seq! if exploration has achieved its goals
  (the ten attempts are finished; the triggering event is
  \lstinline!success! and the guard's
  condition is \lstinline!i==0!).  If the event \lstinline!success!
  is triggered from \lstinline!Explore! and \lstinline!i>0!, the
  counter \lstinline!i! is decreased at the transition from
  \lstinline!Explore! back to \lstinline!MoveBase! (which is an alternative to having actions on state entry or exit in UML).

\looseness=-1
Finally, transitions are triggered by events, which are signals coming from the actions or from the environment. In behavior trees, the control-flow is determined by the actions' outcome, which is either success or fail. In our state machine, we also used \lstinline!success! or \lstinline!fail! as the events originating from action results for some transitions. For others, we used more domain-specific events. For instance, \lstinline!timeout! is an external event from the system, which ends the state \lstinline!MoveBase! and then the whole system; alternatively, the event \lstinline!waypoint reached! moves the system to the state \lstinline!Explore!, where the exploration either continues with the next waypoint or also ends the whole system. If no waypoint is left (indicated in the guard on one of the state's leaving transitions), then \lstinline!popwaypoint()! would fail. Since in the behavior tree, this would be handled by the dedicated action, and we have no specialized state for it, we utilize a method \lstinline!waypointsleft()! to avoid going into \lstinline!MoveBase! again.

\subsubsection*{State-Machine Concepts}

\begin{figure}[t]

	\small

	\begin{tabular}{
			>{\hspace{-2mm}}l
			l
			>{\hspace{-4mm}}l
			>{\hspace{-2mm}}c
		}

		\raisebox{2ex}{Initial state}
		& \includegraphics[
		height=6.0mm,
		clip,
		trim=2mm 1mm 3mm 2mm]{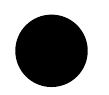}

		&\raisebox{2ex}{Final \rlap{state}}
		& \includegraphics[
		height=8.0mm,
		clip,
		trim=3mm 3mm 3mm 3mm]{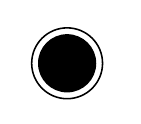}

		\\
		\raisebox{6ex}{State}
		&  \includegraphics[
		height=15mm,
		clip,
		trim=1mm 2mm 2mm 1mm]{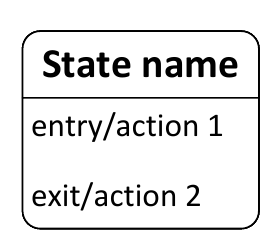}

		& 	
		\raisebox{8ex}{\multirow{2}{15mm}{Composite state with nested-states}}
		& \includegraphics[
		height=15mm,
		clip,
		trim=1mm 2mm 1mm 2mm]{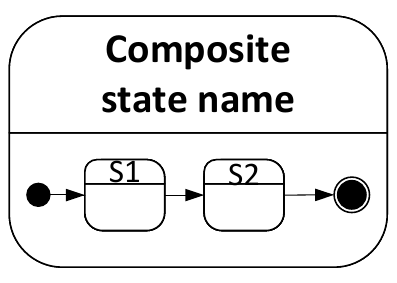}

		\\

		{Trans\rlap{ition}}
		& \includegraphics[
		height=5mm,
		clip,
		trim=0mm 4mm 2mm 2mm]{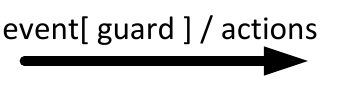}

	\end{tabular}

	\caption{Node (state) types and transitions (visual syntax) for \sms, according to UML.}
	\label{fig:smsyntax}
	\vspace{-3mm}
\end{figure}

\Sms and hierarchical state machines have been used for decades in
different domains with slight variations in syntax and semantics
\cite{crane2007uml, von1994comparison}.  While there exist multiple
syntaxes with minor variations, the UML's state diagram, a graphical
representation of \sms in UML, might be the most common visual syntax,
especially in the software modeling community (see
\cref{fig:smsyntax}).

\looseness=-1
\bchg{A state-machine model is a directed graph of three primary
  elements: states, transitions and actions. In the UML semantics
  \cite{uml2017, DOUGLASS2011257, crane2007uml}, states represent
  patterns of behavior. The behaviors executed by the robot are
  represented as actions. From the perspective of computational
  execution semantics, UML's \sms support both the execution semantics
  of Mealy and Moore machines \cite{samek2009crash,
    kubatova2013petri}. When an action is associated to a transition
  (action-on-transition), similar to Mealy machines, the action is
  executed at the transition. When the action is associated to a state
  (action-on-state), similar to Moore machines \cite{Hajji2022}, the
  action is executed at the state entry or exit.  Most state-machine
  implementations are event-driven.}

Hierarchical state machines, also known as statecharts
\cite{harel1987statecharts}, are an extension of \sms that allow
nesting (hierarchy) and concurrency \cite{harel1987statecharts,
  luttgen2000compositional}. Thus, hierarchical state machines are
similar to \sms but can additionally separate behaviors into
sub-states, or composite states, thereby increasing the modularity of
the model and decreasing the number of transitions needed.

\bchg{Just like \bts in robotics, \sms provide graphical models to
  represent the behavior of a robotic agent.  Colledanchise and
  \"Ogren~\cite{colledanchise2016behavior} have shown how \bts
  generalize \sms in robotics. In the rest of the paper, we focus on
  implementations in robotics; the robotic implementations of \sms
  will differ slightly from the UML notation.}

\section{Methodology}
\label{sec:methodology}
\looseness=-1 We now describe our methodology to (RQ1) identify relevant DSLs and the concepts offered by \bts and \sms, to (RQ2) analyze their implementations, and to (RQ3) identify and analyze open-source robotics projects using these \chg{DSLs}.

\subsection{Identifying Languages and Concepts (RQ1)}
\label{subsec:conceptsmeth}
\looseness=-1
\bchg{To identify behavior-tree DSLs, we searched on GitHub for projects with behavior trees in robotics using different search terms, including ``behavior tree robotics'' and ``behavior tree robot.'' The search returned projects using different behavior-tree DSLs. We identified the imported DSLs in the projects. We focused on DSLs in Python and C++, the most used programming languages in robotics. We also looked at the behavior-tree literature in robotics to identify DSLs.}

\looseness=-1
A similar search for state-machine DSLs yielded hundreds of results on GitHub. Therefore, we decided to use the ROS wiki (\textsf{\href{http://wiki.ros.org/}{wiki.ros.org}}) to find state-machine \chg{DSLs}. The wiki is commonly used by developers to publish their open-source languages and tools for developing robotics applications. By searching it instead of GitHub or Google, we ensured that the identified languages support ROS, so we excluded non-robotics projects in later steps. 

To ensure the relevance of the identified \chg{DSLs (offered as libraries)} for real-world robotics applications for both \bts and \sms, we focused on maintained libraries and for that we applied the following \emph{exclusion criteria}: (1) lack of documentation, (2) out-dated libraries not maintained anymore (last commit older than 2019), (3) no ROS support (this was checked specifically for the identified \bt \chg{DSLs} since we collected them through GitHub) and (4) no mined projects. 

Thereafter, to understand the key characteristics and modeling concepts offered by the included \chg{DSLs}, we identified their main language constructs and studied the relation between them. For our comparison, we collected behavior-tree concepts by an exploratory literature search\,\cite{millington2009artificial,colledanchise2018behavior,colledanchise2017behavior} and snowballing. For the identified \chg{libraries}, we inspected their documentation\,\cite{pytreedoc,btscppdoc} and wrote scripts \chg{that execute small tests} to understand better their semantics.  A similar approach was followed to collect state-machine concepts from the literature\,\cite{bohren2010smach, schillinger2016flexbe, schillinger2015approach, kohlbrecher2016comprehensive} and the documentation of the identified libraries\,\cite{smachdoc,flexbedoc}. In this process, we focused on behavior-tree concepts and whether \sms offer direct support for similar concepts or they need to be expressed indirectly. Our analysis was iterative, to ensure a proper reflection of the concepts in the different \chg{DSLs}. \chg{We scoped the comparison to concepts offered by the included DSLs that met our criteria}.

\subsection{Language Implementations (RQ2)}
\label{subsec: implementationsmeth}
After identifying relevant libraries that support state-machine and behavior-tree modeling languages and analyzing their syntax and semantics, we wanted to understand the implementation design of the libraries. We built our findings from inspecting different sources. Specifically, we inspected the implementations of the libraries on GitHub, documentations \cite{btscppdoc, pytreerosdoc, pytreedoc, flexbedoc, smachdoc}, tutorials \cite{btscpptutorial, pytreerostutorial, flexbetutorial, smachtutorial}, and related publications \cite{mood2be, colledanchise2021implementation, schillinger2016flexbe, romay2017collaborative, der2015approach, bohren2010smach, kohlbrecher2016comprehensive}. 

We focused on the language design of the libraries and the concrete syntax offered. We went through the tutorials of the libraries and related publications to reflect on the dynamicity of the libraries. By dynamicity, we refer to the ability of runtime modification of models. Finally, we examined the models of concurrency used in the libraries by examining their implementations on GitHub and the documentation.

\subsection{Identifying Languages Projects and Analysis (RQ3)}
\label{subsec: projectanalysismeth}
To understand the usage of the identified \chg{DSLs} in robotics projects: (1) we mined GitHub for open-source repositories using these languages in their implementation of robotics missions, then (2) we compared the popularity of the \chg{DSLs} in open-source projects over time and analyzed a sample of these projects in terms of the usage of the languages concepts and the structure of the models, and finally (3) we observed how the \chg{DSLs} users have carried out \chg{code-reuse} in practice. \chg{In the sequel}, we describe these steps in detail.

\begin{figure}[t]
	\begin{center}
		\includegraphics[
		width=.7\linewidth
		]{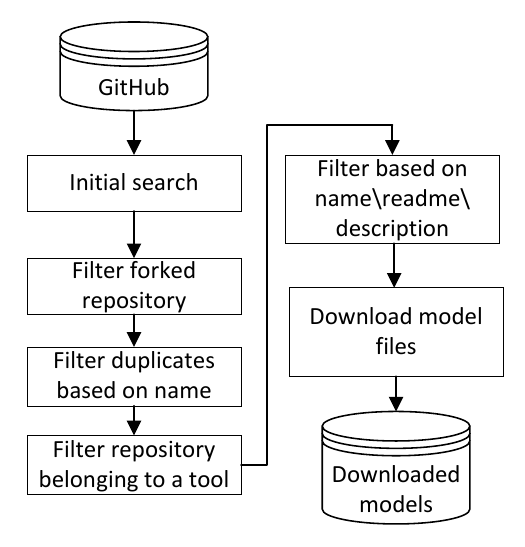}
	\end{center}
	\vspace{-.5cm}
	\caption{Filtering steps to identify relevant repositories.}%
	\label{fig:filterrepos}
	
	\vspace{-3mm}
	
\end{figure}
\subsubsection*{Mining GitHub}
\label{subsec: minegithubmeth}

Open-source projects have been used in software-engineering and robotics research communities to understand real-world applications. Different open-source platforms, such as GitHub, provide great opportunities for researchers to identify the state-of-practice in robotics and software engineering\,\cite{almarzouq2020mining, malavolta2021mining, robles2017extensive}. Motivated by our previous work\,\cite{ghzouli2020behavior} and other researchers who mined GitHub, we used GitHub as a source for mining open-source robotics projects. 

For the identified behavior-tree and state-machine DSLs, we investigated how they are used in the source code of robotics projects. In \BTCPP, the term \emph{main\_tree\_to\_execute} refers to the entry point tree in the XML source code, while the term \emph{py\_trees\_ros, smach\_ros}, and \emph{flexbe\_core} are used to import the languages \pytreesros, \smach and \flexbe, respectively. These terms must be used in the source code of the targeted languages. We created a Python script to mine GitHub repositories for those terms with a simple text-match in source code, using GitHub's code search API (\textsf{\href{https://github.com/PyGithub/PyGithub}{github.com/PyGithub/PyGithub}}).

Next, we wanted to filter-out projects belonging to a course or tutorial, to better reflect on actual usage of the libraries in robotics. Inspired by the work of Malavolta et al.\,\cite{almarzouq2020mining} in providing mining guidelines for robotics software, we used a similar filtering mechanism with minor changes to adopt it to our goal. \Cref{fig:filterrepos} shows our filtering mechanism. After mining GitHub for open-source projects (step 1), we excluded forked repositories using a Python script (step 2). By filtering forked repositories, we excluded duplicate models. After a quick inspection, we noticed that there are duplicate repositories that were not forked, instead they might have been cloned and re-uploaded to the user's GitHub account. To filter them out, we extracted the repositories with the same project name belonging to different GitHub users using a Python script, inspected them manually to ensure they are duplicates, and deleted them from our lists (step 3). In our previous inspection, we noticed that repositories with the keyword (tool) in their name often belonged to software tools where the targeted libraries are requirements for their functionalities. Since our goal is collecting the implementations of robotics projects rather than supportive tools, we excluded repositories with the keyword (tool) in their name using a Python script (step 4). We also excluded the following organizations as users, since they either correspond to  tutorials repositories of the library creators, or they are known tools using the identified languages as requirements (BehaviorTree, splintered-reality, team-vigir, ros-planning, ros-infrastructure, carla-simulator). Finally, we used different Python scripts to inspect the name, readme, and description of the repositories, and exclude those with the following keywords:  (assignment, course, tutorial, introduction) (step 5). This ensured the exclusion of projects belonging to (1) an assignment or (2) a course. By the end of this step, we had a list of relevant repositories that match our criteria. 

\looseness=-1 Our next step was to download the files that contain the models from these repositories (step 6). Through our analysis while extracting concepts in \Cref{sec:concepts}, we identified specific terms used in each \chg{DSL} when constructing the behavior model. In \BTCPP, the term \emph{main\_tree\_to\_execute} refers to the entry point when constructing the tree. In \pytreesros, \emph{add\_child} is usually used when adding nodes to the tree. Finally, \emph{StateMachine.add} and \emph{OperatableStateMachine.add} are used in \smach and \flexbe, respectively, to add states. Using these terms, we matched the files in the GitHub repositories that contain them and downloaded them. For all the previous steps, we used Python regular expressions, Python's Requests API, and the code-search API from GitHub. \chg{All code} can be found in our online appendix\,\cite{appendix:online}.

\subsubsection*{Analyzing Models}

\looseness=-1 To understand the usage of \bts and \sms in \chg{open-source robotics applications} (RQ3), we analyzed the mined projects from two different perspectives: the popularity of the \chg{DSLs} over time and the structure of models. As a start, to understand the popularity of \chg{the DSLs}, we extracted the creation and last commit dates for the repositories to plot the number of active projects per year. We assume a project is active in the duration between its creation date and last commit. We used all mined projects before filtering and only excluded those belonging to the \chg{DSLs}' organizations/creators, which are the following: BehaviorTree, team-vigir, pschillinger, FlexBE, splintered-reality, ros, ros-visualization. The collected projects were mined until 31/12/2021 to capture activity until the end of 2021.

Moving to the analysis of the models, we were interested in understanding the usage of \bts and \sms in \chg{real robotics} projects. Consequently, we used the list of relevant repositories from the filtering step and their downloaded models. At the time of the analysis of the behavior-tree models, we got 75 models. Although an updated mining at the beginning of 2022 to cover the projects until the end of 2021 yielded new models and projects, we decided to keep the same sample, since it covers different sizes of models and domains. For state-machine models, we randomly sampled 75 models to match the number of behavior-tree models to have comparable results.  The state-machine mining yielded thousands of models; consequently, we randomly sampled projects. We defined two project pools according to the model size (defined as number of nodes, see below for metric details), one for normal size [2--6] and another for large size [7--66], then we randomly sampled from each pool a number of projects using the sampling API (\textsf{\href{https://pandas.pydata.org/docs/reference/api/pandas.DataFrame.sample.html}{pandas.pydata.org/docs/reference/api/pandas.DataFrame.sa\-mple.html}} from the Python DataFrame library. The model size range for each pool was decided based on the data distribution. The total sample size corresponded to the number of models belonging to behavior-tree projects meeting our repository inclusion criteria (75 models).

We calculated metrics that capture the core structural aspects of the models and report on the usage of behavior-tree and state-machine concepts. Since we are analyzing two different types of structures, directed cyclic graphs in \sms and directed acyclic trees in \bts, it is challenging to use similar metrics. Thus, we calculated some common metrics between the two architectures and other metrics only for one architecture. To distinguish between them, we are using the abbreviations SM and BT to indicate which architecture metric belongs to. The metrics are:
\begin{itemize}
	
	\item Model size (BT.size, SM.size): the total number of nodes excluding the root node in \bts and the total number of states in \sms. 
	
	\item The tree depth (BT.depth): the number of edges from the root node to the deepest node of the tree\,\cite{chidamber1994metrics}. Considering the example in \cref{fig:figure-1}, the \bt model has a BT.depth of five. 
	
	\item Average branching factor (BT.ABF): the average number of children of each node. Considering the example in \cref{fig:figure-1}, the behavior-tree model has a BT.ABF of \num{1.6}.
	
	\item Nesting level (SM.nesting): the composition or nested hierarchy level induced by counting the number of levels within the state-machine compositions\,\cite{berger2014towards, cruz2007using, cruz2005metrics}. We consider the main state machine as level one. In the example in \cref{fig:smexmsmach}, this state-machine model has a nesting level of three; where the main container \lstinline!Inspect_SM! is level 1, the nested container \lstinline!Iterator_10_attempts! is level 2 and the \lstinline!Exploration_seq! is level 3.
	
	\item Node type percentage (N.pct): the frequency of a node type with respect to the total number of nodes. This metric captures the usage of the different composite node types in \bts and the different container constructs in \sms. 
	
\end{itemize}

\looseness=-1 The extraction of these metrics relies heavily on the implementations of the \chg{DSLs}, so we used different extraction methods for each language. For behavior-tree \chg{DSLs}, we inspected the code of both the tutorials of the libraries and a randomly sampled subset of mined models to understand how users implement them. We noticed that users follow different implementation styles, and that the models tend to be deeply intertwined with the rest of the code. 
While some metrics could be calculated automatically, for others, the models needed to be manually extracted. 
To calculate BT.size and N.pct, we extracted a function name for each node type based on the documentation of the libraries, then used a Python script to count the number of text matches. For leaf nodes, no automatic counting was possible since the libraries do not impose a specific implementation structure. We counted and calculated the percentage of leaf nodes, BT.depth and BT.ABF manually from the manually extracted models.
To extract behavior-tree models, we were able to use a visual editor shipped with one of the identified libraries (Groot for \BTCPP, explained shortly), where the behavior-tree language is realized as an external DSL. The other identified library  (\pytreesros, explained shortly) constituted an internal DSL, where we needed to manually extract the model from the source code by identifying the respective library API calls used to construct the model. We considered every tree with a root node as a behavior-tree model. 

\looseness=-1
Moving to state-machine \chg{DSLs}, we also checked the code of tutorials of the libraries and a random sample of mined models. An implementation pattern was clear for each library, hence an automatic extraction of the metrics was possible. Using Comby (\textsf{\href{https://github.com/comby-tools/comby}{github.com/comby-tools}}), a parser that detects syntax in code based on user-supplied patterns\,\cite{van2019lightweight}, we defined a syntax pattern to match the start and end of a state-machine container to isolate it from the rest of the code. Then, we wrote a Python script to extract the state-machine model from the rest of the code. After extracting the model, another script was used to count the number of states and composite states to calculate SM.size and N.pct. Similar to the composite nodes count in \bts , a text match for the function name was used to count the different container types. Finally, we wrote a script for visualizing the extracted model to facilitate model inspection using SMCat (\textsf{\href{https://github.com/sverweij/state-machine-cat}{github.com/sverweij/state-machine-cat}}). We could not use the viewers provided by \flexbe and \smach, because the models are intertwined with the code, and each project has different dependency requirements to run it. It was easier to isolate the model from the projects and visualize it using an external viewer. We wrote a Python script to transfer the extracted model code into the SMCat syntax and generate SMCat files. SMCat was executed on those files to generate SVG format files that visualize a corresponding state-machine model. Extracting the SM.nesting required manual work. Using the extracted visualized models, we went through all models and counted the nesting levels. All scripts and SVG files can be found in our online appendix \cite{appendix:online}.

\subsubsection*{Analyzing Reuse}
\looseness=-1 The final aspect we analyzed is reuse, because it is one of the major issues in robotics software engineering\,\cite{garcia.ea:2023:robotvar, garcia2019robotics,garcia2020robotics, brugali2010component, brugali2009software, nesnas2006claraty}.
\bchg{Reuse mechanisms are important to scale and sustain the use of \bts and \sms in practice. We analyzed the current state-of-practice for reuse in the studied models.}
By reuse we refer to reusing the code of a robotic skill (also known as action) instead of writing a new skill from scratch, or reusing the code of a repeated task (composed of different skills) in the same model or in a different one. 

We observed reuse in the sampled projects using a mixture of visual and code-level inspection to detect any reused skill or task. We inspected the reuse of a skill or task in the model and across the different models of a project. A task is usually expressed as a sub-tree in \bts and as a composite state in \sms. A skill is expressed as a leaf node in \bts and as an action associated to a state in \sms. By first checking the model visualization, we could detect task-level and/or skill-level reuse in a model. We also relied in the task-level reuse on similar model structure in case of a slightly different combination of skills with a resembling structure. A deeper inspection of the code-level implementation of models and skills for the reused tasks and skills follows to understand what type of reuse mechanism is used.


\begin{table}[t]
	\renewcommand{\arraystretch}{1.15}
	\caption{Behavior-tree and state-machine \chg{DSLs} identified. We analyzed those in bold}%
	\label{tbl:btsmlanguages}
	\begin{tabularx}{\linewidth}{%
			>{\small\raggedright}p{3.1cm}  
			>{\small}p{1.6cm}
			>{\small}p{0.5cm}
			>{\small}p{5mm}
			>{\small}p{14mm}}

		& \textsf{target}
		&
		&
		& \textsf{last} \\
		 \textsf{name}
		& \textsf{language}
		& \textsf{ROS}
		& \textsf{doc.}
		& \textsf{update}\\
		\midrule
		\hspace{-.1cm}{\bfseries{\BTCPP}} \newline
		\rlap{\scriptsize \MYhref{https://github.com/BehaviorTree/BehaviorTree.CPP}{github.com/BehaviorTree/BehaviorTree.CPP}}
		& \bfseries{C++}
		& \bfseries{yes}
		& \bfseries{\text{\cite{btscppdoc}}}
		& \bfseries{2021/05}
		\\[+.5cm]
		\hspace{-1mm}{\bfseries{\pytrees}} \newline
		\rlap{\scriptsize \MYhref{https://github.com/splintered-reality/py\_trees}{github.com/splintered-reality/py\_trees}}
		& \bfseries{Python}
		& \bfseries{no}
		& \bfseries{\text{\cite{pytreedoc}}}
		& \bfseries{2021/05} 
		\\[+.5cm]
		\hspace{-1mm}{\bfseries{\pytreesros}}\newline
		\rlap{\scriptsize \MYhref{https://github.com/splintered-reality/py\_trees\_ros}{github.com/splintered-reality/py\_trees\_ros}}
		& \bfseries{Python}
		& \bfseries{yes}
		& \bfseries{\text{\cite{pytreerosdoc}}}
		& \bfseries{2021/05} 
		\\[+.5cm]
		\hspace{-1mm}BT++\newline
		\rlap{\scriptsize \MYhref{https://github.com/miccol/ROS-Behavior-Tree}{github.com/miccol/ROS-Behavior-Tree}}
		& C++
		& yes
		& \text{\cite{rosbtdoc}}
		& 2018/10 
		\\[+.5cm]
		\hspace{-1mm}SkiROS2\newline
		\rlap{\scriptsize \MYhref{https://github.com/RVMI/skiros2}{github.com/RVMI/skiros2}}
		& Python
		& yes
		& \text{\cite{skiros2}}
		& 2020/11
		\\[+.5cm]
		\hspace{-1mm}pi\_trees\newline
		\rlap{\scriptsize \MYhref{https://github.com/pirobot/pi\_trees}{github.com/pirobot/pi\_trees}}
		& Python
		& yes
		& n/a
		& 2017/10
		\\[+.5cm]
		\hspace{-1mm}Beetree\newline
		\rlap{\scriptsize \MYhref{https://github.com/futureneer/beetree}{github.com/futureneer/beetree}}
		& Python
		& yes
		& n/a
		& 2016/03
		\\[+.5cm]
		\midrule
		\hspace{-1mm}{\bfseries{SMACH}}\newline
		\rlap{\scriptsize \MYhref{https://github.com/ros/executive\_smach}{github.com/ros/executive\_smach}}
		& \bfseries{Python}
		& \bfseries{yes}
		& \bfseries{\text{\cite{smachdoc}}}
		& \bfseries{2020/05}
		\\[+.5cm]
		\hspace{-1mm}{\bfseries{FlexBe}}\newline
		\rlap{\scriptsize \MYhref{https://github.com/team-vigir/flexbe\_behavior\_engine}{github.com/team-vigir/flexbe\_behavior\_engine}}
		& \bfseries{Python}
		& \bfseries{yes}
		& \bfseries{\text{\cite{flexbedoc}}}
		& \bfseries{2020/12}
		\\[+.5cm]
		\hspace{-1mm}SMACC\newline
		\rlap{\scriptsize \MYhref{https://github.com/reelrbtx/SMACC}{github.com/reelrbtx/SMACC}}
		& C++
		& yes
		& \text{\cite{smaccdoc}}
		& 2021/05
		\\[+.5cm]
		\hspace{-1mm}RSM\newline
		\rlap{\scriptsize \MYhref{https://github.com/MarcoStb1993/robot\_statemachine}{github.com/MarcoStb1993/robot\_statemachine}}
		& C++
		& yes
		& \text{\cite{steinbrink2020state}}
		& 2020/03
		\\[+.5cm]
		\hspace{-1mm}Decision making\newline
		\rlap{\scriptsize \MYhref{https://github.com/cogniteam/decision\_making}{github.com/cogniteam/decision\_making}}
		& C++
		& yes
		& n/a
		& 2016/07
		\\
		\bottomrule
	\end{tabularx}
	\vspace{-3mm}
\end{table}

\begin{table*}
	\caption{\label{tbl:concepts} Key concepts in our behavior-tree languages compared to our state-machine languages}
	\renewcommand{\arraystretch}{1.1}

	\begin{tabularx}{\linewidth}{
			>{\small\raggedright}p{20mm}
			>{\small}X
			>{\small}X}

		\textsf{behavior-tree}
		& \textsf{behavior-tree languages}
		& \textsf{state-machine languages}
		\\
		\textsf{concept/a\rlap{spect}}
		& \textsf{\pytrees, \pytreesros, \BTCPP}
		& \textsf{\smach, \flexbe}
		\\
		\midrule

		programming model
		& Synchronized and asynchronized are supported, time-triggered, activity-based. Reactive programming can be implemented to an extent using the tick concept and re-ordering of sub-trees.
		& Asynchronous, event-triggered, reactive
		\\\midrule

		\textbf{simple nodes}
		& Execute \textbf{actions} (arbitrary commands, both instantaneous and long-lasting) or evaluate \textbf{conditions} (value translated to Success/Failure).
		& Basic state with associated action/conditions
		\\

		exit status
		& Each node reports success, failure, or an in-operation state (``running'') each time it is triggered. Status report causes the computation (the traversal) to advance to the next node.\looseness=-1
		& Each state reports its status (also known as outcome). Outcomes are user defined. Execute function checks periodically for outcome
		\\\midrule

		\textbf{composite nodes}
		& Define hierarchical traversal, the control-flow for each epoch (tick).  Sequentially composed. Nodes may start concurrent code though.\looseness=-1
		& Similar to container concept. Allow state nesting and control-flow constructs.
		\\

		root
		& Serves as entry point for every traversal.  Has exactly one child node. Root node is re-entered at every epoch.
		& Initial state (first state added to the container).
		\\

		sequence
		& Trigger children in a sequence until the first failure. If no failure return success, otherwise fail.
		& Supported using a predefined container.
		\\

		selector
		& Trigger children in a sequence until the first success. If no success return failure, otherwise succeed.
		& No direct support, could use container to extend.
		\\

		parallel
		& Generalize sequence/selector with a policy parameter. Several polices available, e.g., meeting a minimum number succeeding children.\looseness=-1
		& No direct support, could use guards to extend.
		\\

		goto (jumps)
		& \textls[-5]{No general jump construct, the computation always traverses the tree\rlap{.}}
		& Supported.
		\\\midrule

		\textbf{decorators}
		& \\

		inverter
		& Invert the Success/Failure status of the child
		& No direct support, could use container to extend.
		\\

		succeed
		& Return success ignores the status returned by the child
		& No direct support, could use container to extend.
		\\

		repeat
		& Trigger the child node a set number of times, then succeed.  Fail if the child fails.
		& Could use the \emph{Iterator container} in \smach as repeat-until loop where the desired outcome for breaking the loop is user-defined.
		\\

		retry
		& Run the child node and retry it immediately if it fails for a maximum number of times, otherwise succeed.
		& Similar to repeat concept, \emph{Iterator container} can be modified and used.
		\\ \midrule

		dynamicity
		& Runtime modifications of model (node re-ordering) possible due to the dynamic nature of the implementation.
		& Runtime modifications of model is possible (adding and removing states, changing state instantiation, Re-ordering states).
		\\

		openness
		& New nodes and operators implemented by users as needed.
		& States and new control-flow is possible using the container concept.
		\\
		concurrency
		& interleaving and co-routines; declared via sequence, selector, and parallel nodes.
		& interleaving; declared via concurrency container; states are executed sequentially but not parallel due to using Python threading.
		\\
		\bottomrule

	\end{tabularx}
	\vspace{-3mm}
\end{table*}

\section{Language Concepts (RQ1)}
\label{sec:concepts}

\bchg{The search we conducted yielded 12 DSLs for implementing
  behavior-tree and state-machine models in robotics applications using
  ROS.}  The upper part of \cref{tbl:btsmlanguages} lists the
identified behavior-tree \chg{DSLs}. We focus on analyzing the
\chg{DSLs} in the first three \chg{rows}, set in bold font, for the
following reasons. Among the \chg{DSLs} relevant for robotics, these
three were actively developed when we checked (2021/06/09). Together,
they support ROS systems implemented in Python and C++, the two most
popular programming languages in the robotics community. The latest
version of ROS, ROS~2 \cite{maruyama2016exploring}, is supported by
the included DSLs. \pytrees, the main behavior-tree \chg{DSL} in the
Python community, does not directly target ROS, but robotics in
general. A popular extension, \pytreesros, provides bindings for
ROS. Since \pytrees and \pytreesros are similar, with the only
difference of ROS packaging, we decided to include \pytrees in the
language analysis even though it does not support ROS
directly. \chg{Although we analyzed both \pytrees and \pytreesros, in the
  remainder of this section we only refer to \pytreesros in the
  analysis for brevity, since our findings apply to both of them. }

\looseness=-1
We decided to discard the remaining \chg{DSLs} from our
analysis. BT++ is now obsolete, superseded by \BTCPP after the
developer of BT++ joined the latter as a contributor. SkiROS2 is a software platform for robotic task-level programming that uses design concepts from MDE and \bts as an execution engine. SkiROS2 is a new version of SkiROS \cite{rovida2017skiros, rovida2015design}, which is now obsolete. SkiROS2 does not have open-source projects, so we needed to discard it from further analysis. Beetree and pi\_trees are inactive experiments, now abandoned.

The lower part of \Cref{tbl:btsmlanguages} lists the state-machine
\chg{DSLs} we identified. We analyzed the first two (in bold font) and
discarded the other three, since they match our exclusion criteria
(cf. \cref{subsec:conceptsmeth}). SMACC is an actively maintained
\chg{DSL} that supports ROS systems implemented in C++. Robot
Statemachine (RSM) is another C++ \chg{DSL} with ROS support that also
offers a GUI. Due to a lack of open-source projects on GitHub using
SMACC and RSM, we needed to discard them. Decision making supports ROS
systems in C++, but is no longer maintained. The two included
\chg{DSLs} are actively maintained when we checked (2021/06/09), have
good documentation, support ROS systems in Python, and have multiple
available projects on GitHub for mining. \chg{\flexbe supports ROS2,
  while \smach does not \cite{zutell2022ros}.}

We now reflect on the language concepts offered by the \chg{included}
behavior-tree and state-machine \chg{DSLs}, both at the syntactic and
semantic levels.  The language concepts offered by our behavior-tree
and state-machine \chg{DSLs} are summarized and compared in
\cref{tbl:concepts}. The left-most column names concepts that are
pertinent to the behavior-tree \chg{DSLs}, either due to inclusion or
a striking exclusion from the behavior-tree \chg{DSLs}. The last
column comments briefly on how the respective concepts are handled in
the state-machine \chg{DSLs}.  The remainder of this section discusses
the details, first for \bts and then for \sms.

\begin{figure}[t]
	\centering
	\includegraphics[width=\columnwidth]{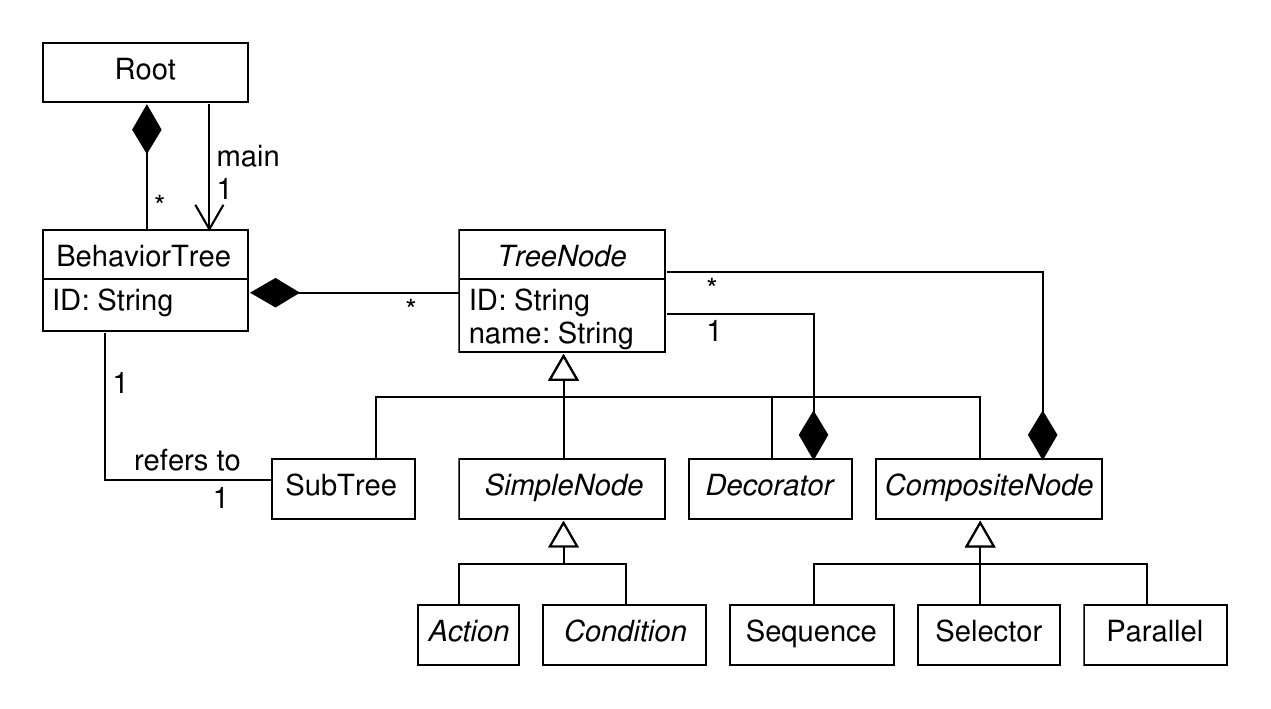} 
	\caption[MM]{\label{fig:metamodel} A meta-model for \BTCPP (reverse-engineered from its XML format)}%
	\vspace{-2mm}
\end{figure}

\subsection{Concepts and Semantics in Behavior-Tree DSLs}
\label{subsec:BTsematicsyntax}

\bchg{The following discussion is based on a broad description of behavior-tree
languages extracted from the available literature and documentation of
\pytreesros and
\BTCPP\,\cite{pytreedoc,btscppdoc}. \Cref{tbl:concepts-vs-langs}
presents the classes of the basic behavior-tree concepts in the analyzed DSLs.}

The variant of \bts used in the analyzed \chg{DSLs} is predominantly a
\emph{time-triggered} \emph{activity-based} behavioral modeling
language\chg{, unlike the implementation of \bts in gaming (see
  \cref{sec:background})}. The computation consists of activities that
have duration and the main control loop triggers the entire model at
(typically) fixed intervals of time like a circuit.  Every tick (or
epoch) triggers a traversal of the entire tree, with diversions
introduced by the various types of nodes.  The traversal can start new
activities, evaluate conditions, access state, and execute basic
actions for side effects. Reactive programming seems not to be
supported first-class, despite reappearing statements to the
contrary,\footnote{For example, the \pytrees\ documentation states
  that the language provides \emph{a good blend of purposeful planning
    towards goals with enough reactivity to shift in the presence of
    important events};
  \url{https://py-trees.readthedocs.io/en/devel/background.html}} but
can be simulated by sufficiently high-frequency model execution.

The \chg{analyzed DSLs support a global storage using the
  \emph{blackboard} behavioral design pattern
  \cite{laker2012blackboard}}, which is a key-value store. No scopes
are supported, all keys are global.  The blackboard is used for
communicating, both within the model and with the rest of the system.
The model and the system read and update the blackboard
asynchronously.

Both \BTCPP and \pytreesros offer the four basic categories of
control-flow nodes and the two basic execution nodes. To illustrate
the abstract syntax, we provide a meta-model that was
reverse-engineered from \BTCPP's XML format in
\cref{fig:metamodel}. These concepts are further detailed in
\cref{tbl:concepts}.

\begin{table}[t]


	\caption{\Bt concepts and corresponding language elements in \BTCPP\, \pytreesros and \pytrees}%
	\label{tbl:concepts-vs-langs}
	\renewcommand{\arraystretch}{1.1}
	\begin{tabularx}{\linewidth}{
			>{\small}p{12mm}
			>{\small\raggedright}p{30.5mm}
			>{\small}p{28mm}}
		\textsf{concept}
		& \BTCPP
		& \pytrees \pytreesros
		\\\midrule

		Simple Node
		& subclasses of \lstinline$ActionNode$ \newline \lstinline$ConditionNode$
		& \lstinline$behaviour.Behaviour$
		\\

		Composite
		& subclasses of \lstinline$ControlNode$
		& classes in \newline \lstinline$composites$
		\\

		Sequence
		& \lstinline$Sequence$,  \lstinline$SequenceStar$ \newline \lstinline$ReactiveSequence$
		& \lstinline$composites.Sequence$
		\\

		Selector
		& \lstinline$Fallback$, \lstinline$FallbackStar$ \newline \lstinline$ReactiveFallback$
		& \lstinline$composites.Selector$ \newline \lstinline$composites.Chooser$
		\\

		Decorator
		& subclasses of \lstinline$DecoratorNode$
		& classes in \newline \lstinline$decorators$
		\\

		Parallel
		& \lstinline$ParallelNode$
		& \lstinline$composites.Parallel$  \\
		\bottomrule
	\end{tabularx}
	\vspace{-2mm}

\end{table}

\subsubsection*{Simple Nodes}

Recall that simple nodes, or leaves in the syntax tree, are either
conditions or actions.  Actions realize the basic computation in the
model. Users of our \chg{DSLs} need to implement custom action
nodes---classes obeying the \lstinline!Action! interface that contain
the Python or C++ code to be executed whenever a node is
ticked. Conditions calculate a value of a Boolean predicate and
convert it to a success or failure value.

\BTCPP supports different types of synchronous and asynchronous action
nodes. Meanwhile \pytreesros supports mainly synchronous type. The
simplest action nodes are synchronous, so they terminate quickly and
return success or failure immediately. Asynchronous nodes may also
return a `running' status and use some form of concurrency to continue
operation.  The execution engine will attempt to trigger them at the
next epoch again.

\subsubsection*{Composite Nodes}

Recall that composite nodes are internal nodes of a \bt. Their main
function is to define the order of traversal at every time epoch (at
every trigger). Unlike the simple nodes, which need to be implemented
by the users, our DSLs provide a range of predefined composite
nodes. The \emph{root} node is the composite node that serves as an
entry point for every traversal, it contains another node as the
body. This node is re-entered to start every traversal.

\bchg{The semantics of the different composite nodes in our studied
  DSLs follow the literature on behavior trees, which we described in
  \cref{sec:background}. A \emph{sequence} node is similar to a
  \emph{forall} higher-order function, which is standard in many
  programming languages. A \emph{selector} node is similar to
  \emph{exist}. We confirm that in the implementations,
  \emph{parallel} node is really a misnomer: It does not execute nodes
  concurrently, but it generalizes sequence and selector to a range of
  \emph{policies}, described in each DSL's documentation.}

\looseness=-1
Since the execution is always a traversal of the entire tree, there is no direct support for jumps (goto). Instead, composite nodes can affect the traversal locally, in stark contrast to \sms. In the studied DSLs, a typical change of control allows an arbitrary change of state, often cross-cutting the syntax tree, depending on the returned status of a node.
\bchg{After each tick of composite nodes, and by propagation also simple nodes, explicit status is returned. Our DSLs support the same set of statuses as in the literature on behavior trees (see \cref{sec:background}). These values propagate upwards during the tree traversal according to the semantics of composite nodes.}

\subsubsection*{Decorators}
\looseness=-1
Decorators are \chg{implemented as } unary composite nodes (only one child) \chg{in the studied DSLs}. They decorate the sub-trees and modify their data or control-flow. \chg{\BTCPP and \pytreesros offer a wide range of constructs for Decorators.} An \emph{Inverter} flips the return status of a child between success and failure.  A \emph{Succeeder} always succeeds regardless the status returned by its child node. A \emph{Repeat} node, which is stateful, acts like a for-loop: it continues to trigger the child for a given number of ticks.  It increments an internal counter at every trigger.  The node succeeds (and resets the counter) on hitting a set bound.  It fails (and resets the counter) if the child fails. A \emph{Retry} node resembles a repeat node. Its main goal is to make a flaky node succeed.  Like Repeat it can run a node up to a set number of times, but unlike Repeat, it only retries when a node fails and it retries immediately without waiting for the next epoch.  It fails if the child failed in a given number of attempts.

\begin{observation}
  \looseness=-1 The concepts offered in \bt languages, and their semantics, stem from domain needs.\@ The studied behavior-tree \chg{DSLs} gather a
  number of constructs based on patterns that, according to users and
  developers, are frequently found in high-level control of autonomous
  systems.
\end{observation}


\subsubsection*{Openness}
\looseness=-1
The openness and indefiniteness of \bts are probably their most
interesting aspects, after the time-triggered coroutine-based model of
computation. Others have also noticed this in the context of
variability in DSLs \cite{tolvanen2019domain}.

\looseness=-1
Unlike in Ecore\,\cite{dslbook} or
UML, the languages' meta-models are \emph{not fixed}.  The \chg{DSLs provide the} meta-classes for composite nodes, while it
leaves the simple nodes abstract or only gives them bare bones
functionality (cf. \cref{fig:metamodel}). A user of the languages is expected to first
extend the meta-model by implementing the basic action nodes, then
link them together in a syntax tree, possibly using an external XML
file. This practice vaguely resembles stereotyping\,\cite{uml2017}.
Obviously, a user of Ecore can extend the meta-model classes and give
them new functionality at runtime as well. However, such use of Ecore
is considered advanced and is seen rather rarely.  The difference is
that of degree: there is essentially no way to consider using behavior-tree \chg{DSLs}
without creating custom nodes.

\subsubsection*{Prerequisites (User Demographics)}

The open nature of behavior-tree DSLs means that the experience of building and debugging models resembles very much language-oriented programming as practiced in the modeling- and language-design research community.  One constantly deals with meta-classes, composing them, traversing them, and so on.  Anybody familiar with building DSLs on top of Ecore or similar frameworks will definitely experience a \emph{déjà vu}, when using either \pytreesros\ or \BTCPP.\@

Given that many robotics engineers, and many ROS users, lack formal
training in computer science and software
engineering\,\cite{DBLP:conf/icse/AlamiDW18}, it is surprising to us
that this design seems to be well received in the community.   Even
within software engineering, language implementation and
meta-programming skills are often considered advanced.  Yet, using
behavior-tree \chg{DSLs} requires such skills.
A challenge for the modeling community is lurking here: to design a
behavior-tree DSL that, while remaining flexible and easy to integrate
with large and complex existing code bases, is much easier to use for
a regular robotics programmer.

\begin{observation}
	The flexibility and extensibility of \chg{the studied} behavior-tree \chg{DSLs} require language-oriented
	programming skills from developers.  The software-language engineering community
	could contribute by designing an accessible, but still flexible, dialect
	of \BT.
\end{observation}

\subsubsection*{Separation of Concerns}

Behavior-tree \chg{DSLs support} plat\-form-spe\-ci\-fic models (PSMs) built as part of a specific
robotics system to control behaviors at runtime.  The models are used
to simplify and conceptualize the description of behavior.  The
ability to reuse the same models with other hardware or similar
systems is not (yet!) a primary concern.  The studied behavior-tree DSLs not only are PSMs, but
tend to be very tightly integrated with the system. Custom nodes tend to refer to
system elements directly and interact with the system API.\@ As a
result, it is hard to use \chg{created} models separately from the robot.
While Groot can visualize a standalone XML file of a model, a working
build environment of ROS is needed just to visualize the syntax of a
\pytreesros\ model.  This may mean not only an installation of
suitable Python and ROS libraries, but, for example, a working
simulation of the robot, or even the hardware environments.  You need
to launch the system and inject a visualization call to inspect the
model!

It is in principle possible with both \chg{DSLs} to build models that
are completely decoupled from the system.  It suffices to route all
communication with the system via the blackboard. \BTCPP\ provides
dedicated XML primitives for this purpose, allowing the entire
behavior to be programmed in XML, provided the rest of the system can
read from and write to the blackboard.  This separation allows models
to be processed outside the system for visualization, testing,
grafting into other systems, and so on.  We definitely think this is a good
architectural practice to follow.  Nevertheless, it is not what we
observed in real-world models (cf. \cref{sec:models}).  Most models
mix the specification of behavior deeply with its implementation, making
separation virtually impossible.

\begin{observation}
	Behavior-tree models \chg{implemented with our DSLs} tend to be deeply intertwined with behavioral
	glue code linking them to the underlying software system. This makes
	operating on models outside the system difficult, hampering
	visualization, testing, and reuse.
\end{observation}


\subsection{Concepts and Semantics in State-Machine DSLs}
\label{subsec:SMsematicsyntax}

%
%
%
%
%
%
%
%
%

\chg{In the following}, we reflect on the syntax and
semantics that we observed in the state-machine
languages State MACHine (\smach) \cite{bohren2010smach} and Flexible Behavior
Engine (\flexbe) \cite{schillinger2015approach}.

\bchg{The implementation of \sms in both DSLs is event-triggered, or reactive, which is similar to the traditional implementation of \sms in different domains. Compared to the studied behavior-tree DSLs, the state-machine DSLs do not have an explicit notion of ticks and of reoccurring traversals. States are only entered once at the beginning of the model execution, revisited only if the control-flow gets there as a result of reactions to external events. In \flexbe, the current state could be interrupted and a transition based on external events could be triggered, allowing reactive programming. The interruption is possible due to an autonomy-level threshold that is associated with a state to allow human-in-the-loop decision incorporation. In contrast, no reactivity is available in \smach, and the current active state blocks the execution until an outcome is returned.}

\looseness=-1
Both DSLs resemble Moore machines (action-on-state)
\cite{moore1956gedanken} from the computational perspective. Users are
expected to implement a state-based interface as a Python class,
\lstinline!State! in \smach and \lstinline!EventState! in
\flexbe. Unlike the UML's state-machine implementation of Moore
machines, there is no entry/exit actions associated to the state
interface. There are different sets of provided functions, with the
constructor \lstinline!__init__! and the execution loop
\lstinline!execute! being the most important parts to implement the
associated action to a state. The constructors in both DSLs require
listing the outcomes to be returned by a state (triggering
events). Other types of state interface exist in both languages, like
\lstinline!ConditionState! in \smach and \lstinline!LockableState!  in
\flexbe. More details can be found in their documentation
\cite{flexbedoc, smachdoc}.

\looseness=-1
To deal with the data flow between states and the system,
\emph{UserData}, a locally scoped key-value dictionary is
used. Compared to \emph{blackboard}, \emph{UserData} defines local
input-output ports to access the needed data, while \emph{blackboard}
is more of a global storage accessed by any node. It has been
highlighted by online discussion forums\footnote{For example, the
  problem is raised in \BTCPP GitHub issues page: \#18 \#41 \#44} and
by researchers \cite{colledanchise2021implementation} that the current
implementation of \emph{blackboard} is causing name-clashes and
unwanted overwrites when the \bts grow. Interestingly, by getting
inspiration from the scoped input-output ports in state-machine
languages (\textsf{\href{https://github.com/BehaviorTree/BehaviorTree.CPP/issues/41}{github.com/BehaviorTree/BehaviorTree.CPP/issues/41}}),
developers of our behavior-tree DSLs added changes to scope
\emph{blackboard} (\textsf{\href{https://www.behaviortree.dev/migrationguide}{behaviortree.dev/migrationguide},
  \href{https://py-trees.readthedocs.io/en/devel/changelog.html\#x-2019-11-15-blackboards-v2}{py-trees.readthedocs.io/en/devel/changelog.html\#x-2019-11-15-blackboards-v2}}).

\smach and \flexbe support hierarchical nesting by means of the
design-pattern container,
  so users can create hierarchical state machines. A container is
simply a Python module that could be extended to support different
execution semantics. This design pattern is also used in the studied
languages to define constructs for common high-level control-flow
patterns. We distinguish between nesting container and control-flow
containers according to their functionality. \lstinline!StateMachine!
in \smach and \lstinline!OperatableStateMachine! in \flexbe are for
defining a state-machine model. Both are also used for nesting and
creating hierarchical state machines.

\chg{In \smach, \lstinline!Concurrence!, \lstinline!Sequence! and
  \lstinline!Iterator! are some of the offered control-flow
  containers.} \lstinline!Sequence! acts similar to the behavior-tree
control-flow node Sequence. The states are executed sequentially
according to a predefined order. \lstinline!Iterator! is a
repeat-until loop that iterates over states until a specified outcome
is reached. Finally, \lstinline!Concurrence! allows multiple
sub-states to be active at the same time through threading. Similar to
the behavior-tree Parallel node, no parallelism of execution is
available. An outcome policy defines the concurrency container outcome
using a key-value dictionary (container outcomes are the keys and the
sub-states' potential outcomes are the values). In \flexbe, only
concurrency behavior is supported through
\lstinline!ConcurrencyContainer!. An example of \smach iterator is
provided in \cref{fig:smexmsmach}.

In addition to offering frequent control-flow patterns,
\flexbe provides meta-classes for common states through the API
\lstinline!flexbe_state! and a separate state library
\lstinline!generic_flexbe_states!.
A documentation of the states from
\lstinline!flexbe_state! is available in
\cite[app.\,A.1]{schillinger2015approach}.

\looseness=-1
Providing constructs for frequent control-flow patterns seems common
in \chg{the studied} behavior-tree and state-machine DSLs. This might
relate to the nature of robotic missions that tend to have a sequence
of actions or some iterative tasks, enforcing the need for the
\chg{behavior-modeling} language to accommodate this type of
behaviors.  Implementing behavior design-patterns as language
constructs in modeling languages is sometimes a problematic design
decision, because it might increase language complexity. However, as
stated by Bosch \cite{bosch1997design}, it is actually the lack of
expressive constructs that increase language complexity and user
overhead. A language should fulfill its domain needs. \bchg{In our
  previous study \cite{ghzouli2020behavior}, we found a clear
  difference in the supported constructs in our comparison between the
  studied UML behavior-modeling languages (state diagram and activity
  diagram) and behavior-tree DSLs. A similar observation regarding the
  need to customize UML modeling languages was reported by Whittle et
  al. \cite{whittle2013state}. It is actually preferred in practice to
  use DSLs over UML modeling languages due to the expressiveness
  needed in robotics \cite{de2021survey}.  Thus, it is surprising to
  observe such a design decision by language developers, and it seems
  like a need for robotics applications.}

\begin{observation}
	Constructs for frequent control-flow patterns seem to be a common need in modeling robotics missions. Our studied state-machine and behavior-tree DSLs accommodate these needs using containers and composite node concepts, respectively.
\end{observation}

\subsubsection*{Openness}

Openness is a common feature in the studied languages due to the
nature of robotic missions. Similar to \bts, the state-machine DSLs do
not constrain users with fixed models and implementations. Users are
provided with meta-classes for containers and states, and they can
extend them. The extension of control-flow types is possible, however
we have not observed any such customization in the analyzed projects
(see \cref{sec:models}), just like the behavior-tree projects. Further
investigation by including the users of languages would be required to
determine the reason.

This design pragmatically supports openness of the language and makes
adaptation to diverse scenarios in robotics easy. The openness seems
to be required due to a lack of agreement in the robotics community
about the ideal control model for robot behavior.  Since this question
is likely to remain open for a long time, the design allows users to
adapt the language as they see fit when building robots.


\section{Language Implementation (RQ2)}
\label{sec:implementation}

This section focuses on analyzing the behavior-tree libraries in the
first three rows in \Cref{tbl:btsmlanguages}, and the state-machine
libraries in the first two rows in the second half of the table, all
set in bold font. For the identified libraries, we broaden the scope
of our analysis by inspecting their implementation techniques and
practices.

\begin{figure*}[t]
	\begin{center}
		\includegraphics[
		width=\linewidth,
		clip,
		trim=25 40 10 5
		]{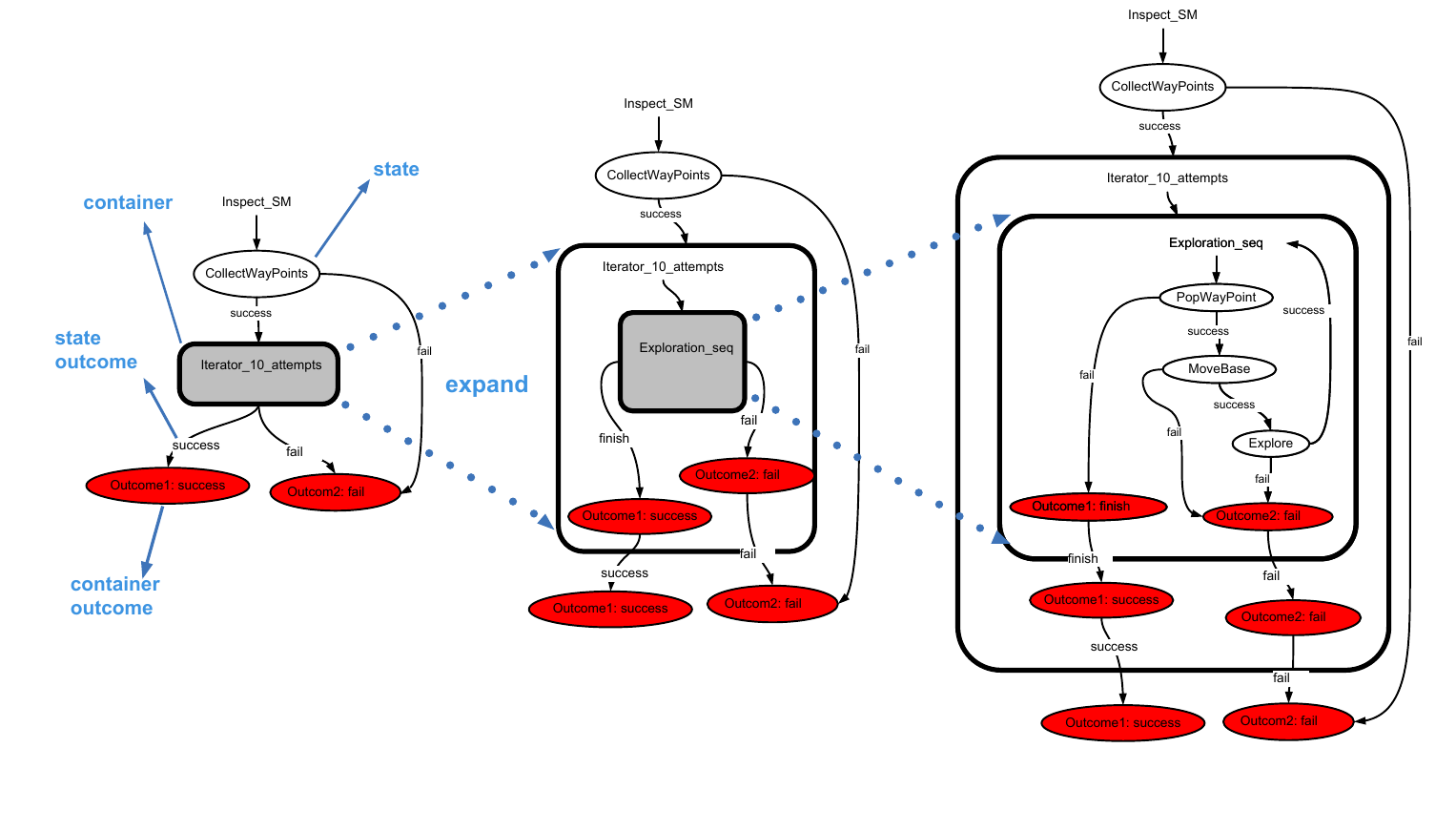}
	\end{center}
	\caption{A state-machine example representing the  mission in \cref{fig:figure-1} and \cref{fig:smexm}, using smach\_viewer syntax. On the left, a top view is provided. The middle and right illustrations represent expanded iterator and nesting containers, respectively}\label{fig:smexmsmach}	
\end{figure*} 

\subsection{Behavior-Tree  DSLs: Language Design}

Turning our attention to how behavior-tree languages are implemented from the
language-design perspective, the first striking observation is that
both languages are predominantly distributed as libraries, not as
language tool chains, or modeling environments.  \BTCPP\ is
implemented as a \texttt{C++} library, packaged as a ROS component,
easy to integrate with a ROS-based codebase\,\cite{mood2be}.  In
contrast, \pytrees\ is a pure \texttt{Python} library. It has an
extension \pytreesros\, which packages \pytrees\ as a ROS package and adds ROS-specific nodes.

\subsubsection*{Syntax and Visualization}
\label{subsec:BTconcretesyntax}

Both DSLs come with ways to visualize models as graphs. \BTCPP\ even has a graphical editor and a visual runtime monitor for its models called Groot (which created the \bt in \cref{fig:figure-1}).

\looseness=-1 Nevertheless, it is important to understand that \bts
are \emph{not} a visual modeling language in a traditional sense.
First, in both libraries, the models are constructed in a text editor,
in a mixture of C++, respectively Python.  Second, the models are
constructed directly in abstract syntax, by instantiating and wiring
abstract syntax types. For convenience, and to support Groot, \BTCPP\
offers an XML format, which can be used to write the tree syntax in
static files. \Cref{ls:xmlnotation} shows the XML file for the mission
displayed in \Cref{fig:figure-1}.  This file is interpreted
at runtime, and an abstract syntax tree is constructed from it
dynamically. Third, crucially, the types of nodes (and, thus, the XML
file in \BTCPP) do not constitute the entire meaning of the model. An
important part of the model is embedded in C++/Python code that is
placed in the methods of the custom node classes. This part of the
model is neither modifiable nor presentable in the graphical tools.

Finally, recall that \BTCPP is realized as an external DSL through Groot and the XML-like format, while \pytreesros constitutes an internal DSL, since it does not have similar tools. From our experience analyzing their models (cf. \cref{sec:models}), we can confirm that the \BTCPP models are much easier to comprehend, and the availability of its visual editor Groot has made it faster to analyze the \bt models than \pytreesros models.
\lstset{
	language=XML,
	caption={\label{ls:xmlnotation}The same example in \Cref{fig:figure-1} shown in \BTCPP XML notation.},
	columns=fullflexible,
	backgroundcolor=\color{backcolour},
	keywordstyle=\color{red},
	stringstyle=\color{blue-violet},
	morekeywords={encoding,
		main_tree_to_execute,name,num_attempts},
	classoffset=1,
	morekeywords={False},
	keywordstyle=\color{blue},
	morekeywords={encoding,
		root,BehaviorTree,SequenceStar,Sequence,Action,Negation,RetryUntilSuccesful}
}
\begin{lstlisting}
<root main_tree_to_execute="MainTree" >
     <BehaviorTree ID="MainTree">
		  <SequenceStar name="MainSeq">
            <Action ID="CollectWaypoints"/>
            <RetryUntilSuccesful num_attempts="10" >
                <Negation>
                    <Sequence name="ExplorationSeq">
                        <Action ID="PopWaypoint"/>
                        <Action ID="MoveBase"/>
                        <Action ID="Explore"/>
                    </Sequence>
                </Negation>
            </RetryUntilSuccesful>
            </SequenceStar>
     </BehaviorTree>
</root>

\end{lstlisting}

\subsubsection*{Concurrency}
\label{subsec:BTconcurrency}

The design of our behavior-tree DSLs does not prescribe the model of
concurrency, and implementations vary.
For instance, \bchg{the \BTCPP\ engine uses a single-threaded execution engine. All nodes are supposed to return a status immediately. If an operation lasts longer, it should spawn an asynchronous logic and inform the behavior-tree engine that it continues to run (or that it completed, if ticked again upon the completion).  So, the programmer has complete freedom to choose the concurrency mechanism. }

\begin{observation}
	The implementations of our behavior-tree DSLs support both interleaving and true concurrency indirectly by resorting to the underlying ROS platform. The model of concurrency is not defined strictly in the language, but is, instead, left largely to the users.

\end{observation}

\subsubsection*{An internal or external DSL?}
\label{subsec:BTimpl_opennness}

Our DSLs are unusually open.  \BTCPP\ is technically an external
DSL, but its implementation exposes aspects of dynamic internal DSLs.
The programmer can both create models in XML (external, static), and
create new node types or modify the shape of the syntax tree at
runtime (dynamic).  \pytreesros\ is an entirely dynamic DSL, where new
node types and Python code can be freely mixed, like in internal DSLs.

\subsubsection*{An interpreter or a compiler?}
\label{subsec:BTimpl_interpreter}

Our DSLs' models are interpreted.  Once the abstract syntax tree is
constructed, the user is supposed to call a method to trigger the
model once, or to trigger it continuously at a fixed frequency.  This
does not seem to depart far from other applications of
models-at-runtime\,\cite{blair2009models,bencomo14modelsatruntime}.  \BTCPP\ uses template
meta\-pro\-gramming instead of \chg{code-generation}, which allows to offer a
bit of type-safety when implementing custom tree nodes, without
exposing users to any specialized code-generation tools. Using the
library appears like using a regular C++ library.  As expected, no
static type safety is offered in \pytreesros.
\vspace{-2mm}

\subsection{State-Machine DSLs: Language Design}
\label{statemachineArchitecture}

\smach and \flexbe are open-source software frameworks written in
Python for building and monitoring \sms and hierarchical state
machines. Similar to \pytreesros, \smach can be used without the ROS
system and it has a ROS binding through \smachros.

\looseness=-1
Looking at how these state-machine DSLs are implemented from a language-design
perspective, there is a difference in how \smach and \flexbe are
realized. Although \flexbe can be used directly as a Python library,
it is used as a modeling environment, since it offers a graphical user
interface (GUI) called FlexBE
App (\textsf{\href{https://github.com/FlexBE/flexbe_app}{github.com/FlexBE/flexbe\_app}}). It
is used to construct \sms using a graphical notation, and to monitor
and modify the \sm during runtime. On the other hand, \smach is mainly
used as a pure Python library, and it has a package \smachros to
integrate it with ROS. It has a GUI called
smach\_viewer (\textsf{\href{http://wiki.ros.org/smach_viewer}{wiki.ros.org/smach\_viewer}}) for
inspecting and debugging an already created state machine, but not
constructing or modifying it. Although \smach is ROS independent, its
viewer cannot be executed without having a running ROS.

\subsubsection*{Syntax and Visualization}
\label{subsec:SMSyntax}
\smach and \flexbe have their own notation, different from the UML
standard in \cref{fig:smsyntax}. \smach uses an ellipse shape for
states and a double-frame ellipse shape for nested machines, and
\flexbe uses rectangle shapes. \flexbe uses a dedicated notation for
initial and final states like UML, while \smach does
not. \Cref{fig:smexmsmach} shows a state-machine example using the
smach\_viewer syntax. \bchg{Events are called outcomes in \smach and
  \flexbe. \smach distinguishes between a container outcome and a
  state outcome in the visual syntax. The outcome of a state that
  caused a transition (triggering event) is represented on the
  transitioning arrow. The outcome of a container, like the
  \lstinline!Interator_10_attempts! in \cref{fig:smexmsmach}, is
  represented explicitly as a red ellipse, which looks like a
  state. This deviates from the typical state-machine syntax.}

The two languages are used differently for constructing the
state-machine model. \flexbe mixes graphical and textual syntax. The
graphical syntax is constructed using the FlexBE App (\flexbe
graphical GUI) in a drag-and-drop manner, then an interpreter
generates the textual syntax as Python code. The generated code
represents the model structure, and certain parts of it can be edited
to provide flexibility to developers. For each state, a stub is
auto-generated for class methods consisting of basic functions, e.g.,
a constructor, execution function, and so on. Developers are expected
to complete the auto-generated stubs for states in Python code.
Meanwhile, \smach users construct the model's abstract syntax directly
in a text editor in Python.

\bchg{\flexbe uses MDE techniques in the construction of the model. It
  makes use of model-based design processes through (semi-)automatic
  generation of code from the models. Syntactic and semantic checks
  for model consistency are also supported. Consequently,
  separation-of-roles is supported in \flexbe.} \chg{\BTCPP also
  supports separation-of-roles \cite{mood2be}, in addition to
  separation-of-concerns,} by providing a GUI for constructing the
graphical model and generating the textual syntax in an XML format
from the model, which is useful, but the language still requires
manual repetitive coding. While \flexbe reduces the manual coding by
generating boilerplate code, such as the import of needed libraries
and default functions for the state initialization. If adopted by
behavior-tree languages, this functionality might reduce programming
effort and reduce syntactic errors. However, since custom-execution
functions still need some manual coding, it is not fully applicable
without some programming background.

\begin{observation}
  One of the state-machine DSLs has used concepts from model-driven
  engineering that reduce repetitive coding and code syntactic
  errors. Enhancing the user support of behavior-tree DSLs for similar
  features could reduce programming effort and syntactic errors.
\end{observation}

\subsubsection*{Concurrency}
\label{subsec:SMconcurrency}
\bchg{Similar to \bts, the model of concurrency is not strictly
  enforced in the implementation. The user can exploit any concurrency
  semantics available in the ROS platform.  In both libraries, every
  state-machine container is executed in its own thread, and its
  \emph{UserData}, the locally scoped dictionary of key-value pairs,
  is not shared between the different state-machine containers. To
  facilitate data sharing between two running \sms, one needs to
  resort to external synchronization and communication mechanisms.}

The analyzed DSLs offer a concurrency container. In both \chg{DSLs},
Python threading is used allowing multiple states to be active at the
same time, each running at its own rate. States are activated
sequentially, not parallel due to using the Python threading
(interleaving).

\subsubsection*{An internal or external DSL?}
\label{subsec:SMimpl_opennness}
The openness of \flexbe is similar to other modeling languages using
model-based design. Similar to \BTCPP, it is realized as an external
DSL. \chg{State-machine models} can be created using the graphical
notation in the FlexBE App.  \chg{Syntactic and minor semantic
  verifications run before generating code to ensure model
  consistency, such as the existence and correctness of the associated
  \emph{UserData} with each state.}  Extension of the model and states
is possible offline. Once the model and states are instantiated,
runtime modification is limited to the structure of the state-machine
model, e.g., add and remove states, and transition function, while
modifying the implementation of a state is not possible. On the other
hand, \smach is similar to \pytreesros and it is realized as an
internal DSL, where extending the model and states is done easily by
simply editing the Python code.

\subsubsection*{An interpreter or a compiler?}
\label{subsec:SMimpl_interpreter}
\looseness=-1 Both libraries execute the state-machine model through interpretation at runtime. \flexbe allows modifications at runtime.
In \flexbe, modification to a running \sm is simply realized through patches---a standard for minimal summaries of the modifications. Patches are applied by taking the difference between old and new code and only re-importing the changes to the source code. In order to apply the modifications, the modified state is locked to prevent transitions during the modification using the Python threading lock(\textsf{\href{https://docs.python.org/3/library/threading.html}{docs.python.org/3/library/threading.html}}), then the Python module is reloaded (using the built-in reload() function).
No modification is applied before the \flexbe engine has run verification checks to ensure consistency (cf. \cref{subsec:SMSyntax}). So, overall, the \flexbe developers implemented a pragmatic state-machine swap at runtime, largely relying on Python's reload support. \smach has no verification step for the created model before execution.

%


\section{Behavior-Tree and State-Machine Models (RQ3)}%
\label{sec:models}
Our mining returned hundreds of GitHub projects per DSL. In the following, we report on our \chg{exploratory} investigation of the usage of state-machine and behavior-tree \chg{DSLs} in these projects. First, \chg{we} present an overview of the different languages' popularity among the mined open-source projects. Second, we provide our analysis results for a sample of these projects. Our analysis covers the structure of the sampled models and their code reuse among the projects.

\begin{figure}[t]
	
		\includegraphics[
		width=\linewidth,
		clip,
		trim=6 8 10 10
		]{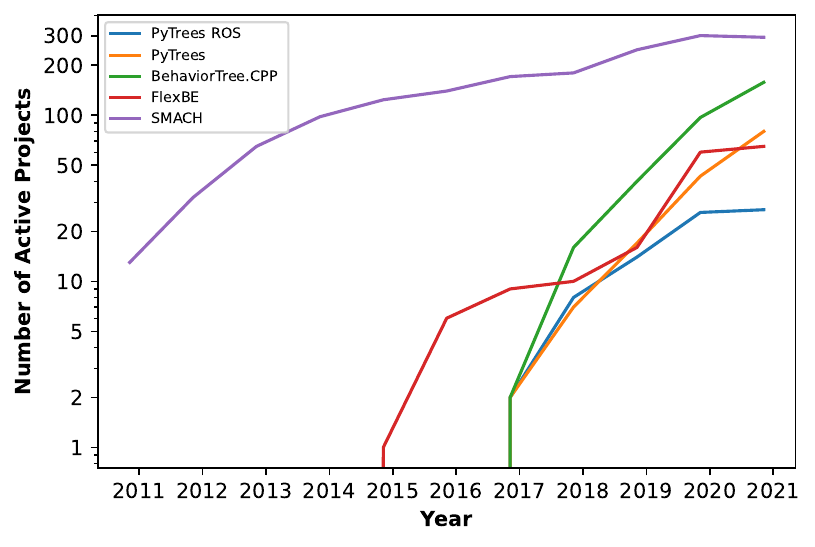}
	
	
	\caption{Usage of the state-machine and behavior-tree DSLs in open-source projects on GitHub over time}%
	\label{fig:trend}
	
	\vspace{-3mm}
	
\end{figure}

\subsection{Language Popularity}

The initial mining returned \num{1086} projects using our state-machine \chg{DSLs} and \num{271} projects \chg{using} our behavior-tree \chg{DSLs} (see \cref{subsec: minegithubmeth} for the search terms). To visualize the language (i.e., library) use in open-source projects over time, we tracked the number of active projects per year per library in these mined projects and plotted each library's trend line. 

\begin{figure*}[t]
	\begin{center}
		\includegraphics[
		width=\linewidth,
		clip,
		trim=20 20 20 15
		]{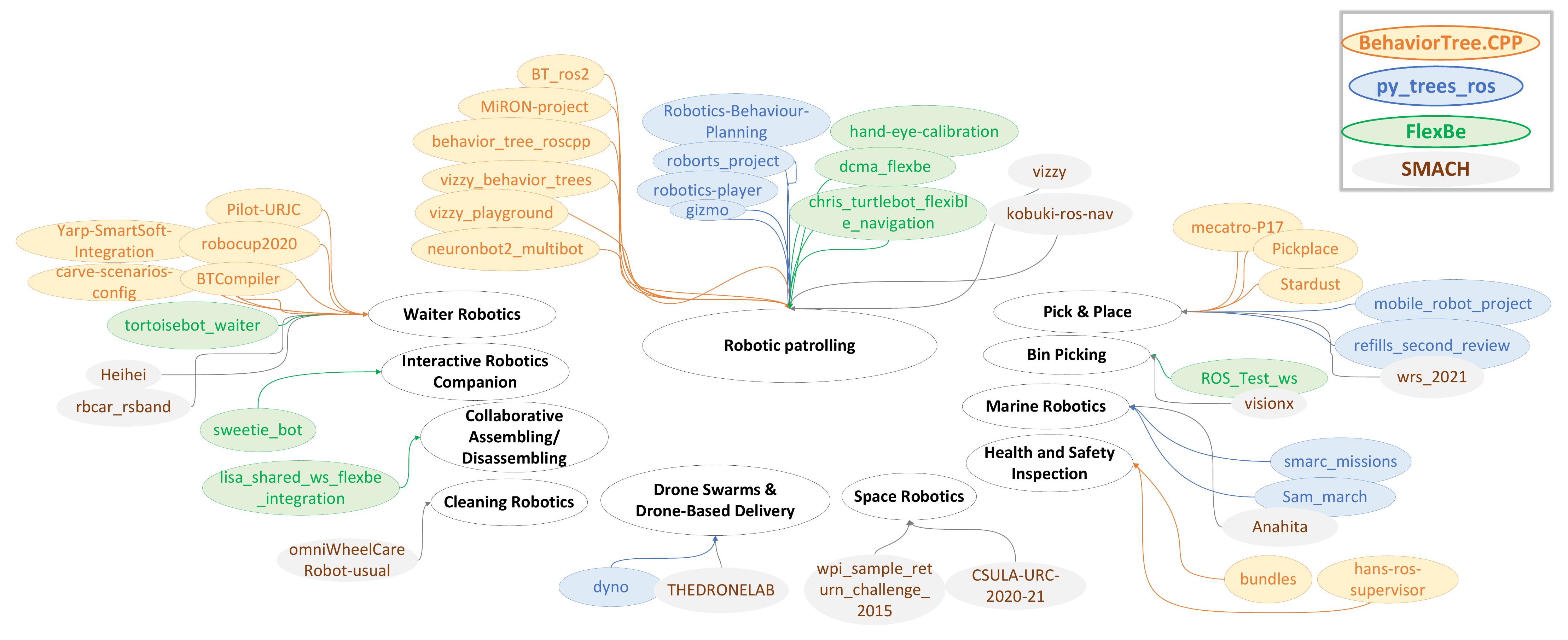}
	\end{center}
	
	\caption{An overview of the domains of the analyzed projects}%
	\label{fig:projectsdomains}
\end{figure*}

\Cref{fig:trend} shows the trend. Until 2015, \smach was the only available language among our five DSLs, and its usage has been increasing steadily over the past five years (1.2 times more each year). In 2015, the other four languages were released. However, their usage only started to increase significantly in 2018. In general, the use of the languages has grown over the past five years. In 2021, \BTCPP's and \pytrees's usage in \chg{open-source} projects reached almost ten times what they were in 2018. We hypothesize that this significant increase in the last three years is due to ROS adopting it as the core component for its navigation stack Navigation~2\,\cite{macenski2020marathon}. For \pytrees, it could be because it was the only Python implementation for \bts offering a stable and actively maintained language and a wide range of constructs for node types. Although the \flexbe usage in open-source projects is lower than \smach (78\% lower than \smach in 2021), its overall usage growth since 2018 is much higher than that of \smach. In 2021, the \flexbe usage was \num{6.5} times higher than in 2018, while that of \smach only increased around 1.6 times.


\begin{table}[H]
	\caption{Overview on our sampled GitHub projects that use behavior-tree and state-machine DSLs to define robot behavior \label{tab:githubprojects-1}}
	\vspace{-3mm}
	\begin{tabular}[h]
		{
			>{\footnotesize\raggedright}p{45mm}
			>{\footnotesize}p{15mm}
			>{\footnotesize}p{1mm}
			>{\footnotesize}p{10mm}}
		
		\textsf{project, GitHub link}
		& \textsf{language} 
		& \rotatebox{90}{\hspace{-.05cm}\textsf{models}}
		& \textsf{Type} 
		\\
		\midrule
		sam\_march\newline
		\rlap{~~\scriptsize \MYhref{https://github.com/KKalem/sam\_march}{KKalem/sam\_march}}
		& \texttt{\pytreesros}
		& 1
		& \MYhref{https://smarc.se/}{research}
		\\
		mobile\_robot\_project\newline
		\rlap{~~\scriptsize \MYhref{https://github.com/simutisernestas/mobile\_robot\_project}{simutisernestas/mobile\_robot\_project}}
		& \texttt{\pytreesros}
		& 1
		& unknown
		\\    
		smarc\_missions\newline
		\rlap{~~\scriptsize \MYhref{https://github.com/smarc-project/smarc\_missions}{smarc-project/smarc\_missions}}
		& \texttt{\pytreesros}
		& 2
		& \MYhref{https://smarc.se/}{research}
		\\
		dyno\newline
		\rlap{\scriptsize  \MYhref{https://github.com/samiamlabs/dyno}{samiamlabs/dyno}}
		& \texttt{\pytreesros}
		& 2
		& \MYhref{https://dynorobotics.se/}{company}
		\\    
		gizmo \newline
		\rlap{~~\scriptsize \MYhref{https://github.com/peterheim1/gizmo}{peterheim1/gizmo}}
		& \texttt{\pytreesros}
		& 8
		& unknown
		\\
		roborts\_project \newline
		\rlap{~~\scriptsize \MYhref{https://github.com/Taospirit/roborts\_project}{Taospirit/roborts\_project}}
		& \texttt{\pytreesros}
		& 1
		& unknown
		\\
		robotics-player \newline
		\rlap{~~\scriptsize \MYhref{https://github.com/braineniac/robotics-player}{braineniac/robotics-player}}
		& \texttt{\pytreesros}
		& 1
		& unknown
		\\  
		refills\_second\_review\newline
		\rlap{~~\scriptsize \MYhref{https://github.com/refills-project/refills\_second\_review}{refills-project/refills\_second\_review}}
		& \texttt{\pytreesros}
		& 1
		& \MYhref{http://www.refills-project.eu/}{research}
		\\
		\textls[-10]{Robotics-Behaviour-Pla\rlap{nning}} \newline
		\rlap{~~{\scriptsize \MYhref{https://github.com/jotix16/Robotics-Behaviour-Planning}{jotix16/Robotics-Behaviour-Planning}}}
		& \texttt{\pytreesros}
		& 3
		& unknown
		\\
		pickplace\newline
		\rlap{~~\scriptsize \MYhref{https://github.com/ipa-rar/pickplace}{ipa-rar/pickplace}}
		& \texttt{\BTCPP}
		& 1
		& \MYhref{https://www.ipa.fraunhofer.de/}{research}
		\\
		stardust\newline
		\rlap{~~\scriptsize \MYhref{https://github.com/julienbayle/stardust}{julienbayle/stardust}}
		& \texttt{\BTCPP}
		& 4
		& \MYhref{https://julienbayle.github.io/stardust/}{research}
		\\
		neuronbot2\_multibot\newline
		\rlap{~~\scriptsize \MYhref{https://github.com/skylerpan/neuronbot2\_multibot}{skylerpan/neuronbot2\_multibot}}
		& \texttt{\BTCPP}
		& 2
		& unknown
		\\
		mecatro-P17\newline
		\rlap{~~\scriptsize \MYhref{https://github.com/alexandrethm/mecatro-P17}{alexandrethm/mecatro-P17}}
		& \texttt{\BTCPP}
		& 11
		& unknown
		\\
		Yarp-SmartSoft-Integra\rlap{tion}\newline
		\rlap{~~\scriptsize \MYhref{https://github.com/CARVE-ROBMOSYS/Yarp-SmartSoft-Integration/}{CARVE-ROBMOSYS/Yarp-SmartSoft-Integration}}
		& \texttt{\BTCPP}
		& 1
		& \MYhref{https://robmosys.eu/carve/}{research}
		\\
		bundles\newline
		\rlap{~~\scriptsize \MYhref{https://github.com/MiRON-project/bundles}{MiRON-project/bundles}}
		& \texttt{\BTCPP}
		& 5
		& \MYhref{https://robmosys.eu/miron/}{research}
		\\
		BTCompiler\newline
		\rlap{~~\scriptsize \MYhref{https://github.com/CARVE-ROBMOSYS/BTCompiler}{CARVE-ROBMOSYS/BTCompiler}}
		& \texttt{\BTCPP}
		& 8
		& \MYhref{https://robmosys.eu/carve/}{research}
		\\
		BT\_ros2\newline
		\rlap{~~\scriptsize \MYhref{https://github.com/Adlink-ROS/BT\_ros2}{Adlink-ROS/BT\_ros2}}
		& \texttt{\BTCPP}
		& 2
		& \MYhref{https://www.adlinktech.com/en/ROS2-Solution.aspx}{company}
		\\
		vizzy\_behavior\_trees\newline
		\rlap{~~\scriptsize \MYhref{https://github.com/vislab-tecnico-lisboa/vizzy\_behavior\_trees}{vislab-tecnico-lisboa/vizzy\_behavior\_trees}}
		& \texttt{\BTCPP}
		& 7
		&  \MYhref{https://vislab.isr.tecnico.ulisboa.pt/}{research}
		\\
		hans-ros-supervisor\newline
		\rlap{~~\scriptsize \MYhref{https://github.com/kmi-robots/hans-ros-supervisor}{kmi-robots/hans-ros-supervisor}}
		& \texttt{\BTCPP}
		& 1
		& \MYhref{https://robots.kmi.open.ac.uk/index.html\#meetHans}{research}
		\\    
		Pilot-URJC\newline
		\rlap{~~\scriptsize \MYhref{https://github.com/MROS-RobMoSys-ITP/Pilot-URJC}{MROS-RobMoSys-ITP/Pilot-URJC}}
		& \texttt{\BTCPP}
		& 2
		& \MYhref{https://robmosys.eu/mros/}{research}
		\\
		vizzy\_playground\newline
		\rlap{~~\scriptsize \MYhref{https://github.com/vislab-tecnico-lisboa/vizzy\_playground}{vislab-tecnico-lisboa/vizzy\_playground}}
		& \texttt{\BTCPP}
		& 6
		& \MYhref{https://vislab.isr.tecnico.ulisboa.pt/}{research}
		\\
			behavior\_tree\_rosC++\newline
		\rlap{~~\scriptsize \MYhref{https://github.com/ParthasarathyBana/behavior\_tree\_rosC++}{ParthasarathyBana/behavior\_tree\_rosC++}}
		& \texttt{\BTCPP}
		& 1
		& unknown
		\\

		\bottomrule
	\end{tabular}
\end{table}%

\begin{table}[H]
	\ContinuedFloat
	\caption{Sampled projects overview (continued) \label{tab:githubprojects-2}}
	\begin{tabular}[h]
		{
			>{\footnotesize\raggedright}p{45mm}
			>{\footnotesize}p{15mm}
			>{\footnotesize}p{1mm}
			>{\footnotesize}p{10mm}}
		\textsf{project, GitHub link}
		& \textsf{language} 
		& \rotatebox{90}{\hspace{-.05cm}\textsf{models}}
		& \textsf{Type}
		\\
		\midrule
		MiRON-project\newline
		\rlap{~~\scriptsize \MYhref{https://github.com/ajbandera/MiRON-project}{ajbandera/MiRON-project}}
		& \texttt{\BTCPP}
		& 1
		& \MYhref{https://robmosys.eu/miron/}{research}
		\\
		robocup2020\newline
		\rlap{~~\scriptsize \MYhref{https://github.com/IntelligentRoboticsLabs/robocup2020}{IntelligentRoboticsLabs/robocup2020}}
		& \texttt{\BTCPP}
		& 2
		& \MYhref{https://robolab.science.uva.nl/}{research}
		\\
		carve-scenarios-config\newline
		\rlap{~~\scriptsize \MYhref{https://github.com/CARVE-ROBMOSYS/carve-scenarios-config}{CARVE-ROBMOSYS/carve-scenarios-config}}
		& \texttt{\BTCPP}
		& 1
		& \MYhref{https://carve-robmosys.github.io/}{research}
		\\
		
		ROS\_Test\_ws\newline
		\rlap{~~\scriptsize \MYhref{https://github.com/QuiN-cy/ROS\_Test\_ws}{QuiN-cy/ROS\_Test\_ws}}
		& \texttt{\flexbe}
		& 6
		& unknown
		\\
		lisa\_shared\_ws\_flexbe\_integration\newline
		\rlap{~~\scriptsize \MYhref{https://github.com/lawrence-iviani/lisa\_shared\_ws\_flexbe\_integration}{lawrence-iviani/lisa\_shared\_ws\_flexbe\_integration}}
		& \texttt{\flexbe}
		& 4 
		& \MYhref{https://www.dropbox.com/s/1ddzlee5ejinccg/liviani\_defense\_presentation\_hires.pdf}{research}
		\\
		sweetie\_bot\newline
		\rlap{~~\scriptsize \MYhref{https://github.com/sweetie-bot-project/sweetie\_bot/}{sweetie-bot-project/sweetie\_bot}}
		& \texttt{\flexbe}
		& 5
		& \MYhref{https://sweetie.bot/}{company}
		\\
		chris\_turtlebot\_flexible\_navigation\newline
		\rlap{~~\scriptsize \MYhref{https://github.com/CNURobotics/chris\_turtlebot\_flexible\_navigation/}{CNURobotics/chris\_turtlebot\_flexible\_navigation}}
		& \texttt{\flexbe}
		& 3
		& \MYhref{https://dvic.devinci.fr/}{research}
		\\
		hand-eye-calibration\newline
		\rlap{~~\scriptsize \MYhref{https://github.com/tku-iarc/hand-eye-calibration}{tku-iarc/hand-eye-calibration}}
		& \texttt{\flexbe}
		& 2
		& \MYhref{https://www.iarc.tku.edu.tw/}{research}
		\\
		dcma\_flexbe\newline
		\rlap{~~\scriptsize \MYhref{https://github.com/andy-Chien/dcma\_flexbe}{andy-Chien/dcma\_flexbe}}
		& \texttt{\flexbe}
		& 8
		& unknown
		\\
		tortoisebot\_waiter\newline
		\rlap{~~\scriptsize \MYhref{https://github.com/rigbetellabs/tortoisebot\_waiter}{rigbetellabs/tortoisebot\_waiter}}
		& \texttt{\flexbe}
		& 2
		& \MYhref{https://rigbetellabs.com/}{company}
		\\
		Anahita\newline
		\rlap{~~\scriptsize \MYhref{https://github.com/AUV-IITK/Anahita}{AUV-IITK/Anahita}}
		& \texttt{\smach}
		& 2
		& \MYhref{https://auviitk.com/}{research}
		\\
		wrs\_2021\newline
		\rlap{~~\scriptsize \MYhref{https://github.com/hentaihusinsya/wrs\_2021}{hentaihusinsya/wrs\_2021}}
		& \texttt{\smach}
		& 2
		& unknown
		\\
		Heihei\newline
		\rlap{~~\scriptsize \MYhref{https://github.com/Luobokeng2021/Heihei}{Luobokeng2021/Heihei}}
		& \texttt{\smach}
		& 2
		& unknown
		\\
		THEDRONELAB\newline
		\rlap{~~\scriptsize \MYhref{https://github.com/DeVinci-Innovation-Center/THEDRONELAB}{DeVinci-Innovation-Center/THEDRONELAB}}
		& \texttt{\smach}
		& 2
		& \MYhref{https://dvic.devinci.fr/}{research}
		\\
		kobuki-ros-nav\newline
		\rlap{~~\scriptsize \MYhref{https://github.com/vibin18/kobuki-ros-nav}{vibin18/kobuki-ros-nav}}
		& \texttt{\smach}
		& 3
		& unknown
		\\
		visionx\newline
		\rlap{~~\scriptsize \MYhref{https://github.com/tsoonjin/visionx}{tsoonjin/visionx}}
		& \texttt{\smach}
		& 3
		& unknown
		\\
		wpi\_sample\_return\_challenge\_2015\newline
		\rlap{~~\scriptsize \MYhref{https://github.com/rcxking/wpi\_sample\_return\_challenge\_2015}{rcxking/wpi\_sample\_return\_challenge\_2015}}
		& \texttt{\smach}
		& 3
		& \MYhref{https://www.nasa.gov/press/2014/september/nasas-2015-sample-return-robot-challenge-open-for-registration/}{research}
		\\
		rbcar\_rsband\newline
		\rlap{~~\scriptsize \MYhref{https://github.com/darshan-kt/rbcar\_rsband}{darshan-kt/rbcar\_rsband}}
		& \texttt{\smach}
		& 4
		& unknown
		\\
		vizzy\newline
		\rlap{~~\scriptsize \MYhref{https://github.com/vislab-tecnico-lisboa/vizzy}{vislab-tecnico-lisboa/vizzy}}
		& \texttt{\smach}
		& 6
		& \MYhref{https://vislab.isr.tecnico.ulisboa.pt/}{research}
		\\
		omniWheelCareRobot-usual\newline
		\rlap{~~\scriptsize \MYhref{https://github.com/Art-robot0/omniWheelCareRobot-usual}{Art-robot0/omniWheelCareRobot-usual}}
		& \texttt{\smach}
		& 7
		& unknown
		\\
		CSULA-URC-2020-21\newline
		\rlap{~~\scriptsize \MYhref{https://github.com/CSULA-URC/CSULA-URC-2020-21}{CSULA-URC/CSULA-URC-2020-21}}
		& \texttt{\smach}
		& 11
		& \MYhref{https://www.ieee-ras.org/images/Standards/meeting\_june\_2021/8-Robotics\_Presentation\_2021-06-Guzman.pdf}{research} 
		\\		
		\bottomrule
	\end{tabular}
\end{table}%

\subsection{Characteristics of the Models}
\label{modelCharacteristics}
\looseness=-1 After applying the multiple filtering steps in our mining, we ended up with \num{620} projects that include \num{2507} state-machine models used in them, and \num{169} projects that include \num{658} behavior-tree models. \cref{tbl:finalnum} shows the number of projects and models per DSL. In a project, a file with a full definition of a model is counted as a model, resulting in projects with multiple models. \chg{We} randomly sampled \num{75} models per language.  We analyzed a total of 150 models of \bts and \sms (75 models each) belonging to 43 projects (25 \bts and 18 \sms), as summarized \chg{in} \Cref{tab:githubprojects-1}. The difference in the number of projects stems from the choice to sample the same number of models rather than projects.

\looseness=-1 In our sample dataset, we found eleven different domains, as shown in \Cref{fig:projectsdomains}. The dominating domains are robotic patrolling, waiter robotics and pick\&place applications. \bchg{We identified three types of projects: (1) research projects belonging to labs, research groups, or competing teams in events, (2) company projects belonging to companies producing robotic solutions, and (3) unknown projects, for which we could not find enough information to characterize. \Cref{tab:githubprojects-1} presents the types of our projects with a link to their corresponding organization and their GitHub repository by hovering over the blue text.}

\begin{table}[t]
	\centering
	\caption{Projects and models extracted per library}%
	\label{tbl:finalnum}
	\begin{tabularx}{\linewidth}{lp{1.1cm}p{1.1cm}p{1.8cm}p{1cm}}
		& \smach & \flexbe & \BTCPP & \pytreesros  \\
		\midrule
		\textsf{projects} & \ \ 560     & \ \ 60   & 141   & 28      \\
		\textsf{models}   & 2065    & 442  &595    & 63    \\
		\bottomrule
	\end{tabularx}
\end{table}

\looseness=-1 We noticed a variation between the structural properties of behavior-tree and state-machine models \chg{in the analyzed sample}. 
As a start, state-machine models have a right-skewed model-size distribution with an average model size of 9 across the population. Similarly, behavior-tree models have a right-skewed model-size distribution, but the average model size is three times larger than that of \sms (the average model size is 26 across the population). Interestingly, the average branching factor BT.ABF is 3, which is small, meaning the developers kept the trees in a somehow manageable size. The reason behavior-tree models seem to have a larger model size compared to \sms could be linked to the former representing both execution nodes and control-flow nodes, which state-machine models lack \cite{colledanchise2016behavior}. Another reason could be linked to state-machine models becoming complex once their size is big, affecting their understandability\,\cite{cruz2010impact, genero2002defining}. Consequently, developers might try to keep them small. For \bts, \chg{there is no work in the literature that} confirms the optimal size of models. However, through our analysis of models we noticed that the combination of moderate size and branching of nodes makes a model easy to navigate and understand. By moderate, we refer to the average reported values for both BT.size and BT.ABF.

Another structural property we were interested to capture is nesting and depth. According to existing empirical studies\,\cite{cruz2005investigating, cruz2005empirical} the nesting level affects the understandability of state-machine models, so shallow models should be preferred. 


In our sample, state-machine models have an average nesting level of one, and behavior-tree models have an average depth of five (similar across \chg{the different DSLs}). In state-machine models, those using \flexbe implemented hierarchical state machines, where the nesting level is above one, more often than models using \smach (11 out of the 14 \hsm models).
It appears that \chg{the developers in our sample projects kept the} models shallow, both when building state-machine and behavior-tree models.
Unfortunately, there is yet no study on the impact of a behavior-tree's depth on its complexity that could justify the developers' tendency to design shallow models. A justification could link the design decision to understandability, which would be similar to state-machine models. However, a proper investigation is required to confirm.

\looseness=-1
Now we turn to the use of different language constructs  in \chg{the sampled} models, starting with \bts.
In our behavior-tree sample, 66\% of the nodes in the dataset are leaf nodes ($1,228$ out of $1,850$ nodes), and 34\% are composite nodes. 
\Cref{tab:compositepct} summarizes the usage of composite nodes for each library and for the total population of behavior-tree models.

Most of the composite nodes in our projects are of type Sequence (56\% of total composite nodes) followed by Selector type (21\% of total composite nodes). The Parallel node concept, which generalizes Sequence and Selector, was not used much, only for 7\% of all composite nodes. This might explain why standard libraries for programming languages normally do not include generalizations of existential and universal quantifier functions (exists and forall)---these use cases seem to be rare.  The re-entrant nature of the behavior-tree \chg{DSLs} allows to use Parallel to wait until a minimum number of sub-trees succeed. This, however, does not seem to be used as often as we expected.

\looseness=-1 Decorators are used relatively rarely in \pytreesros\ models, where they constitute 6\% of the composite nodes. This is likely explained by the fact that it is easier to apply the transforming operations directly in the Python code, using Python syntax, than elevating it to behavior-tree abstract-syntax constructors.  The situation is different with \BTCPP, where decorators are used almost three times as often (19\% of composite nodes).  Here, the benefit of using the decorators (data-flow operators) of the \bt\ instead of C++ allows them to be visualized and monitored in the graphical editor (Groot). No such tool is available for \pytrees, so likely larger parts of the model may ``leak'' to the code. This demonstrates that users of \bts often have a choice of what is in scope and what is out of scope for a model. By model scope, we refer to the boundaries of a model which decide what is included in the model and what is left to be programmed outside of the model.
This is a property that clearly distinguishes general-purpose languages (GPLs), such as Python and C++, from DSLs. Yet, in our experience, the competence of deciding the model scope and the precision level is only rarely (with exceptions \cite{dslbook}) discussed in the teaching and research literature.

\begin{table}[t]

	\caption{ \label{tab:compositepct} Usage of different behavior-tree
		composite nodes to the total of them per library and in
		total for all models}
	\begin{tabularx}{\linewidth}{%
			>{\small\raggedright}p{4.9cm}
			>{\small}p{.4cm}
			>{\small}p{.4cm}
			>{\small}p{.4cm}
			>{\small}p{.4cm}}
	
	& \multicolumn{4}{c}{\small\sf ~~~composite nodes}\\
	
	\textsf{library}
	& \rotatebox{90}{\footnotesize Sequence}
	& \rotatebox{90}{\footnotesize Selector}
	& \rotatebox{90}{\footnotesize Decorator}
	& \rotatebox{90}{\footnotesize Parall\rlap{el}}
	\\\midrule
	
	\bfseries \BTCPP & 57\% & 19\% & 19\% & 6\%\\
	\bfseries \pytreesros & 53\% & 28\% & 6\% & 13\%\\

	\midrule
	
	share in population of all models
	& 56\%  & 21\% & 16\%& 7\%\\
	\bottomrule
	
\end{tabularx}
	\vspace{-3mm}
\end{table}

Finally, we observed that none of the models implement their own custom nodes. They rely on the extensibility of \bts using new custom operators (decorators). The available off-the-shelf decorators in \BTCPP and \pytreesros were sufficient to create a custom behavior to change an action/condition status, or to customize an action duration, e.g., wanting to execute an action without waiting, retrying an action $n$ times before given up, or repeating an action $n$ times. 

Going back to \cref{fig:figure-1}, the decorator (\texttt{RetryUntil\-Succesful}) was used to create a conditional loop that executes the sub-tree under (\texttt{ExplorationSeq}) $10$ times, unless the task fails, which is inverted into a success by an (\texttt{Inverter}). The developers were able to model this without having to use a while-loop or a similar general control-flow structure in the script.

\begin{observation}
	The studied behavior-tree \chg{DSLs} offer a range of concepts that are well suited to roboticists, but the usage of the offered concepts differs according to the \chg{GUI support} of the languages. 
	
\end{observation}

\looseness=-1
For state-machine \chg{DSLs}, we focused on the control-flow containers as a concept equivalent to the composite nodes in \bts. As mentioned in \cref{subsec:SMsematicsyntax}, \smach offers a range of control-flow containers, sharing only Concurrence behavior with \flexbe. For this reason Sequence and Iterator data is reported only for \smach. 
In general, control-flow constructs in our sampled state-machine models are used far less than in \bts. Only 12\% of the state-machine models used some kind of control-flow construct, while in the behavior-tree models all of them used some type of control-flow node. Concurrence is used in 11\% of all models, with models using \flexbe being responsible for the majority of this number (7 out of the 8 models used \flexbe). Although, Concurrence is used less in \smach than in \flexbe, it makes for 67\% of the total control-flow constructs used in \smach projects. Iterator is the second most used control-flow construct in the projects using \smach with 33\% of the total control-flow types. Sequence is never used in our sample.

It is not clear why the offered control-flow constructs are not as popular in \chg{our analyzed sample of} state-machine \chg{models} compared to the behavior-tree \chg{models}. It might be related to the syntax of state-machine models, or users not being familiar with them. By syntax we mean that \sms have a structure that is easy to understand, where states represent the status of a system and transitions facilitate the control-flow between states. Adding a layer of complexity to the model by using the control-flow constructs offered by the \chg{DSLs} is less needed. Also, it might be a similar case to the use of \pytreesros's Decorators, where it is easier to apply the pattern directly in Python syntax, than elevating it to the state-machine abstract syntax. 

\begin{table}[t]
\vspace{+.3cm}
	\caption{ \label{tab:reusestat} Frequency of reuse patterns per library for skill and task level}
	\begin{tabularx}{\linewidth}{
		p{1.8cm}|
		p{.5cm}
		p{.5cm}|
		p{.5cm}
		p{.5cm}|
		p{.5cm}
		p{.5cm}}
		\textsf{library}
		& \multicolumn{2}{p{1.8cm}|}{\textsf{intra-model referencing}}
		& \multicolumn{2}{p{1.9cm}|}{\textsf{clone-and-own}}
		& \multicolumn{2}{p{1.8cm}}{\textsf{inter-model referencing}}
		\\[+.2cm]
		& \textsf{\footnotesize skills} & \textsf{\footnotesize tasks}& \textsf{\footnotesize skills}  & \textsf{\footnotesize tasks} & \textsf{\footnotesize skills}  & \textsf{\footnotesize tasks} \\
		\midrule
		 \smach & - & - &  20\% & 13\% & 35\% & 8\%  \\
		 \flexbe & 1.3\% & -  & - & 16\% & 32\% & - \\[+.2cm]
		\textbf{total for state machines} & \bfseries 1.3\% & -  & \bfseries 20\% & \bfseries 29\%& \bfseries 67\% & \bfseries 8\% \\
		\midrule
		\BTCPP & 25\% & 13\% & 3\% & 39\% & 37\% & 3\%  \\
		\pytreesros & 5\% & 5\% & 13\% & 9\% & 19\% & - \\[+.2cm]
		\textbf{total for behavior trees} & \bfseries 30\% & \bfseries 18\% & \bfseries 16\% & \bfseries 48\%& \bfseries 56\% & \bfseries 3\%  \\
		\bottomrule               
	\end{tabularx}
	\vspace{-2mm}
\end{table}

\begin{figure*}[t]
	
	\includegraphics[
	clip,
	trim= 5mm 2mm 9mm 3mm,
	width=\linewidth
	]{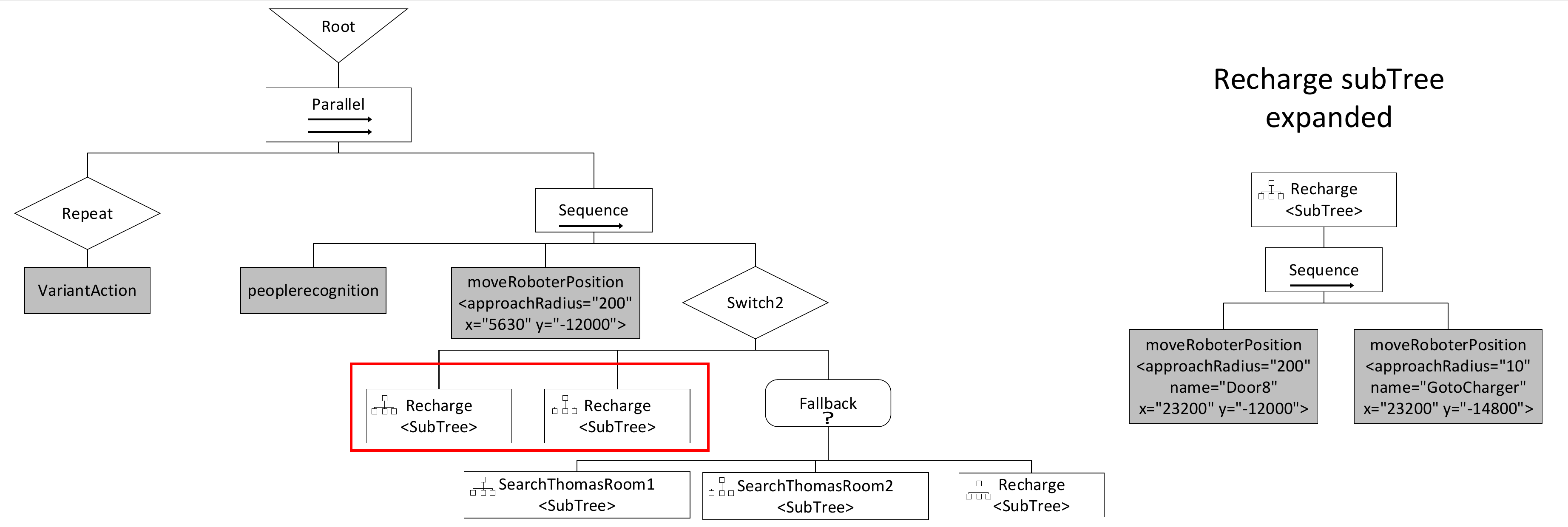}
	
	\caption[bundles model 2]{\Bt model of a retirement home robot from project bundles. The red box highlights an example of a intra-model referencing for a sub-tree Recharge (expanded on the right side). A legend is shown in \cref{fig:btsyntax}.}%
	\label{fig:bundles2}

\end{figure*}

\subsection{Reuse}
\label{subsec:reuse}

After presenting the core structural characteristics of our \chg{models in the sample}, we now shift to reuse as one of the major issues in robotics engineering software\,\cite{garcia2019robotics,garcia2020robotics, brugali2010component, brugali2009component, brugali2009software, nesnas2006claraty} and control architectures\,\cite{colledanchise2017behavior, kortenkamp2016robotic}. To facilitate reuse, a decomposition of a robotic mission or behavior into modular components should be supported by the modeling language. A behavior-modeling language supporting modular design---consequently, reusable components---is an important aspect for overcoming challenges of robotics control architectures and enhancing robotics software maintainability and quality\,\cite{brugali2010component, brugali2009component, colledanchise2017behavior, kortenkamp2016robotic}.

\looseness = -1 In our work, reuse refers to the ability to use already implemented skill code (also known as action in other contexts), or reusing code of a repeated task (composed of different skills) in the same model or across models in a project. We use the terms skill-level and task-level in the remainder to reference each. For skill-level and task-level code reuse, we observed three patterns of reuse \chg{in our sample}: \emph{intra-model referencing}, \emph{reuse by clone-and-own}\,\cite{dubinsky2013exploratory} and \emph{inter-model referencing}.

\emph{Intra-model referencing} was mostly used by \bt models. On the task-level, it is implemented  by creating a sub-tree for a repeated activity, then re-using it by reference in multiple branches in the model after passing new values to its parameters (usually by writing a new value to a blackboard). A skill-level implementation defines a leaf node as a function in the main model execution file, then reuses it by reference after passing new values to its parameters. 

The behavior-tree models often exploited \emph{intra-model referencing} on skill-level (30\% of the behavior-tree models), while on task-level it was only used in 18\% of the models.
This pattern was rarely used in state-machine models. Only one reused its skills by reference, and few reused on the task-level.
\Cref{fig:bundles2} shows an excerpt from one of our behavior-tree models, presenting the different tasks for a robot in a retirement home. The red box highlights an example of intra-model referencing, where the developer wrapped the moving activity in the sub-tree (\lstinline!Recharge!) and reused it in multiple parts of the model. An example on the skill-level is shown in the action \lstinline!moveRoboterPosition!, which was used in multiple parts in the model, only changing the values of parameters (name, approachRadius, x, and y). 

\emph{Reuse by clone-and-own} was the most used pattern for task-level reuse in the DSLs. In projects with multiple behavior-tree models, we observed that, when two \bts have the same activities, the similar parts (a sub-tree or the entire model) are reused after some minor changes, such as adding new nodes or removing old ones. Similarly in \sms, minor modifications are introduced to a model by adding or removing states to accommodate similar tasks. On the skill-level, clone-and-own was mainly implemented by minor modifications to the skill code.

\begin{figure*}[t]
	
	\includegraphics[
	clip,
	trim=7mm 4.5mm 11mm 5mm,
	width=\linewidth
	]{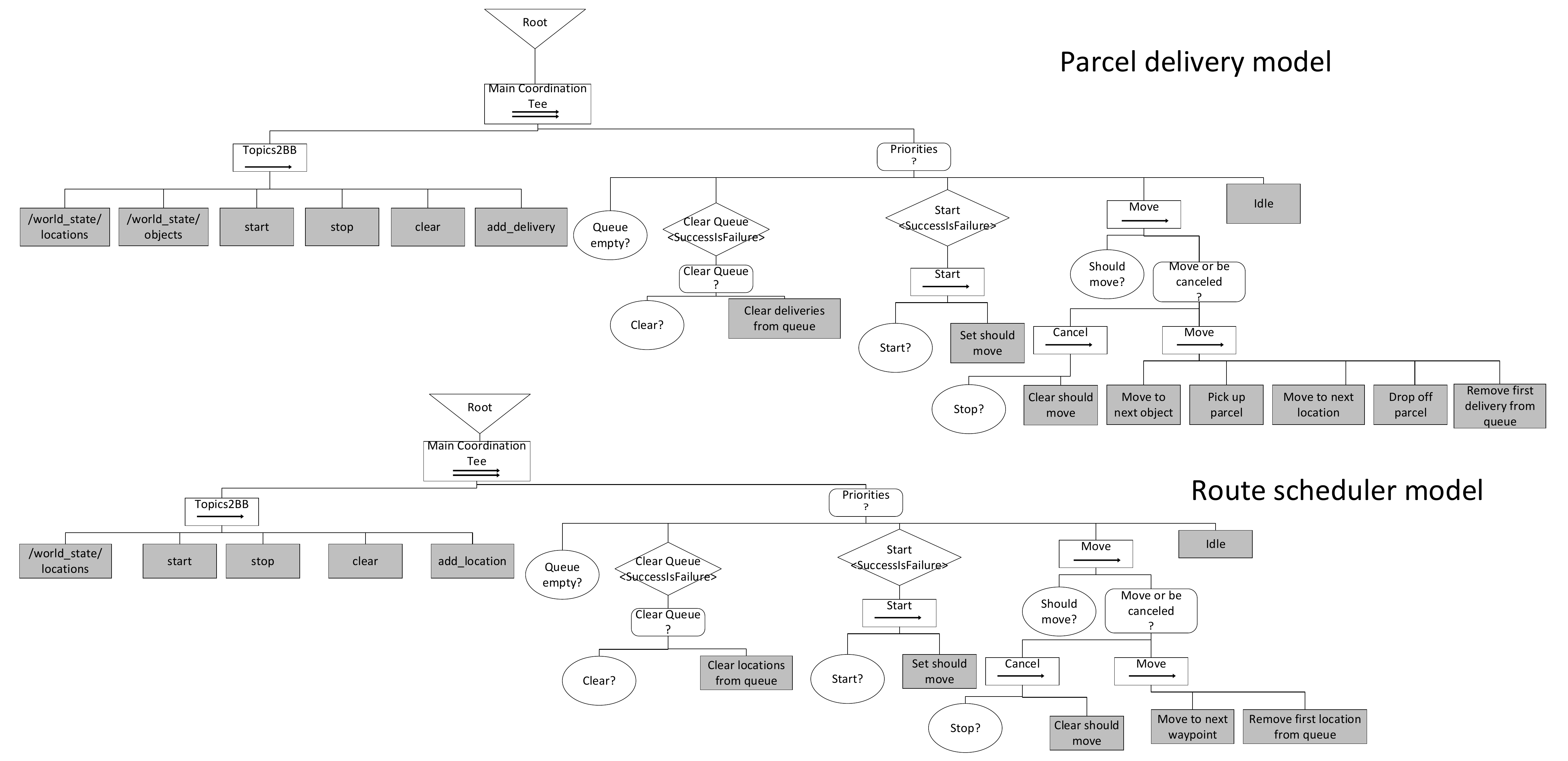}
	
	\caption[M1 and M2 BT model]{An example of clone-and-own referencing in \Bts from project Dyno. Each model belong to a different mission (M1) parcel delivery, and (M2) a route scheduler. Legend in \Cref{fig:btsyntax}.}%
	\label{fig:dynocombined}
	
	
\end{figure*}

Users of \bts and \sms favored using clone-and-own for task-level reuse \chg{in our sample}. It was the most used pattern for task-level reuse compared to the other two patterns (48\% of models in \bts and 29\% of models in \sms used it). Clone-and-own was not as popular for skill-level reuse as task-level reuse. It was the least-used pattern in behavior-tree models (16\% of the behavior-tree models) and second used pattern in state-machine models with only 20\% of models reusing skills using this pattern.  
The Dyno project in \cref{fig:dynocombined}, a drone-based parcel delivery project, includes two behavior-tree models: one for a parcel delivery mission (M1) and another one for a route scheduler mission (M2). These models are an example of task-level clone-and-own, where the developer reused the entire behavior-tree model for two different missions that share similar activities after proper modification depending on the mission.\footnote{The model can be found in full-size in the online appendix,\,\cite{appendix:online} in addition to the models of the other projects.}

\emph{Inter-model referencing} was the most used pattern for skill-level reuse in the DSLs.
Repeated skills were implemented as action nodes or associated with states in external files that were imported later in the main model execution file. This allows projects with multiple models having similar skills to reuse these skills in different parts of the model. Similar implementation paradigm was used for task-level by defining a repeated task as a behavior-tree or state-machine model in an external file then it is imported and invoked in the main model execution file. The repeated task is plugged as a sub-tree in the main \bt or as nested \sm in the main state-machine model. 

Inter-model referencing was the most used pattern for skill-level reuse among the different languages (used in 67\% of the state-machine models and 56\% of the behavior-tree models). For task-level, only two projects from the state-machine dataset used it and one project from the behavior-tree dataset (corresponding to 8\% of the state-machine models and 3\% of the behavior-tree models).

An interesting observation was that inter-model referencing through predefined skills offered by \flexbe developers as either an API \lstinline!flexbe_state! (\textsf{\href{https://github.com/FlexBE/generic_flexbe_states}{git\-hub.com/Flex\-BE/generic\_flexbe\_states}}), or as separate state libraries \lstinline!generic_flexbe_states! (\textsf{\href{https://github.com/FlexBE/generic_flexbe_states}{git\-hub.com/Flex\-BE/generic\_flexbe\_states}}) and \lstinline!Flexible Navigation! (\textsf{\href{https://github.com/FlexBE/flexible_navigation/tree/ros2-devel-alpha/flex_nav_flexbe_states}{github.com/FlexBE/flexible\_navigation/tree/ros2-devel-al\-pha/flex\_nav\_flexbe\_states}}). The skills were used in 73\% of the \flexbe models. Similar behavior was observed across projects of \smach and \pytreesros libraries by reusing \lstinline!ROS Navigation Stack! skills, such as \lstinline!move_base! package (in 55\% of the \smach projects and 60\% of the \pytreesros projects). The data shows that predefined skills APIs could facilitate reuse, and could be worth considering by developers to facilitate reuse.

In general, \chg{we observed in our sample that the} choice of reuse mechanism is influenced by the implementation of the \chg{DSL}, and whether one reuses skills or tasks. It was noticeable that inter-model referencing was favored for skill-level reuse among the different \chg{DSLs}, since defining a skill in an external file and importing it is a pretty straight-forward mechanism. For task-level reuse, inter-model referencing was the least-used method, while clone-and-own was more favored. Similar missions might share a set of skills, but the combination of them to form a task changes to accommodate specific needs, thus clone-and-own is the most used pattern for task-level reuse. Zooming into reuse \chg{for each DSL}, projects using \BTCPP have reported higher reuse frequency compared to projects using \pytreesros. These numbers can be related to \BTCPP having a dedicated XML format to express the behavior-tree model and Groot for visualizing the model, instead of intertwining the model with the code like in \pytreesros. Visualization and abstraction of the model could be an aspect to explore for pushing toward reusing in robotic missions.

\looseness=-1
We conjecture that the identified simple reuse mechanisms suffice for the identified robotics projects. It is less clear whether it would be useful to have safer and richer reuse mechanisms known from mainstream programming languages, including name-spacing and safe reuse contracts (interfaces), as they tend to be heavyweight for users to learn and use. More research is needed to determine whether sufficiently lightweight and safe reuse mechanism could be realized.


\section{Threats to Validity}
\label{sec:validity}
\looseness=-5 \textbf{Internal Validity.}  We provide a dataset of behavior-tree and state-machine models in robotics. Our dataset could include projects from courses or tutorials, and we could have missed projects with behavior-tree or state-machine models expressed in our libraries.  To mitigate that, we applied a multi-step filtration mechanism that was adopted to our observations during the mining process. We derived our filtration steps from our data observations. After each step, we checked randomly a sample of the results to see if we needed to adopt the filtration mechanism.

Another threat that could affect the results, is possible errors in our Python scripts calculating the model metrics. In \bts, as a form of quality check, we manually counted node types and checked the script results against them. We excluded commented parts and unused node types in the behavior-tree codes. In \sms, we rechecked the automatic counting results by going through the models.

\medskip

\noindent{\textbf{External Validity.}} 
A threat is the generalizability of our quantitative results and the identified reuse patterns, since we used a random sample of the mined models. We mitigated that by aggregating the models into pools of sizes according to their data distribution, then randomly sampling our dataset. This way, we made sure to capture different projects with different model sizes. In addition, our project domains, presented in \cref{fig:projectsdomains}, show that our random sample covers different domain categories.

\looseness=-1
The list of identified \chg{open-source} robotics projects might be missing examples from Bitbucket and GitLab. Both platforms are used in the robotics community; however, they do not provide a code search API, which makes it difficult to conduct a code-level search. We conducted a less precise query in Bitbucket and GitLab using \emph{behavior trees} and \emph{state machine} as search terms in the web interface. However, for \bts, we could not identify any real robotics projects from that search. For \sms, GitLab returned results for which it was hard to check if any of the subject libraries were used. We favored using GitHub for the flexibility of the provided APIs, for their integration with Python and other languages for mining, and because it is a platform widely used by developers to publish open-source projects.

We have only considered projects using Python and C++ libraries with ROS support, while there might be other open-source robotics projects out there. We acknowledge that limiting our search to ROS-supported languages might have resulted in missing other robotics projects. However, we focused on the two dominant languages in ROS, assuming that this is the most representative framework for \chg{open-source} robotics.
\section{Related Work}
\label{sec:relatedwork}

In the literature, researchers compare---theoretically---\bts to
popular control architectures in robotics, such as finite state
machines, the subsumption architecture, and decision trees. They show
that \bts generalize them \cite{colledanchise2018behavior,
  colledanchise2016behavior, colledanchise2017behavior}. They also
discuss advantages and disadvantages of each control architecture
based on design properties that are important for robotic control
architectures. Colledanchise et
al.\,\cite{colledanchise2016advantages} describe the drawbacks of
using state machines for multi-robot scenarios in comparison to \bts
using illustrative examples. In another work, Colledanchise et
al.\,\cite{colledanchise2021implementation} illustrate how to
represent different kinds of behaviors from state machines in behavior
trees, provide available software libraries and show what kind of
representations they support. Modularity, reactiveness expressiveness,
and readability are some of the properties of \bts that are studied in
a formal-theoretical manner and compared to state machines by Biggar
et al.\,\cite{biggar2021expressiveness}.  Colledanchise and
Natale\,\cite{colledanchise2021implementation} highlight that
behavior-tree tools are less mature than tools available for state
machines; however, their findings do not build on an analysis of
available tools. In our work, we observed good language-design
practices for the studied behavior-tree and state-machine DSLs
(tools), and some sub-optimal ones. However, we cannot conclude that
the studied state-machine DSLs were better designed than the
behavior-tree DSLs. Each language design had its own advantages and
disadvantages.
	
The related work only bases its findings to a very limited extent on
software engineering practice in real-world robotics projects. In this
work, we scope the analysis to behavior-tree DSLs in robotics from the
software-language perspective, and we compare them to state-machine
DSLs, which has not been done so far in the literature. Compared to
previous work, we shift the comparison of \bts and \sms from the
theoretical perspective to the real world. We do not report on the
advantages, or disadvantages, of one model over the other. We report
on the similarities and differences between the supporting languages
of both models in robotics. We report on the state-of-practice by
analyzing the usage of behavior-tree and state-machine DSLs in
open-source projects.  In our work, we focus on the structural
properties and reuse-of-code in open-source projects, using the
identified languages. Consequently, our work complements the related
work. In addition, we provide behavior-tree and state-machine models
as a community dataset, which has not been done so far. The
dataset can be used for further research.

 
\section{Conclusion}%
\label{sec:conclusion}
\looseness=-1 We presented a study of behavior-tree and state-machine DSLs, and of their use in \chg{open-source ROS} robotics appli\-cations. We systematically compared the concepts available in popular behavior-tree \chg{DSLs} and contrasted them with well-established \chg{DSLs} for \chg{reactive modeling}---the language of state machines. We mined open-source projects from code repositories and extracted the behavior-tree and state-machine models from their codebases. We analyzed the structure of a sample of these models, as well as how they use the concepts and how they reuse model code. We contribute a dataset of models in an online appendix\,\cite{appendix:online}, together with scripts and additional data.

\subsubsection*{Results and Implications}
\looseness=-1
The paper displays popular DSLs designed \emph{outside} of the language-engineering community for the vibrant domain of robotics.  We believe that studying modeling and language-engineering practices is beneficial for both communities, as it helps to improve language-engineering methods and tools, as well as to improve the actual practices and languages.

\looseness=-1
Our analysis showed that both groups of DSLs follow good design practices and even obtain some improvement by learning from each other.
As a start, having a GUI that provides constructing, editing, and runtime monitoring of models is a good practice. A GUI supports visualization and building abstractions. Both \flexbe and \BTCPP provide similar tools, which is reflected in their users reusing code more often compared to \smach and \pytreesros.

\looseness=-1
In addition, our results illustrate that many of the modeling and language-engineering methods are relevant in practice.
Developing DSLs in a rather pragmatic way, without hundreds of pages of specification documents and with a basic, but extensible meta-model, or even without an explicitly defined meta-model seems to be successful. Such a strategy seems to attract practitioners not trained in language and modeling technology, allowing practitioners who come from lower-level programming paradigms to raise the level of abstraction and effectively implement missions of robots in higher-level representations.

\looseness=-1
Concepts from software engineering are adopted in the design of the studied languages. \BTCPP is built with separation-of-concerns and separation-of-roles in mind\,\cite{mood2be}, making it possible to decouple models from the robotic system for visualization and testing. During model analysis, we managed to load models easily in Groot without the need to setup ROS or fulfill other requirements of the robotic projects. Also, inspecting the XML representation of the models was relatively easy. These are two features that we appreciated during the analysis. Unfortunately the state-machine DSLs and \pytreesros lack an easily accessible visualization tools, which required us to build custom ways to inspect models.

\bchg{Not all the studied languages support the \chg{separation-of-roles} to the same extent. \flexbe uses code generation from model-based design for boilerplate code, \chg{while \BTCPP requires custom code in an internal DSL, which requires programming skills}. Although \chg{code-generation}, and external DSLs, are more constraining than the flexibility offered by programming in a general-purpose language with an internal DSL, support from code generators for creating repetitive parts seems valuable, as more aspects of modeling can be performed by domain experts. Furthermore, syntactic and semantic checks of model consistency are another natural feature of external DSLs exploited in \flexbe (these can be realized in internal DSLs as well, albeit at a higher cost). Still, the users of all the studied languages need to be familiar with language-oriented programming (meta-programming). This is despite the fact that most robot developers are not trained in this field. This poses an interesting challenge for the language design community to come up with designs that would reduce the friction of integrating models with the rest of the robot architecture.}


\looseness=-1
Promoting and adopting mode-driven engineering practices has been on the rise in the robotics community to improve the reusability and maintainability of systems\,\cite{de2021survey,gherardi2014modeling,schlegel2015model,wigand2017modularization, brugali2009component,brugali2010component,garcia2019robotics, robotics2017robotics,garcia.ea:2023:robotvar}. Through our analysis, we observed common practices across the different languages that are adopted from model-driven engineering specifically, and from software engineering generally, which seem well received by practitioners. In addition, improvements to these behavior-modeling languages could be adopted and evaluated in the robotics community.

\subsubsection*{Future Work}
\bchg{In the future, we would like to build on top of current observations and conduct studies with users in the loop. Our goal would be to compare the usability, comprehension and expressiveness of the studied \chg{behavior-modeling} languages, and the needed improvements. Specifically, a valuable study would be to systematically explore the realization of mission requirements (e.g., specified in natural textual language) in different behavior-modeling languages and measuring expressiveness, but also non-functional aspects such as succinctness and intuitiveness of the resulting model.}

\looseness=-1
Also, we would like to provide good design guidelines for behavior-modeling languages that promote MDE practices but are compatible with the current state-of-practice. Finally, during our analysis we noticed that most projects lack mission specifications.  We want to use our dataset to develop automatic tools to generate mission specifications in natural language from existing behavior-tree models. By achieving that we help in creating automated re-engineering tools for legacy projects that could be reused by other practitioners instead of starting from scratch.

\looseness=-1
\bchg{Another valuable piece of future work would be to gather empirical data on the use of behavior-modeling languages in robotics projects in general, without the limitation to behavior-tree and state-machine DSLs. While these are, as we argue, the most prominent kinds of modeling languages used, we believe that other languages are also used. However, identifying other languages as used in open-source projects is not trivial and would constitute one or even multiple future studies.}



\section*{Acknowledgments}
This work was partially supported by the Wallenberg AI, Autonomous Systems and Software Program (WASP) funded by the Knut and Alice Wallenberg Foundation.

\ifCLASSOPTIONcaptionsoff
  \newpage
\fi



\bibliographystyle{IEEEtran}

\bibliography{main}

\begin{thebibliography}{100}
\providecommand{\url}[1]{#1}
\csname url@samestyle\endcsname
\providecommand{\newblock}{\relax}
\providecommand{\bibinfo}[2]{#2}
\providecommand{\BIBentrySTDinterwordspacing}{\spaceskip=0pt\relax}
\providecommand{\BIBentryALTinterwordstretchfactor}{4}
\providecommand{\BIBentryALTinterwordspacing}{\spaceskip=\fontdimen2\font plus
\BIBentryALTinterwordstretchfactor\fontdimen3\font minus
  \fontdimen4\font\relax}
\providecommand{\BIBforeignlanguage}[2]{{%
\expandafter\ifx\csname l@#1\endcsname\relax
\typeout{** WARNING: IEEEtran.bst: No hyphenation pattern has been}%
\typeout{** loaded for the language `#1'. Using the pattern for}%
\typeout{** the default language instead.}%
\else
\language=\csname l@#1\endcsname
\fi
#2}}
\providecommand{\BIBdecl}{\relax}
\BIBdecl

\bibitem{garcia2020robotics}
S.~Garc{\'\i}a, D.~Str{\"u}ber, D.~Brugali, T.~Berger, and P.~Pelliccione,
  ``Robotics software engineering: A perspective from the service robotics
  domain,'' in \emph{FSE}, 2020.

\bibitem{menghi.ea:2019:tse}
\BIBentryALTinterwordspacing
C.~Menghi, C.~Tsigkanos, P.~Pelliccione, C.~Ghezzi, and T.~Berger,
  ``Specification patterns for robotic missions,'' \emph{{IEEE} Trans. Software
  Eng.}, vol.~47, no.~10, pp. 2208--2224, 2021. [Online]. Available:
  \url{https://doi.org/10.1109/TSE.2019.2945329}
\BIBentrySTDinterwordspacing

\bibitem{dragule2021bookchapter}
S.~Dragule, S.~Garcia, T.~Berger, and P.~Pelliccione, ``Languages for
  specifying missions of robotic applications,'' in \emph{Software Engineering
  for Robotics}, A.~Cavalcanti, B.~D. ad~Rob~Hierons, J.~Timmis, and
  J.~Woodcock, Eds.\hskip 1em plus 0.5em minus 0.4em\relax Springer, 2021.

\bibitem{michaud2016behavior}
F.~Michaud and M.~Nicolescu, ``Behavior-based systems,'' in \emph{Springer
  handbook of robotics}.\hskip 1em plus 0.5em minus 0.4em\relax Springer, 2016,
  pp. 307--328.

\bibitem{kortenkamp2016robotic}
D.~Kortenkamp, R.~Simmons, and D.~Brugali, ``Robotic systems architectures and
  programming,'' in \emph{Springer handbook of robotics}.\hskip 1em plus 0.5em
  minus 0.4em\relax Springer, 2016, pp. 283--306.

\bibitem{brugali2010component}
D.~Brugali and A.~Shakhimardanov, ``Component-based robotic engineering (part
  ii),'' \emph{IEEE Robotics \& Automation Magazine}, vol.~17, no.~1, pp.
  100--112, 2010.

\bibitem{colledanchise2018behavior}
M.~Colledanchise and P.~{\"O}gren, \emph{Behavior Trees in Robotics and Al: An
  Introduction}.\hskip 1em plus 0.5em minus 0.4em\relax CRC Press, 2018.

\bibitem{chen2018development}
J.~Chen and D.~Shi, ``Development and composition of robot architecture in
  dynamic environment,'' in \emph{RCAE}, 2018.

\bibitem{heckel2010representational}
F.~W. Heckel, G.~M. Youngblood, and N.~S. Ketkar, ``Representational complexity
  of reactive agents,'' in \emph{CIG}, 2010.

\bibitem{isla2005gdc}
\BIBentryALTinterwordspacing
D.~Isla, ``Handling complexity in the {Halo 2 AI},'' \emph{GDC 2005
  Proceedings}, 2005. [Online]. Available:
  \url{https://www.gamasutra.com/view/feature/130663/gdc_2005_proceeding_handling_.php?page=2}
\BIBentrySTDinterwordspacing

\bibitem{Mcquillan2015}
K.~Mcquillan, ``A survey of behaviour trees and their applications for game
  {AI},'' 2015, course CP5330 final report, James Cook University.

\bibitem{biggar2021expressiveness}
O.~Biggar, M.~Zamani, and I.~Shames, ``An expressiveness hierarchy of behavior
  trees and related architectures,'' \emph{IEEE Robotics and Automation
  Letters}, vol.~6, no.~3, pp. 5397--5404, 2021.

\bibitem{iovino2022programming}
M.~Iovino, J.~F{\"o}rster, P.~Falco, J.~J. Chung, R.~Siegwart, and C.~Smith,
  ``On the programming effort required to generate behavior trees and finite
  state machines for robotic applications,'' \emph{arXiv preprint
  arXiv:2209.07392}, 2022.

\bibitem{bagnell2012integrated}
J.~A. Bagnell, F.~Cavalcanti, L.~Cui, T.~Galluzzo, M.~Hebert, M.~Kazemi,
  M.~Klingensmith, J.~Libby, T.~Y. Liu, N.~Pollard \emph{et~al.}, ``An
  integrated system for autonomous robotics manipulation,'' in \emph{IROS},
  2012.

\bibitem{colledanchise2016advantages}
M.~Colledanchise, A.~Marzinotto, D.~V. Dimarogonas, and P.~{\"O}gren, ``The
  advantages of using behavior trees in multi robot systems,'' in \emph{47th
  International Symposium on Robotics ({ISR})}, 2016.

\bibitem{colledanchise2016behavior}
M.~Colledanchise and P.~{\"O}gren, ``How behavior trees modularize hybrid
  control systems and generalize sequential behavior compositions, the
  subsumption architecture, and decision trees,'' \emph{IEEE Transactions on
  robotics}, vol.~33, no.~2, pp. 372--389, 2016.

\bibitem{colledanchise2018learning}
M.~Colledanchise, R.~Parasuraman, and P.~{\"O}gren, ``Learning of behavior
  trees for autonomous agents,'' \emph{IEEE Transactions on Games}, vol.~11,
  no.~2, pp. 183--189, 2018.

\bibitem{marzinotto2014towards}
A.~Marzinotto, M.~Colledanchise, C.~Smith, and P.~{\"O}gren, ``Towards a
  unified behavior trees framework for robot control,'' in \emph{ICRA}, 2014.

\bibitem{ogren2012increasing}
P.~{\"{O}}gren, ``Increasing modularity of uav control systems using computer
  game behavior trees,'' in \emph{{AIAA} Guidance, Navigation, and Control
  Conference ({GNC})}, 2012.

\bibitem{rovida2017extended}
F.~Rovida, B.~Grossmann, and V.~Kr{\"u}ger, ``Extended behavior trees for quick
  definition of flexible robotic tasks,'' in \emph{IROS}, 2017.

\bibitem{garcia2019high}
S.~Garc{\'\i}a, P.~Pelliccione, C.~Menghi, T.~Berger, and T.~Bures,
  ``High-level mission specification for multiple robots,'' in \emph{SLE},
  2019.

\bibitem{klockner2013interfacing}
A.~Kl{\"o}ckner, ``Interfacing behavior trees with the world using description
  logic,'' in \emph{{AIAA} Guidance, Navigation, and Control Conference
  ({GNC})}, 2013.

\bibitem{dslbook}
A.~Wasowski and T.~Berger, \emph{Domain-specific Languages: Effective Modeling,
  Automation, and Reuse}.\hskip 1em plus 0.5em minus 0.4em\relax Springer,
  2023, http://dsl.design.

\bibitem{de2021survey}
E.~de~Ara{\'u}jo~Silva, E.~Valentin, J.~R.~H. Carvalho, and
  R.~da~Silva~Barreto, ``A survey of model driven engineering in robotics,''
  \emph{Journal of Computer Languages}, vol.~62, p. 101021, 2021.

\bibitem{casalaro2021model}
G.~L. Casalaro, G.~Cattivera, F.~Ciccozzi, I.~Malavolta, A.~Wortmann, and
  P.~Pelliccione, ``Model-driven engineering for mobile robotic systems: a
  systematic mapping study,'' \emph{Software and Systems Modeling}, pp. 1--31,
  2021.

\bibitem{nordmann2014survey}
A.~Nordmann, N.~Hochgeschwender, and S.~Wrede, ``A survey on domain-specific
  languages in robotics,'' in \emph{SIMPAR}, 2014.

\bibitem{quigley2009ros}
M.~Quigley, K.~Conley, B.~Gerkey, J.~Faust, T.~Foote, J.~Leibs, R.~Wheeler,
  A.~Y. Ng \emph{et~al.}, ``{ROS}: an open-source robot operating system,'' in
  \emph{ICRA workshop on open source software}, 2009.

\bibitem{ghzouli2020behavior}
R.~Ghzouli, T.~Berger, E.~B. Johnsen, S.~Dragule, and A.~Wasowski, ``Behavior
  trees in action: a study of robotics applications,'' in \emph{SLE}, 2020.

\bibitem{bencomo14modelsatruntime}
N.~Bencomo, R.~B. France, B.~H.~C. Cheng, and U.~A{\ss}mann, Eds.,
  \emph{Models@run.time---Foundations, Applications, and Roadmaps}, vol.
  8378.\hskip 1em plus 0.5em minus 0.4em\relax Springer, 2014.

\bibitem{blair2009models}
G.~Blair, N.~Bencomo, and R.~B. France, ``Models@run.time,'' \emph{IEEE
  Computer}, vol.~42, no.~10, pp. 22--27, 2009.

\bibitem{appendix:online}
\BIBentryALTinterwordspacing
``Online appendix,'' \url{https://bitbucket.org/easelab/behaviortrees}, 2020.
  [Online]. Available: \url{https://doi.org/10.5281/zenodo.7515222}
\BIBentrySTDinterwordspacing

\bibitem{colledanchise2017behavior}
M.~Colledanchise, ``Behavior trees in robotics,'' Ph.D. dissertation, KTH Royal
  Institute of Technology, 2017.

\bibitem{dragule2021survey}
S.~Dragule, T.~Berger, C.~Menghi, and P.~Pelliccione, ``A survey on the design
  space of end-user-oriented languages for specifying robotic missions,''
  \emph{Software and Systems Modeling}, vol.~20, pp. 1123--1158, 2021.

\bibitem{colledanchise2016behavior2}
M.~Colledanchise and P.~{\"O}gren, ``How behavior trees generalize the
  teleo-reactive paradigm and and-or-trees,'' in \emph{IROS}, 2016.

\bibitem{conner2017flexible}
D.~C. Conner and J.~Willis, ``Flexible navigation: Finite state machine-based
  integrated navigation and control for ros enabled robots,'' in
  \emph{SoutheastCon}, 2017.

\bibitem{macenski2020marathon}
S.~Macenski, F.~Mart{\'\i}n, R.~White, and J.~G. Clavero, ``The {M}arathon 2: A
  navigation system,'' in \emph{IROS}, 2020.

\bibitem{zutell2022ros}
J.~M. Zutell, D.~C. Conner, and P.~Schillinger, ``Ros 2-based flexible behavior
  engine for flexible navigation,'' in \emph{SoutheastCon}, 2022.

\bibitem{garcia2020icsedemo}
S.~Garcia, P.~Pelliccione, C.~Menghi, T.~Berger, and T.~Bures, ``Promise:
  High-level mission specification for multiple robots,'' in \emph{ICSE
  Demonstrations}, 2020.

\bibitem{kato2018autoware}
S.~Kato, S.~Tokunaga, Y.~Maruyama, S.~Maeda, M.~Hirabayashi, Y.~Kitsukawa,
  A.~Monrroy, T.~Ando, Y.~Fujii, and T.~Azumi, ``Autoware on board: Enabling
  autonomous vehicles with embedded systems,'' in \emph{ICCPS}, 2018.

\bibitem{Dosovitskiy17}
A.~Dosovitskiy, G.~Ros, F.~Codevilla, A.~Lopez, and V.~Koltun, ``{CARLA}: {An}
  open urban driving simulator,'' in \emph{Proceedings of the 1st Annual
  Conference on Robot Learning}, 2017, pp. 1--16.

\bibitem{carlarunner}
``{ScenarioRunner} for {CARLA} documentation,''
  \url{https://carla-scenariorunner.readthedocs.io}, 2022.

\bibitem{iovino2022survey}
M.~Iovino, E.~Scukins, J.~Styrud, P.~{\"O}gren, and C.~Smith, ``A survey of
  behavior trees in robotics and {AI},'' \emph{Robotics and Autonomous
  Systems}, vol. 154, p. 104096, 2022.

\bibitem{colledanchise2019analysis}
M.~Colledanchise and L.~Natale, ``Analysis and exploitation of synchronized
  parallel executions in behavior trees,'' \emph{arXiv preprint
  arXiv:1908.01539}, 2019.

\bibitem{millington2009artificial}
I.~Millington and J.~Funge, \emph{Artificial intelligence for games}.\hskip 1em
  plus 0.5em minus 0.4em\relax CRC Press, 2009.

\bibitem{crane2007uml}
M.~L. Crane and J.~Dingel, ``{UML} vs. classical vs. rhapsody statecharts: not
  all models are created equal,'' \emph{Software \& Systems Modeling}, vol.~6,
  no.~4, pp. 415--435, 2007.

\bibitem{von1994comparison}
M.~von~der Beeck, ``A comparison of statecharts variants,'' in \emph{Third
  International Symposium on Formal Techniques in Real-Time and Fault-Tolerant
  Systems}, 1994.

\bibitem{uml2017}
\BIBentryALTinterwordspacing
{Object~Management~Group}, ``{OMG} unified modeling language 2.5.1,'' 2017.
  [Online]. Available: \url{https://www.omg.org/spec/UML/}
\BIBentrySTDinterwordspacing

\bibitem{DOUGLASS2011257}
B.~P. Douglass, ``Chapter 5 - design patterns for state machines,'' in
  \emph{Design Patterns for Embedded Systems in C}, B.~P. Douglass, Ed.\hskip
  1em plus 0.5em minus 0.4em\relax Newnes, 2011, pp. 257--356.

\bibitem{samek2009crash}
M.~Samek, ``A crash course in {UML} state machines,'' \emph{Quantum Leaps,
  LLC}, 2009.

\bibitem{kubatova2013petri}
H.~Kub{\'a}tov{\'a}, K.~Richta, and T.~Richta, ``Petri nets versus {UML} state
  machines,'' in \emph{Proc. SDOT}, 2013, pp. 53--59.

\bibitem{Hajji2022}
B.~Hajji, A.~Mellit, and L.~Bouselham, \emph{Finite State Machines}.\hskip 1em
  plus 0.5em minus 0.4em\relax Singapore: Springer Singapore, 2022, pp.
  175--205.

\bibitem{harel1987statecharts}
D.~Harel, ``Statecharts: A visual formalism for complex systems,''
  \emph{Science of computer programming}, vol.~8, no.~3, pp. 231--274, 1987.

\bibitem{luttgen2000compositional}
G.~L{\"u}ttgen, M.~Von~der Beeck, and R.~Cleaveland, ``A compositional approach
  to statecharts semantics,'' \emph{ACM SIGSOFT Software Engineering Notes},
  vol.~25, no.~6, pp. 120--129, 2000.

\bibitem{pytreedoc}
D.~Stonier, N.~Usmani, and M.~Staniaszek, ``Py {Trees} library documentation,''
  \url{https://py-trees.readthedocs.io/en/devel/}, 2020.

\bibitem{btscppdoc}
D.~Faconti and M.~Colledanchise, ``{BehaviorTree.CPP} library documentation,''
  \url{https://www.behaviortree.dev}, 2018.

\bibitem{bohren2010smach}
J.~Bohren and S.~Cousins, ``The {SMACH} high-level executive [ros news],''
  \emph{IEEE Robotics \& Automation Magazine}, vol.~17, no.~4, pp. 18--20,
  2010.

\bibitem{schillinger2016flexbe}
P.~Schillinger, S.~Kohlbrecher, and O.~von Stryk, ``Human-robot collaborative
  high-level control with an application to rescue robotics,'' in \emph{IEEE
  International Conference on Robotics and Automation ({ICRA})}, May 2016.

\bibitem{schillinger2015approach}
P.~Schillinger, ``An approach for runtime-modifiable behavior control of
  humanoid rescue robots,'' \emph{Technische Universitat Darmstadt}, 2015.

\bibitem{kohlbrecher2016comprehensive}
S.~Kohlbrecher, A.~Stumpf, A.~Romay, P.~Schillinger, O.~Von~Stryk, and D.~C.
  Conner, ``A comprehensive software framework for complex locomotion and
  manipulation tasks applicable to different types of humanoid robots,''
  \emph{Frontiers in Robotics and AI}, vol.~3, p.~31, 2016.

\bibitem{smachdoc}
J.~Bohren and S.~Cousins, ``{SMACH} library documentation,''
  \url{http://wiki.ros.org/smach/Documentation}, 2010.

\bibitem{flexbedoc}
P.~Schillinger, ``{FlexBE} library documentation,''
  \url{http://philserver.bplaced.net/fbe/documentation.php}, 2016.

\bibitem{pytreerosdoc}
D.~Stonier, N.~Usmani, and M.~Staniaszek, ``{Py Trees ROS} library
  documentation,'' \url{https://py-trees-ros.readthedocs.io/en/devel/}, 2020.

\bibitem{btscpptutorial}
D.~Faconti and M.~Colledanchise, ``{BehaviorTree.CPP} library tutorials,''
  \url{https://www.behaviortree.dev/tutorial_01_first_tree/}, 2018.

\bibitem{pytreerostutorial}
D.~Stonier, N.~Usmani, and M.~Staniaszek, ``{Py Trees ROS} library tutorials,''
  \url{https://py-trees-ros-tutorials.readthedocs.io/en/devel/tutorials.html},
  2020.

\bibitem{flexbetutorial}
P.~Schillinger, ``{FlexBE} library tutorials,''
  \url{http://wiki.ros.org/flexbe/Tutorials}, 2021.

\bibitem{smachtutorial}
J.~Bohren and S.~Cousins, ``{SMACH} library tutorials,''
  \url{http://wiki.ros.org/smach/Tutorials}, 2021.

\bibitem{mood2be}
\BIBentryALTinterwordspacing
D.~Faconti, ``{MOOD2Be}: Models and tools to design robotic behaviors,''
  European Union's Horizon 2020 Research and Innovation Programme, 2019.
  [Online]. Available:
  \url{https://github.com/BehaviorTree/BehaviorTree.CPP/blob/master/MOOD2Be_final_report.pdf}
\BIBentrySTDinterwordspacing

\bibitem{colledanchise2021implementation}
M.~Colledanchise and L.~Natale, ``On the implementation of behavior trees in
  robotics,'' \emph{IEEE Robotics and Automation Letters}, vol.~6, no.~3, pp.
  5929--5936, 2021.

\bibitem{romay2017collaborative}
A.~Romay, S.~Kohlbrecher, A.~Stumpf, O.~von Stryk, S.~Maniatopoulos,
  H.~Kress-Gazit, P.~Schillinger, and D.~C. Conner, ``Collaborative autonomy
  between high-level behaviors and human operators for remote manipulation
  tasks using different humanoid robots,'' \emph{Journal of Field Robotics},
  vol.~34, no.~2, pp. 333--358, 2017.

\bibitem{der2015approach}
P.~Schillinger, ``An approach for runtime-modifiable behavior control of
  humanoid rescue robots,'' Master's thesis, TU Darmstadt, 2015,
  \url{https://www.sim.informatik.tu-darmstadt.de/publ/da/2015_Schillinger_MA.pdf}.

\bibitem{almarzouq2020mining}
M.~AlMarzouq, A.~AlZaidan, and J.~AlDallal, ``Mining {GitHub} for research and
  education: challenges and opportunities,'' \emph{International Journal of Web
  Information Systems}, 2020.

\bibitem{malavolta2021mining}
I.~Malavolta, G.~A. Lewis, B.~Schmerl, P.~Lago, and D.~Garlan, ``Mining
  guidelines for architecting robotics software,'' \emph{Journal of Systems and
  Software}, vol. 178, p. 110969, 2021.

\bibitem{robles2017extensive}
G.~Robles, T.~Ho-Quang, R.~Hebig, M.~R. Chaudron, and M.~A. Fernandez, ``An
  extensive dataset of {UML} models in {GitHub},'' in \emph{2017 IEEE/ACM 14th
  International Conference on Mining Software Repositories (MSR)}.\hskip 1em
  plus 0.5em minus 0.4em\relax IEEE, 2017, pp. 519--522.

\bibitem{chidamber1994metrics}
S.~R. Chidamber and C.~F. Kemerer, ``A metrics suite for object oriented
  design,'' \emph{IEEE Transactions on software engineering}, vol.~20, no.~6,
  pp. 476--493, 1994.

\bibitem{berger2014towards}
T.~Berger and J.~Guo, ``Towards system analysis with variability model
  metrics,'' in \emph{VaMoS}, 2014.

\bibitem{cruz2007using}
J.~A. Cruz-Lemus, M.~Genero, and M.~Piattini, ``Using controlled experiments
  for validating {UML} statechart diagrams measures,'' in \emph{Software
  Process and Product Measurement}.\hskip 1em plus 0.5em minus 0.4em\relax
  Springer, 2007, pp. 129--138.

\bibitem{cruz2005metrics}
------, ``Metrics for {UML} statechart diagrams,'' in \emph{Metrics for
  Software Conceptual Models}, 2005, pp. 237--272.

\bibitem{van2019lightweight}
R.~van Tonder and C.~Le~Goues, ``Lightweight multi-language syntax
  transformation with parser parser combinators,'' in \emph{Proceedings of the
  40th ACM SIGPLAN Conference on Programming Language Design and
  Implementation}, 2019, pp. 363--378.

\bibitem{garcia.ea:2023:robotvar}
S.~Garc{\'\i}a, D.~Str{\"u}ber, D.~Brugali, A.~Di~Fava, P.~Pelliccione, and
  T.~Berger, ``Software variability in service robotics,'' \emph{Empirical
  Software Engineering}, vol.~28, no.~2, pp. 1--67, 2023.

\bibitem{garcia2019robotics}
S.~Garcia, D.~Strueber, D.~Brugali, A.~D. Fava, P.~Schillinger, P.~Pelliccione,
  and T.~Berger, ``Variability modeling of service robots: Experiences and
  challenges,'' in \emph{VaMoS}, 2019.

\bibitem{brugali2009software}
D.~Brugali and E.~Prassler, ``Software engineering for robotics [from the guest
  editors],'' \emph{IEEE Robotics \& Automation Magazine}, vol.~16, no.~1, pp.
  9--15, 2009.

\bibitem{nesnas2006claraty}
I.~A. Nesnas, R.~Simmons, D.~Gaines, C.~Kunz, A.~Diaz-Calderon, T.~Estlin,
  R.~Madison, J.~Guineau, M.~McHenry, I.-H. Shu \emph{et~al.}, ``{CLARAty}:
  Challenges and steps toward reusable robotic software,'' \emph{International
  Journal of Advanced Robotic Systems}, vol.~3, no.~1, p.~5, 2006.

\bibitem{rosbtdoc}
M.~Colledanchise, ``{BT++} library documentation,''
  \url{https://github.com/miccol/ROS-Behavior-Tree/blob/master/BTUserManual.pdf},
  2017.

\bibitem{skiros2}
F.~Rovida, ``{SkiROS2} library documentation,''
  \url{https://github.com/RVMI/skiros2/wiki}, 2020.

\bibitem{smaccdoc}
{RoboSoft AI}, ``{SMACC} library documentation,'' \url{https://smacc.dev/},
  2018.

\bibitem{steinbrink2020state}
M.~Steinbrink, P.~Koch, S.~May, B.~Jung, and M.~Schmidpeter, ``State machine
  for arbitrary robots for exploration and inspection tasks,'' in
  \emph{Proceedings of the 2020 4th International Conference on Vision, Image
  and Signal Processing}, 2020, pp. 1--6.

\bibitem{maruyama2016exploring}
Y.~Maruyama, S.~Kato, and T.~Azumi, ``Exploring the performance of ros2,'' in
  \emph{Proceedings of the 13th International Conference on Embedded Software},
  2016, pp. 1--10.

\bibitem{rovida2017skiros}
F.~Rovida, M.~Crosby, D.~Holz, A.~S. Polydoros, B.~Gro{\ss}mann, R.~Petrick,
  and V.~Kr{\"u}ger, ``{SkiROS} — a skill-based robot control platform on top
  of {ROS},'' in \emph{Robot operating system (ROS)}.\hskip 1em plus 0.5em
  minus 0.4em\relax Springer, 2017, pp. 121--160.

\bibitem{rovida2015design}
F.~Rovida and V.~Kr{\"u}ger, ``Design and development of a software
  architecture for autonomous mobile manipulators in industrial environments,''
  in \emph{ICIT}, 2015.

\bibitem{laker2012blackboard}
P.~Laker, ``Blackboard design pattern,'' \url{
  https://social.technet.microsoft.com/wiki/contents/articles/13215.blackboard-design-pattern.aspx},
  2012.

\bibitem{tolvanen2019domain}
J.-P. Tolvanen and S.~Kelly, ``How domain-specific modeling languages address
  variability in product line development: Investigation of 23 cases,'' in
  \emph{23rd International Systems and Software Product Line Conference}, ser.
  SPLC, 2019.

\bibitem{DBLP:conf/icse/AlamiDW18}
A.~Alami, Y.~Dittrich, and A.~Wasowski, ``Influencers of quality assurance in
  an open source community,'' in \emph{11th International Workshop on
  Cooperative and Human Aspects of Software Engineering ({CHASE})}, 2018.

\bibitem{moore1956gedanken}
E.~F. Moore \emph{et~al.}, ``Gedanken-experiments on sequential machines,''
  \emph{Automata studies}, vol.~34, pp. 129--153, 1956.

\bibitem{bosch1997design}
J.~Bosch, ``Design patterns as language constructs,'' \emph{Journal of Object
  Oriented Programming}, 1997.

\bibitem{whittle2013state}
J.~Whittle, J.~Hutchinson, and M.~Rouncefield, ``The state of practice in
  model-driven engineering,'' \emph{IEEE software}, vol.~31, no.~3, pp. 79--85,
  2013.

\bibitem{cruz2010impact}
J.~A. Cruz-Lemus, A.~Maes, M.~Genero, G.~Poels, and M.~Piattini, ``The impact
  of structural complexity on the understandability of {UML} statechart
  diagrams,'' \emph{Information Sciences}, vol. 180, no.~11, pp. 2209--2220,
  2010.

\bibitem{genero2002defining}
M.~Genero, D.~Miranda, and M.~Piattini, ``Defining and validating metrics for
  {UML} statechart diagrams,'' in \emph{QAOOSE}, 2002.

\bibitem{cruz2005investigating}
J.~A. Cruz-Lemus, M.~Genero, M.~Piattini, and A.~Toval, ``Investigating the
  nesting level of composite states in {UML} statechart diagrams,'' \emph{Proc.
  QAOOSE}, vol.~5, pp. 97--108, 2005.

\bibitem{cruz2005empirical}
------, ``An empirical study of the nesting level of composite states within
  {UML} statechart diagrams,'' in \emph{International Conference on Conceptual
  Modeling}.\hskip 1em plus 0.5em minus 0.4em\relax Springer, 2005, pp. 12--22.

\bibitem{brugali2009component}
D.~Brugali and P.~Scandurra, ``Component-based robotic engineering (part
  i)[tutorial],'' \emph{IEEE Robotics \& Automation Magazine}, vol.~16, no.~4,
  pp. 84--96, 2009.

\bibitem{dubinsky2013exploratory}
Y.~Dubinsky, J.~Rubin, T.~Berger, S.~Duszynski, M.~Becker, and K.~Czarnecki,
  ``An exploratory study of cloning in industrial software product lines,'' in
  \emph{CSMR}, 2013.

\bibitem{gherardi2014modeling}
L.~Gherardi and D.~Brugali, ``Modeling and reusing robotic software
  architectures: The hyperflex toolchain,'' in \emph{ICRA}, 2014.

\bibitem{schlegel2015model}
C.~Schlegel, A.~Lotz, M.~Lutz, D.~Stampfer, J.~F. Ingl{\'e}s-Romero, and
  C.~Vicente-Chicote, ``Model-driven software systems engineering in robotics:
  covering the complete life-cycle of a robot,'' \emph{it-Information
  Technology}, vol.~57, no.~2, pp. 85--98, 2015.

\bibitem{wigand2017modularization}
D.~L. Wigand, A.~Nordmann, M.~Goerlich, and S.~Wrede, ``Modularization of
  domain-specific languages for extensible component-based robotic systems,''
  in \emph{IRC}, 2017.

\bibitem{robotics2017robotics}
{SPARC~Robotics}, ``Robotics 2020 multi-annual roadmap for robotics in
  {E}urope,'' \emph{SPARC Robotics, EU-Robotics AISBL, The Hauge, The
  Netherlands}, 2017.

\end{thebibliography}

%

\vskip -2\baselineskip
\begin{IEEEbiography}[{\includegraphics[width=1in,height=1.25in,clip,keepaspectratio]{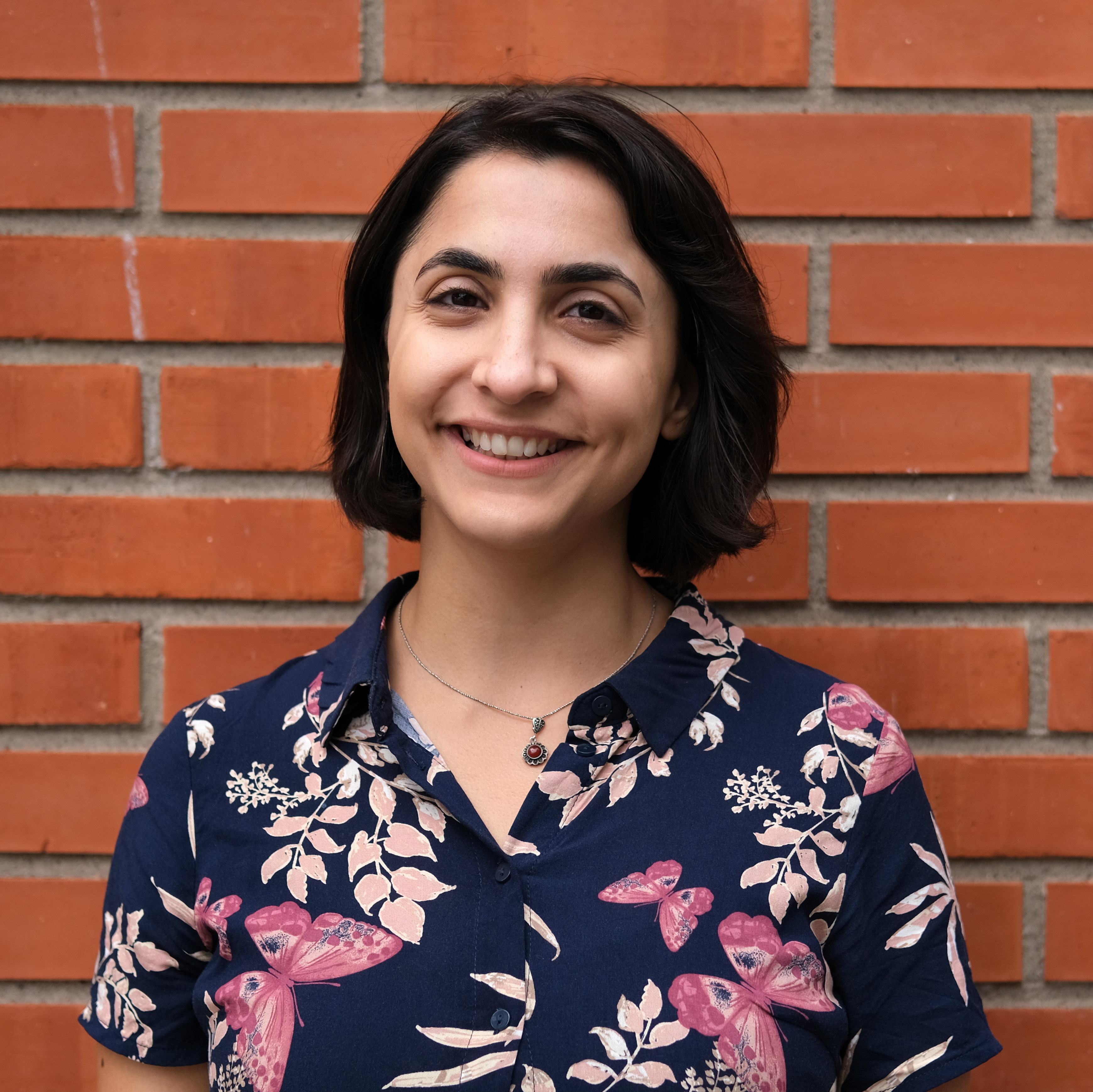}}]{Razan Ghzouli} is currently working towards a PhD degree at Chalmers University of Technology, Gothenburg, Sweden, where she is part of the software engineering division at the Department of Computer Science and Engineering.
She received her master degree in applied data science from the university  of Gothenburg, Sweden and her bachelor degree in computer and automation engineering from Damascus university, Syria. 
Her PhD focuses on facilitating the migration to model-based design and systems to enable reusable and maintainable robotic missions.
\end{IEEEbiography}

\vskip -2\baselineskip
\begin{IEEEbiography}[{\includegraphics[width=1in,height=1.25in,clip,keepaspectratio]{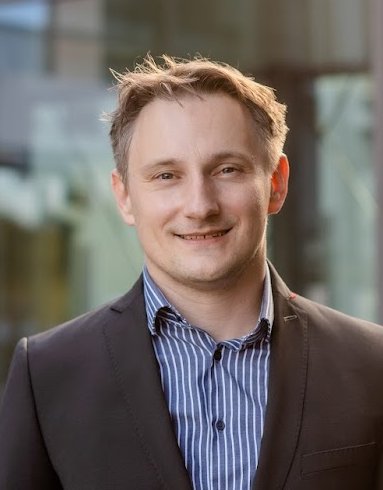}}]{Thorsten Berger} is a Professor in Computer Science at Ruhr University Bochum in Germany. After receiving the PhD degree from the University of Leipzig in Germany in 2013, he was a Postdoctoral Fellow at the University of Waterloo in Canada and the IT University of Copenhagen in Denmark, and then an Associate Professor jointly at Chalmers University of Technology and the University of Gothenburg in Sweden. He received competitive grants from the Swedish Research Council, the Wallenberg Autonomous Systems Program, Vinnova Sweden (EU ITEA), and the European Union. He is a fellow of the Wallenberg Academy---one of the highest recognitions for researchers in Sweden. He received two \emph{best-paper} and two \emph{most-influential-paper} awards. His service was recognized with distinguished reviewer awards at the tier-one conferences ASE 2018 and ICSE 2020, and at SPLC 2022. His research focuses on model-driven software engineering, program analysis, and empirical software engineering. 
\end{IEEEbiography}

\vskip -2\baselineskip
\begin{IEEEbiography}[{\includegraphics[width=1in,height=1.25in,clip,keepaspectratio]{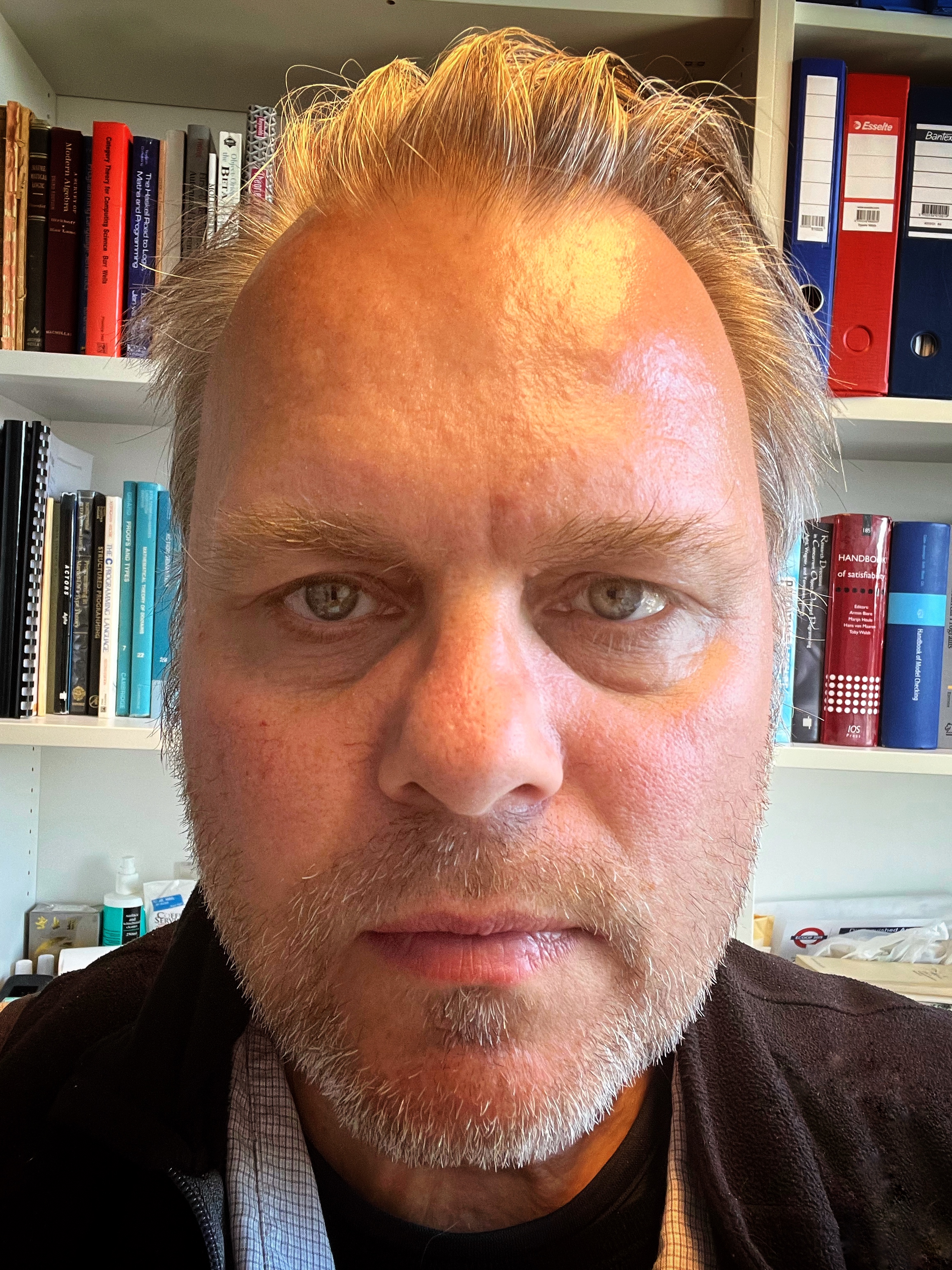}}]{Einar Broch Johnsen}
  is a Professor at the Department of Informatics of the University of
  Oslo in Norway.  He is the strategy director of Sirius, a center for
  research-driven innovation with long-term funding from the Research
  Council of Norway. He has been prominently involved in many national
  and European research projects; in particular, he was the
  coordinator of the EU FP7 project Envisage (2013-2016) on formal
  methods for cloud computing and the scientific coordinator of the EU
  H2020 project HyVar (2015-2018) on hybrid variability systems.  His
  research focuses on formal methods, programming models and
  methodology, and model-based analysis in domains such as distributed and
  concurrent systems, cloud computing, digital twins and robotics.
\end{IEEEbiography}

\vskip -2\baselineskip
\begin{IEEEbiography}[{\includegraphics[width=1in,height=1.25in,clip,keepaspectratio]{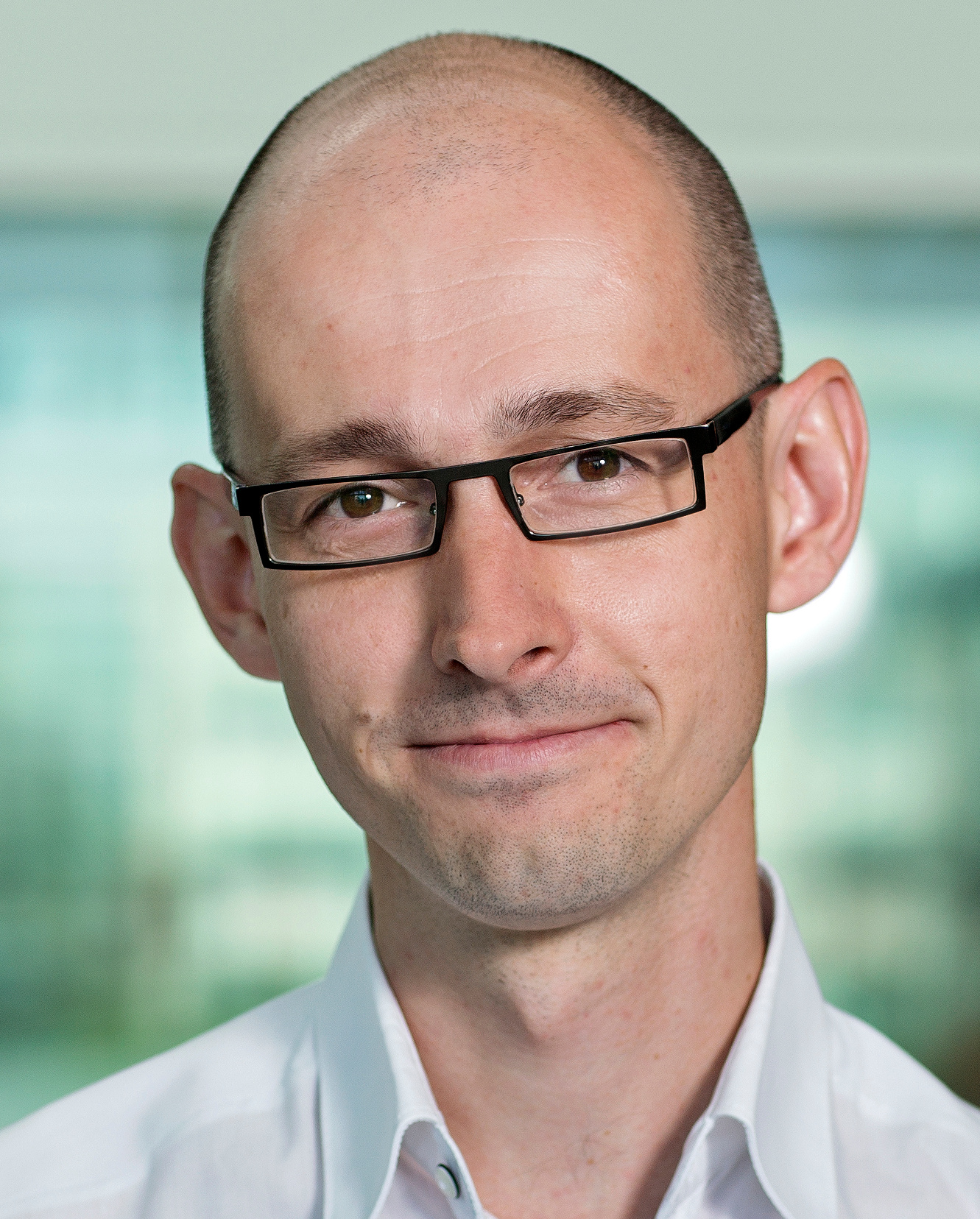}}] {Andrzej Wasowski} is Professor of Software Engineering at the IT University of Copenhagen. He has also worked at Aalborg University in Denmark, and as visiting professor at INRIA Rennes and University of Waterloo, Ontario. His interests are in software quality, reliability, and safety in high-stake high-value software projects. This includes semantic foundations and tool support for model-driven development, program analysis tools, testing tools and methods, as well as processes for improving and maintain quality in software projects.  Many of his projects involve commercial or open-source partners, primarily in the domain of robotics and safety-critical embedded systems. Recently he coordinates the Marie-Curie training network on Reliable AI for Marine Robotics (REMARO). Wasowski holds a PhD degree from the IT University of Copenhagen, Denmark (2005) and a MSC Eng degree from the Warsaw University of Technology, Poland (2000).
\end{IEEEbiography}

\vskip -2\baselineskip
\begin{IEEEbiography}
  [{\includegraphics[width=1in,height=1.25in, trim={100mm 0 60mm 0},
    clip,keepaspectratio]{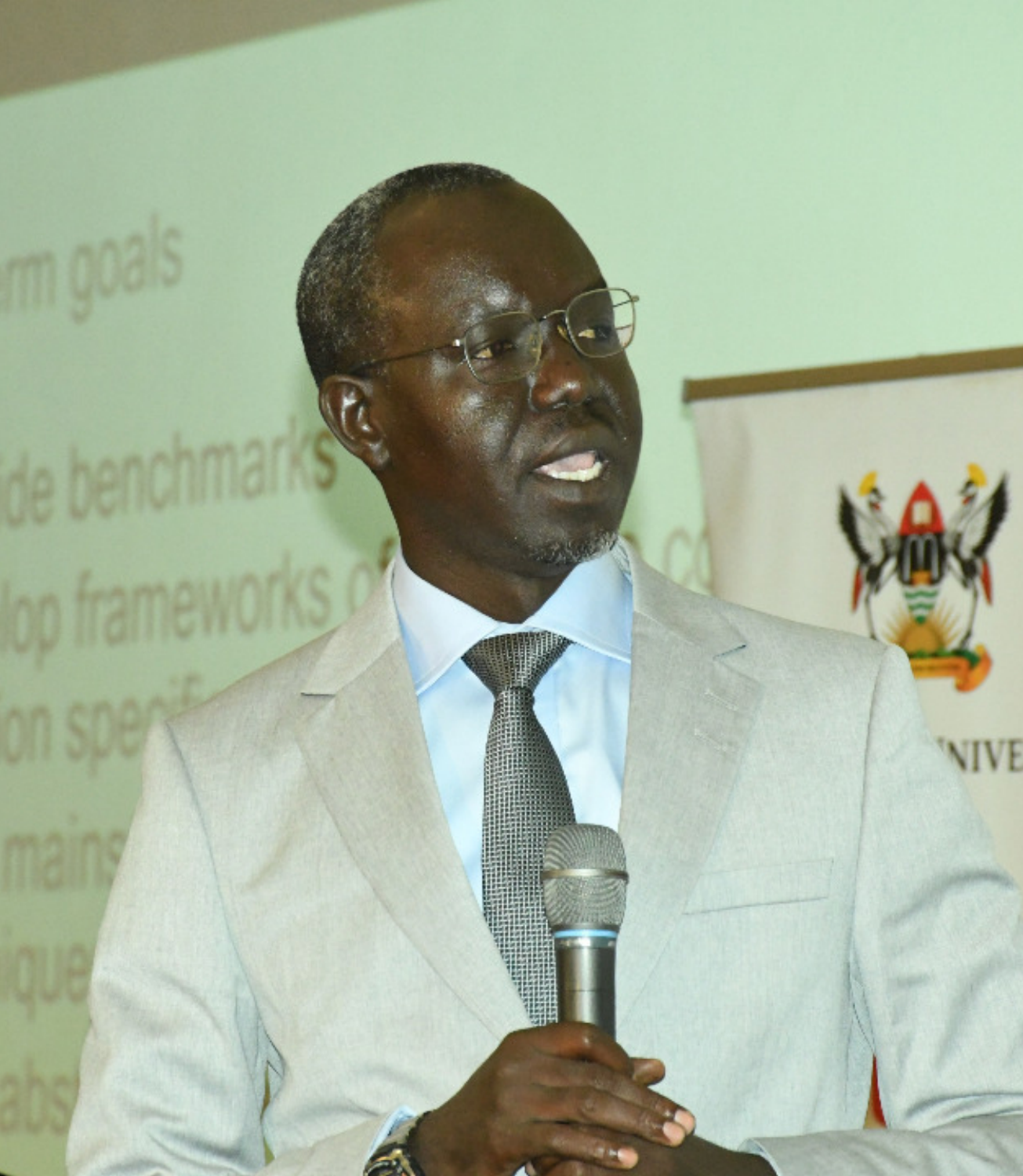}}]{Swaib Dragule} is
  a PhD Fellow in Computer Science and Software Engineering at
  Chalmers University of Technology and Makerere University. He holds
  MSc. and BSc. in computer science. He is an academic staff of
  Makerere university, College of Computing and Information
  Sciences. His research interests are in programming languages,
  domain-specific languages, and robotics.
\end{IEEEbiography}






\end{document}